\documentclass{article}
\pdfoutput=1

\usepackage[preprint]{neurips_2023}

\usepackage[utf8]{inputenc} 
\usepackage[T1]{fontenc}    
\usepackage{hyperref}       
\usepackage{url}            
\usepackage{booktabs}       
\usepackage{amsfonts}       
\usepackage{nicefrac}       
\usepackage{microtype}      
\usepackage{xcolor}         

\usepackage{graphicx}%
\usepackage{multirow}%
\usepackage{amsmath,amssymb,amsfonts}%
\usepackage{amsthm}%
\usepackage{mathrsfs}%
\usepackage[title]{appendix}%
\usepackage{textcomp}%
\usepackage{manyfoot}%
\usepackage{booktabs}%
\usepackage{algorithm}%
\usepackage{algorithmicx}%
\usepackage{algpseudocode}%
\usepackage{listings}%

\usepackage[labelfont=bf]{caption}
\usepackage{adjustbox}
\usepackage{enumitem}
\usepackage{cleveref}

\hypersetup{
    colorlinks,
    linkcolor={red!50!black},
    citecolor={blue!50!black},
    urlcolor={blue!80!black}
}

\newif\ifshowpartitle 
\newcommand{\partitle}[1]{\ifshowpartitle\textbf{#1}\fi} 
\showpartitletrue 

\newif\ifshowtables 
\showtablestrue 

\title{Beyond Preferences in AI Alignment}

\author{%
  Tan Zhi-Xuan \\
  MIT \\
  \And
  Micah Carroll \\
  UC Berkeley \\
  \And
  Matija Franklin \\
  University College London \\
  \And
  Hal Ashton \\
  University of Cambridge \\
}

\begin{document}

\maketitle

\begin{abstract}
The dominant practice of AI alignment assumes (1) that preferences are an adequate representation of human values, (2) that human rationality can be understood in terms of maximizing the satisfaction of preferences, and (3) that AI systems should be aligned with the preferences of one or more humans to ensure that they behave safely and in accordance with our values. Whether implicitly followed or explicitly endorsed, these commitments constitute what we term a  \emph{preferentist} approach to AI alignment. In this paper, we characterize and challenge the preferentist approach, describing conceptual and technical alternatives that are ripe for further research. We first survey the limits of rational choice theory as a descriptive model, explaining how preferences fail to capture the thick semantic content of human values, and how utility representations neglect the possible incommensurability of those values. We then critique the normativity of expected utility theory (EUT) for humans and AI, drawing upon arguments showing how rational agents need not comply with EUT, while highlighting how EUT is silent on which preferences are normatively acceptable. Finally, we argue that these limitations motivate a reframing of the targets of AI alignment: Instead of alignment with the preferences of a human user, developer, or humanity-writ-large, AI systems should be aligned with normative standards appropriate to their social roles, such as the role of a general-purpose assistant. Furthermore, these standards should be negotiated and agreed upon by all relevant stakeholders. On this alternative conception of alignment, a multiplicity of AI systems will be able to serve diverse ends, aligned with normative standards that promote mutual benefit and limit harm despite our plural and divergent values.
\end{abstract}

\section{Introduction}

Recent progress in the capabilities of AI systems, as well as their increasing adoption in society, has led a growing number of researchers to worry about the impact of AI systems that are misaligned with human values. The roots of this concern vary, with some focused on the existential risks that may come with increasingly powerful autonomous systems \citep{carlsmith2022power}, while others take a broader view of the dangers and opportunities presented by potentially transformative AI technologies \citep{prunkl2020beyond,lazar2023ai}. To address these challenges, AI alignment has emerged as a field, focused on the technical project of ensuring an AI system acts reliably in accordance with the values of one or more humans.

Yet terms like ``human values'' are notoriously imprecise, and it is unclear how to operationalize ``values'' in a sufficiently precise way that a machine could be aligned with them. One prominent approach is to define ``values'' in terms of human preferences, drawing upon the traditions of rational choice theory \citep{mishra2014decision}, statistical decision theory \citep{berger2013statistical}, and their subsequent influence upon automated decision-making and reinforcement learning in AI \citep{sutton2018reinforcement}. Whether explicitly adopted, or implicitly assumed in the guise of ``reward'' or ``utility'', this preference-based approach dominates both the theory and practice of AI alignment. However, as proponents of this approach note themselves, aligning AI with human preferences faces numerous technical and philosophical challenges, including the problems of social choice, anti-social preferences, preference change, and the difficulty of inferring preferences from human behavior \citep{russell2019human}.

In this paper, we argue that to truly address such challenges, it is necessary to go beyond formulations of AI alignment that treat human preferences as ontologically, epistemologically, or normatively basic. Borrowing a term from the philosophy of welfare \citep{baber2011preference}, we identify these formulations as part of a broadly \emph{preferentist} approach to AI alignment, which we characterize in terms of four theses about the role of preferences in both descriptive and normative accounts of (human-aligned) decision-making:

\begin{description}[topsep=6pt,itemsep=6pt,leftmargin=0pt,labelsep=3pt]
    \item[\textbf{Rational Choice Theory as a Descriptive Framework.}] Human behavior and decision-making is well-modeled as approximately maximizing the satisfaction of preferences, which can be represented as a utility or reward function.

    \item[\textbf{Expected Utility Theory as a Normative Standard.}] Rational agency can be characterized as the maximization of expected utility. Moreover, AI systems should be designed and analyzed according to this normative standard.

    \item[\textbf{Single-Principal Alignment as Preference Matching.}] For an AI system to be aligned to a single human principal, it should act so as to maximize the satisfaction of the preferences of that human.

    \item[\textbf{Multi-Principal Alignment as Preference Aggregation.}] 
    For AI systems to be aligned to multiple human principals, they should act so as to maximize the satisfaction of their aggregate preferences.
\end{description}

These four theses represent a \emph{cluster} of views, not a unified theory of AI alignment. Still, the ideas they represent are tightly linked, and most approaches to AI alignment assume two or more of the theses. For example, inverse reinforcement learning \citep{ng2000algorithms,hadfield2016cooperative}, reinforcement learning from human feedback \citep{akrour2014programming,christiano2017deep,ouyang2022training}, and direct preference optimization \citep{rafailov2024direct,hejna2024contrastive} all assume that human preferences are well-modeled by a reward or utility function, which can then be optimized to produce aligned behavior. Similarly, worries about deceptive alignment \citep{hubinger2019risks} and goal misgeneralization \citep{langosco2022goal} are typically characterized as a mismatch between a learned utility function and the human-intended utility function; the solution is thus to ensure that the utility functions (and the preferences they represent) are closely matched.

Of course, preferentism in AI alignment is not without its critics. There has been considerable discussion as to whether its component theses are warranted \citep{shah2018coherence,eckersley2018impossibility,hadfield-menell2018incomplete,wentworth2019subagents,wentworth2023subagents,gabriel2020artificial,vamplew2022scalar,korinek2022aligned,garrabrant2022geometric,thornley2023coherence}, echoing similar debates in economics, decision theory, and philosophy. Nonetheless, it is apparent that the dominant practice of AI alignment has yet to absorb the thrust of these debates. Consequently, we believe it is worthwhile to identify the descriptive and normative commitments of preferentist approaches, to state clearly their limitations, and to describe conceptual and technical alternatives that are ripe for further research\ifshowtables\ (Table \ref{tab:preferentist_theses})\fi.

\subsection{Overview}

The rest of this paper is organized as follows: In \textbf{Section \ref{sec:rat_choice}}, we examine rational choice theory as a descriptive account of human decision-making. Drawing upon the tradition of revealed preferences in economics, rational choice theory is often taken for granted by AI researchers seeking to learn human preferences from behavior. In doing so, they assume that human behavior can be modeled as the (approximate) maximization of expected utility, that human preferences can be represented as utility or reward functions, and that preferences are an adequate representation of human values. We challenge each of these assumptions, offering alternatives that better account for resource-limited human cognition, incommensurable values, and the constructed nature of our preferences.

Developing upon these ideas, in \textbf{Section \ref{sec:eut_rat}} we turn to expected utility theory (EUT) as a normative standard of rationality. Even while recognizing that humans often do not comply with this standard, alignment researchers have traditionally assumed that sufficiently advanced AI systems will do so, and hence that solutions to AI alignment must be compatible with EUT. In parallel with recent critiques of this view \citep{thornley2023coherence,thornley2024shutdown,bales2023will,petersen2023invulnerable}, we argue that EUT is both unnecessary and insufficient for rational agency, and hence limited as both a design strategy and analytical lens. Instead of adhering to utility theory, we can design tool-like AI systems with \emph{locally coherent} preferences that are not representable as a utility function. We can also go beyond EUT, building systems that \emph{reason} about preferences in accordance with deeper normative principles.

\ifshowtables
\begin{table}[t]
\centering
\begin{adjustbox}{center, max width=\paperwidth}
\footnotesize
\begin{tabular}{p{3.5cm}p{4.5cm}p{4.5cm}}
\toprule
\textbf{Assumption} & \textbf{Limitations} & \textbf{Alternatives} \\ 
\midrule
\begin{minipage}[t]{\linewidth}
\textbf{\emph{Rational Choice Theory as a Descriptive Framework}} \\
A person's behavior is well-modeled as (approximate) maximization of a utility function representing their preferences.
\end{minipage}
&
\begin{minipage}[t]{\linewidth}
\begin{itemize}[topsep=0pt,itemsep=0pt,leftmargin=0pt]
    \item Humans are not perfectly rational.
    \item Humans are not noisily rational.
    \item Reward or utility functions cannot represent all human preferences.
    \item Preferences do not capture the semantics of human values and reasons, or value commensuration.
\end{itemize}
\end{minipage}
&
\begin{minipage}[t]{\linewidth}
\begin{itemize}[topsep=0pt,itemsep=0pt,leftmargin=0pt]
    \item Resource-rational human models.
    \item Multi-objective and partial order representations of preferences.
    \item Learning the semantics of evaluative and normative concepts.
    \item Modeling how people do or do not commensurate their values.
\end{itemize}
\end{minipage}
\\
\midrule
\begin{minipage}[t]{\linewidth}
\textbf{\emph{Expected Utility Theory as a Normative Standard}} \\
Rational agency consists in maximizing expected utility, and AI systems should be designed or analyzed accordingly.
\end{minipage}
&
\begin{minipage}[t]{\linewidth}
\begin{itemize}[topsep=0pt,itemsep=0pt,leftmargin=0pt]
    \item EUT-style global coherence is not rationally required.
    \item EUT analyses are only weakly informative about AI behavior.
    \item Globally coherent agents are not the only viable design target.
    \item EUT does not explain how (normative) reasons shape our preferences.
\end{itemize}
\end{minipage}
&
\begin{minipage}[t]{\linewidth}
\begin{itemize}[topsep=0pt,itemsep=0pt,leftmargin=0pt]
    \item Agents with incomplete preferences can avoid exploitation.
    \item Mechanistic, economic, or evolutionary analyses of AI behavior.
    \item Locally coherent agents may better preserve tool-like corrigibility.
    \item Theories of normative reasoning can be integrated with AI systems.
\end{itemize}
\end{minipage}
\\
\midrule
\begin{minipage}[t]{\linewidth}
\textbf{\emph{Single-Principal Alignment as Preference Matching}} \\
Alignment with a single person consists in acting to satisfy their preferences.
\end{minipage}
&
\begin{minipage}[t]{\linewidth}
\begin{itemize}[topsep=0pt,itemsep=0pt,leftmargin=0pt]
    \item Alignment with ``human rewards'' assumes that preferences are static, complete, acontextual, and asocial.
    \item Unclear what ``alignment'' means once a person's preferences change or conflict across contexts.
\end{itemize}
\end{minipage}
&
\begin{minipage}[t]{\linewidth}
\begin{itemize}[topsep=0pt,itemsep=0pt,leftmargin=0pt]
    \item For locally scoped AI systems, alignment with normative criteria specific to the task, context, or role.
    \item For globally scoped AI assistants, alignment with the normative ideal of a good assistant.
\end{itemize}
\end{minipage}
\\
\midrule
\begin{minipage}[t]{\linewidth}
\textbf{\emph{Multi-Principal Alignment as Preference Aggregation}} \\
Alignment with society or multiple people consists in acting to satisfy their aggregate preferences.
\end{minipage}
&
\begin{minipage}[t]{\linewidth}
\begin{itemize}[topsep=0pt,itemsep=0pt,leftmargin=0pt]
    \item Aggregation of elicited preferences may not track aggregate value, welfare, or normative acceptability.
    \item Optimizing aggregate preferences is intractable, incentive incompatible.
    \item Aggregation is at odds with the plurality of AI uses \& human interests.
\end{itemize}
\end{minipage}
&
\begin{minipage}[t]{\linewidth}
\begin{itemize}[topsep=0pt,itemsep=0pt,leftmargin=0pt]
    \item Prioritarian, egalitarian, or contractualist elicitation of normative judgments and principles.
    \item Alignment with social, legal, moral norms given our divergent interests.
    \item A plurality of normative standards for a plurality of AI systems.
\end{itemize}
\end{minipage}
\\
\bottomrule
\end{tabular}
\end{adjustbox}
\vspace{3pt}
\caption{Four theses that characterize the preferentist approach to AI alignment, along with a summary of their limitations and alternatives.}
\label{tab:preferentist_theses}
\vspace{-20pt}
\end{table}
\fi

After interrogating these descriptive and normative foundations, in \textbf{Section~\ref{sec:single_align}} we consider what this implies for aligning AI with a single human principal. Since reward functions may not capture even a single human's values, the practice of reward learning is unsuitable beyond narrow tasks and contexts where people are willing to commensurate their values. Furthermore, since preferences are dynamic and contextual, they cannot serve as the alignment target for broadly-scoped AI systems. Rather, alignment with an individual person should be reconceived as alignment with the normative ideal of an assistant. More generally, AI systems should not be aligned with preferences, but with the normative standards appropriate to their social roles and functions \citep{kasirzadeh2023conversation}.

If normative standards are to serve as alignment targets, whose judgments do we consider in determining these (oft-contested) standards? We take up this final topic in \textbf{Section \ref{sec:multi_align}}, critiquing naive preference aggregation as an approach to aligning AI with multiple human principals \citep{fickinger2020multi-principal}. Despite increasing recognition that this approach is inadequate \citep{critch2020ai,gabriel2020artificial,korinek2022aligned}, applied alignment techniques typically aggregate preferences across multiple individuals, overlooking the contested and plural nature of human values, while conflating norm-specific judgments with all-things-considered preferences. As alternatives, we argue that contractualist and agreement-based approaches can better handle value contestation while respecting the individuality of persons and the plurality of uses we have for AI. This motivates a reframing of the aims of AI alignment as they have often been conceived: Our task is not to align a single powerful AI system with the preferences of humanity writ large, but to align a multiplicity of AI systems with the norms we agree that each system should abide by \citep{zhixuan2022what}.

A note on methodology: Whereas most philosophy papers tend to be narrow in scope, this paper is intentionally broad; it covers a wide range of connected topics, and hence makes arguments that are relatively brief. Our aim is not provide a decisive argument for any particular thesis, but to provide a critical review of the role of preferences in AI alignment, while developing a research agenda for alternative approaches that is accessible to an interdisciplinary audience.

\section{Beyond rational choice theory when modeling humans}\label{sec:rat_choice}

\ifshowtables
\begin{table}[h]
\centering
\footnotesize
\begin{adjustbox}{center, max width=\paperwidth}
\begin{tabular}{p{3.5cm}p{4.5cm}p{4.5cm}}
\toprule
\textbf{Assumption} & \textbf{Limitations} & \textbf{Alternatives} \\ 
\midrule
(Noisily) rational models of human decision making. &
\begin{minipage}[t]{\linewidth}
\begin{itemize}[topsep=0pt,itemsep=0pt,leftmargin=0pt]
    \item Failure to account for (systematic) deviations from optimality.
    \item Failure to model resource bounds on human cognition.
\end{itemize}
\end{minipage}
&
\begin{minipage}[t]{\linewidth}
\begin{itemize}[topsep=0pt,itemsep=0pt,leftmargin=0pt]
    \item Resource rationality as a unifying frame for cognitive biases.
   \item Resource rationality as an inductive bias in models of human decisions.
\end{itemize}
\end{minipage}
\\
\midrule
Reward/utility functions as preference representations. &
\begin{minipage}[t]{\linewidth}
\begin{itemize}[topsep=0pt,itemsep=0pt,leftmargin=0pt]
    \item Markovian rewards cannot express time-extended preferences.
    \item Assumes \emph{complete} preferences.
    \item Non-identifiable w/o more structure.
\end{itemize}
\end{minipage}
&
\begin{minipage}[t]{\linewidth}
\begin{itemize}[topsep=0pt,itemsep=0pt,leftmargin=0pt]
    \item Temporal logics \& reward machines for temporal preference structure.
    \item Vector/interval-valued utilities, CP-nets can model incompleteness.
\end{itemize}
\end{minipage}
\\
\midrule
Preferences as representations of human values. &
\begin{minipage}[t]{\linewidth}
\begin{itemize}[topsep=0pt,itemsep=0pt,leftmargin=0pt]
    \item Human preferences are \emph{constructed} from reasons \& values, not basic.
    \item Failure to model the semantics of human values / evaluative concepts.
\end{itemize}
\end{minipage}
&
\begin{minipage}[t]{\linewidth}
\begin{itemize}[topsep=0pt,itemsep=0pt,leftmargin=0pt]
    \item Learning evaluative concepts as ``input features'' to decision-making.
    \item Modeling preference construction as value (non)-commensuration.
\end{itemize}
\end{minipage}
\\
\bottomrule
\end{tabular}
\end{adjustbox}
\vspace{3pt}
\caption{Assumptions, limitations, and alternatives to rational choice theory as a descriptive framework for modeling human preferences, values, and decision making.}
\label{tab:behavioral-preferentism}
\end{table}
\fi

The central tenet of rational choice theory is the assumption that humans act so as to maximize the satisfaction of their preferences, and that both individual and aggregate human behavior can be understood in these terms. As far as theoretical presuppositions go, this assumption has been wildly successful, forming the bedrock of modern economics as a discipline, and influencing a great variety of fields concerned with analyzing human behavior, including sociology \citep{boudon2003beyond}, law \citep{ulen1999rational}, and cognitive science \citep{chater1999ten,jara2020naive}.

\partitle{Revealed preferences and their representation as utility functions.} In its most standard form, rational choice theory assumes that human preferences can be represented as a scalar-valued utility function defined over outcomes --- that is, in terms of a quantity that can be maximized --- and that human choice can be modeled as selecting actions so as to maximize the expected value of this function. The promise this offers is that we can directly derive what a person prefers from what they choose, and furthermore represent how much they prefer it as a scalar value. Such preferences are called \emph{revealed preferences}, because they are supposedly revealed through what a person chooses. This methodology is bolstered by numerous representation theorems \citep{savage1972foundations,bolker1967simultaneous,jeffrey1991logic} showing that \emph{any} preference ordering over outcomes that obeys certain ``rationality axioms'' can be represented in terms of a utility function, such as the famous von Neumann–Morgenstern (VNM) utility theorem \citep{von_neumann1944theory}.

\partitle{Rational choice theory in machine learning.} In keeping with rational choice theory, many machine learning and AI systems also assume that human preferences can be derived from human choices in a more or less direct manner, and furthermore represent those preferences in terms of scalar utilities or rewards. This is most pronounced in the fields of inverse reinforcement learning \citep{ng2000algorithms,abbeel2004apprenticeship,hadfield2016cooperative} and reinforcement learning from human feedback \citep{christiano2017deep,zhu2023principled}, which explicitly assume that the behavior of a human can be described as (approximately) maximizing a sum of scalar rewards over time, and then try to infer a reward function that explains the observed behavior. Similar assumptions can be found in the field of recommender systems \citep{thorburn2022what}, with many papers modeling recommendation as the problem of showing items to users that they are most likely to engage with, which is presumed to be the item they find the most rewarding \citep{li2010contextual,hill2017efficient,mcinerney2018explore}.

\partitle{Boltzmann models of noisily-rational choice.} While these preference-based models of human behavior are rooted in rational choice theory, it is worth noting that they are slightly more complex than ``maximize expected utility'' might imply. In particular, they allow for the fact that humans may not \emph{always} maximize utility, and hence are models of \emph{noisy} or \emph{approximately} rational choice. In machine learning and AI alignment, the most common of such choice models is called \emph{Boltzmann rationality} (after the Boltzmann distribution in statistical mechanics), which assumes that the probability of a choice $c$ is proportional to the exponential of the expected utility of taking that choice:
\begin{equation}
    P(c) \propto \exp\left(\beta \mathbb{E}[U(c)]\right)
\label{eq:boltzmann-rationality}
\end{equation}

\partitle{Justifications and extensions of Boltzmann rationality.} This choice model exhibits a number of practically useful and theoretically appealing properties. For example, by varying the ``rationality parameter'' $\beta$ between zero and infinity, Boltzmann rationality interpolates between completely random choice and deterministic optimal choice \citep{ghosal2023effect}. As an instantiation of Luce's choice axiom \citep{luce1979individual}, it obeys independence of irrelevant alternatives.\footnote{That is, choosing $x$ out of the set $\{x, y, z\}$ has the same probability as first choosing $\{x, y\}$ out of the full set, then choosing $x$ out of $\{x, y\}$.} Boltzmann rationality has also been justified as the maximum entropy distribution\footnote{Maximum entropy distributions are minimally informative in the information theoretic sense, and hence are often advocated for as ``ignorance priors'' in statistical analyses \citep{jaynes1968prior}.} that matches certain constraints implied by observed behavior \citep{ziebart2008maximum, ziebart2010modeling}, or as a thermodynamically-inspired model of bounded rationality where agents have to spend energy investigating which choice leads to the highest utility \citep{ortega2013thermodynamics,jarrett2021inverse}. In addition, Boltzmann rationality has been extended to model other aspects of human behavior besides goal-directed actions, including direct comparisons between options (i.e. stated preferences) \citep{akrour2014programming,christiano2017deep,zhu2023principled}, explicitly stated reward functions \citep{hadfield2017inverse}, entire behavior policies \citep{laidlaw2022boltzmann}, and linguistic utterances \citep{lin2022inferring}, allowing preferences to be inferred from multiple forms of human feedback \citep{jeon2020reward-rational}.

\partitle{Limitations of Boltzmann rationality.} As useful as Boltzmann rationality may be, however, we believe it is important to seek alternatives. For one, it is not the only intuitively plausible model of noisily rational choice: Random-utility models instead model choice as the result of maximization over randomly perturbed utility values, and are widely used in marketing research \citep{horowitz1994advances,azari_soufiani2013generalized}. More crucially, noisy rationality is not enough to account for the full set of ways in which humans fail to act optimally. Richer models of bounded rationality are necessary to accurately infer human preferences and values from their behavior. Most fundamentally, the contents of human motivation are not entirely reducible to bare preferences or utility functions. Instead, we need to enrich our models of human rationality to encompass all the ways in which humans are guided by \emph{reasons for acting}, including the thick evaluative concepts that we apply when deciding between courses of action \citep{blili2024unsocial}. We elaborate upon these limitations in the following sections.

\subsection{Beyond noisily-rational models of human decisions}

The issue with both perfect and noisily-rational models of human decision-making is that they do not account for the systematic deviations from optimality that humans in fact exhibit. As a long line of psychological and behavioral research has shown, humans are \emph{boundedly} rational at best, exhibiting satisficing instead of optimizing behavior, \citep{simon1957behavioral, simon1979rational}. These deviations from optimality include framing effects, loss aversion, anchoring biases, and mis-estimation of high and low probabilities -- phenomena which are better modeled by prospect theory  \citep{kahneman1979prospect,tversky1992advances} than standard rational choice theory. More generally, many of the decision problems that people encounter are computationally intractable to solve optimally, making rational choice a implausible model of human behavior \citep{van2008tractable,bossaerts2019uncertainty,camara2022computationally}. Instead, research suggests that humans make use of a variety of heuristics in order to approximately solve the problems they encounter \citep{gigerenzer2008why}.

\partitle{Challenges to modeling bounded rationality.} How might AI systems that infer human preferences and values account for these findings? One approach might be to incorporate a sufficiently long list of known heuristics and biases into our models of human decision-making, thereby ensuring that preferences can be robustly inferred even in the presence of such biases \citep{evans2016learning, chan2021human}. However, this approach is highly contingent upon on our current state of knowledge about human rationality --- what if we miss out important biases in our models, leading to inaccurate predictions and inferences \citep{christiano2015easy,steinhardt2017latent}? As a potential remedy, \citet{shah2019feasibility} suggest \emph{learning} human biases alongside their preferences. But a conceptual difficulty remains: Without any inductive constraints on the types of errors humans are susceptible to, how can we ensure that human biases are accurately learned? As \citet{armstrong2018occam} show, even inductive preferences for more parsimonious models of human decision-making cannot distinguish intuitively plausible hypotheses from observationally-equivalent but implausible hypotheses, such as the possibility that humans are acting \emph{anti}-rationally by minimizing the satisfaction of their preferences.

\partitle{Resource rationality as a unifying frame.} To address these challenges, we suggest -- in line with prior work -- that \emph{resource rational} analyses of human decision-making might provide an answer: Instead of treating human biases and heuristics as idiosyncratic artifacts, resource rationality posits that seemingly irrational human behavior can often be understood as arising from the rational use of limited computational resources \citep{lieder2020resource}.\footnote{Also known as computational rationality \citep{lewis2014computational,gershman2015computational,oulasvirta2022computational}, algorithmic rationality \citep{halpern2015algorithmic}, and bounded optimality \citep{russell1994provably}.} For example, availability biases towards extreme events can be modeled as a form of resource-rational sampling \citep{lieder2018overrepresentation}, susceptibility to sharing inaccurate information can result from a form of rational inattention \citep{pennycook2021shifting,sims2003implications}, and habitual action can be explained as a mechanism for avoiding costly planning under time constraints \citep{keramati2016adaptive}. Resource rationality thus serves as a \emph{generative principle} for hypothesizing possible deviations from standard rationality, and then testing whether such deviations in fact occur in humans.

\partitle{Resource rationality as an inductive bias.} What does this imply for AI alignment? Most practically, the assumption of resource rationality can be embedded as priors over computation time and representational complexity in probabilistic models of human decision-making \citep{zhi2020online,ho2022cognitive,berke2023thinking,jacob2024modeling}, enabling systems to infer human goals and preferences from failed plans and mistaken reasoning \citep{evans2016learning,alanqary2021modeling,chan2021human}, while accelerating the speed of goal inference \citep{zhixuan2024infinite}. Embedding these priors on human resource bounds provides a strong but flexible inductive bias on the the space of decision procedures that humans might employ. Unlike a simplicity prior, this may avoid concerns about the non-identifiability of human preferences \citep{armstrong2018occam}.

\partitle{The normative appeal of resource rationality.} Indeed, the inductive bias imposed by resource rationality has a \emph{normative} appeal over a simplicity-based approach: It tries to make sense of humans as \emph{rational} creatures, aiming for teleological explanations of our behavior instead of reducing us to mere physical phenomena to be explained by the simplest causal mechanism. At the same time, it is a \emph{forgiving} standard of rationality, allowing room for mistakes when inferring preferences from their decisions, while placing greater evidential weight on decisions made after lengthier deliberation. Both of these features make resource rationality a promising framework for systems that learn our values: Rather than directly associating our behavior with our preferences, preferences are associated with how we would act if we were more thoughtful, reflective, and informed.

\subsection{Beyond reward and utility functions as representations of human preferences}

While resource rationality provides a more flexible framework for modeling the relationship between preferences and behavior, this says little about how preferences themselves should be represented. For the most part, resource rational analyses continue to represent human preferences in terms of scalar costs and rewards, or more generally, utility functions, with the primary innovation being the inclusion of costs on computation \citep{lieder2020resource,callaway2022rational}. Yet, there are many reasons to think that reward functions and utility functions are inadequate representations of human preferences, while also tending to produce conceptual confusion about what they do represent.

\partitle{The limited expressivity of reward functions.} These issues are most easily appreciated in the case of (scalar, Markovian) reward functions. As noted earlier, the reward representation assumes that the utility of a sequence of states and actions $\xi = (s_1, a_1, ..., s_T, a_T)$ can be decomposed into a sum of scalar rewards over time:
\begin{equation*}
    U(\xi) = \textstyle\sum_{t=1}^T R(s_t, a_t)
\end{equation*}
Advocates of the reward representation argue that any task accomplishable by an intelligent agent can be framed as a reward maximization problem \citep{silver2021reward}. As \citet{kasenberg2018norms} point out, however, this minimally requires that all historically relevant information is already included in the representation of each state $s_t$ --- a requirement since stated more formally by \citet{abel2021expressivity} and \citet{bowling2023settling}. This means that without careful feature engineering, reward functions cannot easily express time-extended preferences like the desire to keep a promise, or the value of narrative coherence. Separately, the scalar nature of the (standard) reward representation means that it cannot represent the existence of incomplete preferences due to multiple incommensurable scales of value \citep{vamplew2022scalar,anderson1995value,chang1997incommensurability}: Sometimes, the choices before us may seem good or bad in such distinct ways that it makes no sense to say which is better than another.\footnote{For example, one might have to choose between staying in a democratic country while being at severe risk of poverty, or immigrating to a country with material security but no political freedoms.} As a result, we may have \emph{preferential gaps}: pairs of options where neither option is preferred over the other, nor are they equally preferred.

\partitle{Confusion about what reward functions represent.} Alongside these limitations in expressiveness, there is often slippage among AI researchers regarding the ontological status of reward,\footnote{See \citet{lambert2023entangled} for an overview in the context of reinforcement learning from human feedback.} which is sometimes interpreted as the \emph{intrinsic} desirability of a particular state or action \citep{schroeder2004three}, or as a biological signal that promotes learning \citep{butlin2021ai} or evolutionary success \citep{singh2009rewards}, but is also used to define the \emph{instrumental} value of a state (as in reward shaping \citep{ng1999policy,booth2023perils}), or to demarcate goals (i.e. desired trajectories or states of affairs \citep{molinaro2023goal,davidson2024goals}). While this is partly a testament to the flexibility of reward functions as a mathematical formalism, this also means that distinct normative concepts (preferences, goals, intents, desires, values, etc.) get conflated or subsumed under the label of ``reward''. In alignment research, this manifests as the tendency to frame value alignment in terms of reward learning \citep{hadfield2016cooperative,leike2018scalable}, and to formalize concepts like ``goals'' \citep{langosco2022goal} and ``intents'' \citep{ouyang2022training} as reward functions. This is despite the existence of other useful and potentially more appropriate formalisms, such as the formalization of goals as logical specifications \citep{fikes1971strips}, and the formalization of intentions as (partial) plans \citep{bratman1987intention,bratman1988plans}.

\partitle{Utility functions are more expressive, but insufficiently constrained.} While not without their own interpretive confusions,\footnote{Most prominently, the debate between interpreting utility as cardinal measure of welfare that is comparable across individuals, versus a mere representation of individual preference rankings \citep{strotz1953cardinal,harsanyi1953cardinal}} utility functions are considerably more general than (Markovian) reward functions. For example, they can be defined over arbitrarily long sequences of states, allowing them to capture time-extended preferences. However, what utility functions buy in terms of expressiveness comes at a cost to both identifiability and tractability: If no constraints are placed on the structure of human utility functions, then given some sequence of actions (e.g. a person buying ten apples, then two oranges), it is not possible to disambiguate a reasonable utility function that explains the actions (e.g. by assigning higher utility to an apple over an orange) from a degenerate utility function that assigns a utility of one to exactly the observed sequence.\footnote{See \citet{armstrong2018occam} for a similar argument. Note that these identifiability problems already exist with Markovian reward functions \citep{cao2021identifiability,kim2021reward,skalse2023invariance}, but are made worse once we let go of the Markov assumption altogether.} In addition, many utility functions are intractable to coherently maximize \citep{camara2022computationally} or even to compute.\footnote{For example, a utility function might embed the NP-hard traveling salesperson problem (TSP), by assigning higher utility to road networks with TSP solutions under a certain cost threshold. While a human could hold such preferences, it would generally be very costly for them to check whether those preferences hold.} If we apply the principle of resource rationality here too, this makes intractable utility functions less plausible representations of human preferences. Finally, utility functions are not without their own expressivity limitations: Like scalar rewards, they assume away preference incompleteness due to plural and incommensurable values \citep{chang2021how,eckersley2018impossibility}. Indeed, empirical work shows that incomplete preferences are not just possible, but actual \citep{cettolin2019revealed,nielsen2023revealed}. This means that utility functions are, at best, \emph{approximate} representations of human preferences, not exact ones.

\partitle{Fundamental tensions for any representation of preferences.} It is worth noting that these tensions between expressivity, structure, and tractability apply to \emph{any} representation of human preferences, not just reward or utility functions. Thus, while it might be tempting to ensure expressivity by directly representing human preferences as a (possibly incomplete) list of comparisons over universe trajectories (or a distribution over such comparisons \citep{dumoulin2023density}), such a list would be extremely space-inefficient, while providing little to no action guidance in novel choice situations. Instead, we should recognize that part of what makes reward and utility functions so useful in practice is that they are typically engineered to be \emph{compact} representations of preferences. Practically useful alternatives should maintain this property, while better capturing the richness of human preferences.

\partitle{Alternative representations can better capture temporal structure and value plurality.} Fortunately, many promising options exist: Temporal logics \citep{kasenberg2018norms} and reward machines \citep{icarte2022reward,davidson2024goals} avoid the limitations of traditional reward functions, enabling the expression of time-extended preferences. At the same time, they can be structured in a way that enables effective learning from human behavior \citep{shah2018bayesian,zhou2022hierarchical}. To account for incommensurability and incompleteness, vector-valued reward functions \citep{vamplew2022scalar}, conditional preference networks \citep{boutilier2004cp-nets,cornelio2013updates}, or interval-valued utility functions \citep{denoeux2020interval} can be used, allowing our models to explicitly surface hard choices due to preferential gaps. Many of these representations are also associated with rich compositional semantics, making apparent the complex internal structure of human goals and preferences \citep{gerevini2005plan,davidson2024goals}. Although these formalisms have limitations of their own, they nonetheless embed important insights about how preferences can be computationally represented. As such, they deserve further study by alignment researchers seeking to adequately model human preferences in a general fashion, while also being useful representational tools for today's AI systems.

\subsection{Beyond preferences as representations of human values and reasons}

\partitle{Preferences are constructed, not basic.} Thus far, we have proceeded as if human motivations and values are adequately captured by the concept of ``preference'' as it is used in rational choice theory. But as far as evaluative concepts go, this concept of ``preference'' is an extremely thin one: Mathematically, a ``preference'' is just some ordering of two options, which can be interpreted as either a disposition to choose one option over another, subjective liking of one option over the other \citep{franklin2022recognising}, or an all-things-considered judgment in favor of one of the options. Distinct as these interpretations are, what they share is their highly abstract and general nature --- ``preference'' is a \emph{thin} concept because it does not encode richer semantic information beyond the bare notion of ``betterness''. Insofar as utility functions are interpreted as representations of preferences, this thinness is inherited by them: Utility just represents the mere preferability of some option. But \emph{why} exactly are some options preferred over others? In virtue of what reasons do people make these preference judgments? Without answering these questions, we are unlikely to model how someone's preferences generalize to novel options in ways they would endorse. To do so, we must go beyond preferences as the fundamental unit of analysis, and understand how preferences are \emph{computed} and \emph{constructed} from our reasons and values \citep{warren2011values,lichtenstein2006construction}.

\partitle{Rational choice as action on the basis of reasons.} In making this point, we depart from the domain of rational choice theory, and return to a more basic understanding of what it means to model ourselves as rational agents: We are agents that take ourselves to act on the basis of \emph{reasons} \citep{raz1999engaging,logins2022normative}.\footnote{While some psychological theories deny that reasons are the \emph{causes} or \emph{motivations} for human action (at least typically), they can nonetheless serve as \emph{justifications} for our actions \citep{mercier2011humans,mercier2017enigma}. As such, insofar as our goal is to build AI systems that infer \emph{justified bases of action} from our behavior (and then act according to them), reasons can still play this role.} These reasons might include desires, such as an intrinsic desire to avoid pain \citep{sinhababu2017humean}, evaluative judgments, such as the judgment that a movie is artistic enough to be worth watching \citep{anderson1995value}, or even acts of will, such as the intention to pursue a specific career \citep{chang2009voluntarist}.

\partitle{Evaluative concepts as building blocks for reasons.} What exactly is the content of these reasons? In decision theory and Humean accounts of motivation \citep{sinhababu2017humean}, only beliefs (represented as subjective probabilities) and desires (represented as the utility of some desired outcome) are considered as reasons for action. But even if we set aside other accounts \citep{anderson1995value,chang2004can,parfit2018rationality}, this leaves open what a person's beliefs and desires are \emph{about}. If I desire to be both helpful and honest to others, what does it mean to be helpful or honest?
Acting upon this desire requires applying the concepts of \emph{helpfulness} and \emph{honesty}, which are not just any concepts, but \emph{evaluative concepts}, or \emph{values}. Importantly, most such concepts are not thin ones, like \emph{preference}, \emph{utility} or \emph{goodness}; they are \emph{thick} evaluative concepts --- concepts that comprise both descriptive and normative elements --- such as \emph{beauty}, \emph{humor}, or \emph{health}. As \citet{blili2023making} point out, even the concept of \emph{intelligence} so central to AI is thick in this way.

\partitle{Utility functions as aggregators of distinct evaluative judgments.} How should AI systems model such evaluative concepts, and their relationship to preferences and action? As a first pass, one might turn the utility representation theorems on their head, viewing reward and utility functions as \emph{generators} of human preferences, instead of mere representations of them. Indeed, as gestured at earlier, reward and utility functions are often interpreted in this way, with rewards, costs, and utilities respectively treated as biological signals \citep{singh2009rewards}, energetic expenditure \citep{ab2020inverse}, or units of pleasure \citep{bentham1789introduction}. Preferences can then be treated as \emph{downstream comparisons} of these more basic quantities, as assumed in reinforcement learning from human feedback \citep{christiano2017deep,knox2024models,zhu2023principled}. Taking this line of thought further, one might treat evaluative concepts 
such as ``aesthetic quality'' or ``helpfulness'' as \emph{features} over which a reward or utility function is defined, reducing the problem of ``value learning'' to one of representation or feature learning \citep{barreto2017successor,bobu2022inducing,bobu2024aligning}. On this interpretation, reward and utility functions represent \emph{aggregate} evaluative judgments, with each feature corresponding to a distinct way of valuing the world.

\partitle{Utility functions assume that values are always commensurable.} Although there is much to be said in favor of this approach, we believe that it is not quite enough. For one, it is still subject to the representational limits of reward and utility functions. In particular, if utility functions are used to represent aggregate value judgments, this effectively assumes that distinct human values are \emph{always} commensurable in some way, and that our resulting preferences are always complete. Yet, as value pluralists argue, there are contexts where it seems hard or impossible to commensurate our values \citep{anderson1995value}, resulting in choices where our reasons run short, and we cannot say if one option is ultimately better than another \citep{chang1997incommensurability}.\footnote{See our immigration example from earlier, where it may be unclear how to prioritize between political freedom and material security when deciding whether to migrate.} Even when we do commensurate our values, utility functions do not provide further information on our reasons and justifications for those trade-offs.

\partitle{Evaluative judgments are not reducible to observable features.} For another, by conceiving of evaluative concepts as ``features'', we risk over-simplifying the semantics of many evaluative domains. Consider, for example, the concept of whether a research paper is \emph{novel}, or whether an action is \emph{helpful} or \emph{universalizable}. Applying these concepts requires a complex set of computations: \emph{novelty} involves evaluating the contributions of a paper with respect to a broader field of established knowledge \citep{amplayo2019evaluating}; \emph{helpfulness} involves estimating the goals of the agent being helped, and then judging whether the action aided in achieving that goal \mbox{\citep{ullman2009help}}; \emph{universalizability} involves simulating what would happen if everyone took a particular action \citep{levine2020logic,kwon2023not}. The structured nature of these concepts suggests the need for a suitably rich \emph{language of thought} --- one that captures the compositionality and algorithmic complexity of human conceptual cognition \citep{piantadosi2016four,quilty2023best,wong2023word}.

\partitle{Explicitly modeling processes of evaluation and commensuration.} To begin to capture all of this complexity, we propose that human decisions can be productively modeled as a three-stage process: Evaluate, Commensurate, then Decide (ECD).\footnote{Note that this a \emph{descriptive} framework for modeling how human reasons and values lead to decisions, not a \emph{prescriptive} framework for designing AI systems. We take up the latter topic in Section \ref{sec:rat_choice}.} Given some choice options, a set of \emph{evaluation procedures} compute valuations or rankings of the options under consideration, where each procedure corresponds to a distinct evaluative concept. These valuations serve as inputs to a  \emph{commensuration procedure} \citep{espeland1998commensuration}, which produces, where possible, a context-sensitive value assignment or preference ordering over the options (optionally with justifications for \emph{why} certain trade-offs were made), while leaving certain preferences unspecified when some options are not comparable. Finally, a \emph{decision procedure} computes actions and policies with respect to the (possibly incomplete) preference ordering induced by the evaluation and commensuration procedures, resulting in behavior that approximately satisfies those preferences.\footnote{One possible instantiation of this framework is multi-objective reinforcement learning \citep{vamplew2022scalar}: Each component of a vector-valued reward function can be thought of as a separate evaluation procedure. These can be transformed by the commensuration procedure into a lexicographic ordering (where some dimensions of value matter infinitely more than others) or constrained maximization problem (where some values must stay within a certain range while others are maximized). A planning or learning algorithm then serves as the decision procedure, producing an action policy that satisfies the commensurated preferences.} By explicitly modeling human decisions in this way, we can maintain the distinctness of the values that guide our actions, while foregrounding the ways in which we commensurate our values and dynamically construct our preferences.\footnote{In proposing this framework, we do not mean to imply that humans are \emph{always} going through these stages for every decision; as suggested by the RL formalism, one or more of these procedures may be cached through experience and learning, enabling habitual action without explicitly representing values in the brain \citep{keramati2016adaptive,hayden2021case}. Nonetheless, we can still \emph{rationalize} learned behavior and cached preferences in light of someone's values.} 

\partitle{Learning and specifying evaluative concepts.} This still leaves open the question of how evaluative concepts can be specified or learned. In principle, an AI system could infer such concepts from human decisions by inverting the ECD process, extending inverse reinforcement learning \citep{ziebart2008maximum} and Bayesian inverse planning \citep{baker2009action}. However, decisions alone might provide insufficient information about the nature and structure of our evaluative concepts. Recent advances in large language models (LLMs) suggest a promising alternative: By imitating the distribution of human text, LLMs appear to learn the conceptual roles associated with particular words \citep{piantadosi2022meaning}, and recognize semantic entailments between sentences \citep{merrill2024can}. Correspondingly, they might approximate the semantics of many evaluative concepts \citep{leshinskaya2023value}. This may explain why LLMs can often use evaluative adjectives in their appropriate contexts \citep{mahowald2023discerning}, and even perform rudimentary forms of moral reasoning \citep{jin2022make}. Still, LLMs remain limited in their ability to represent and reason with compositional concepts \citep{dziri2023faith,mahowald2024dissociating,ramesh2024compositional}, and would function as poor models of humans on their own. Instead, we could embed their approximate semantic knowledge into more structured models of human cognition \citep{kwon2023neuro,wong2023word} such as the ECD process described above. In doing so we might eventually model the full richness of human practical reasoning. 

\section{Beyond expected utility theory as a normative standard of rationality}
\label{sec:eut_rat}

\ifshowtables
\begin{table}[h]
\centering
\begin{adjustbox}{center, max width=\paperwidth}
\footnotesize
\begin{tabular}{p{3.5cm}p{4.5cm}p{4.5cm}}
\toprule
\textbf{Assumption} & \textbf{Limitations} & \textbf{Alternatives} \\ 
\midrule
EUT prescribes \& describes behavior of sufficiently rational or intelligent agents.
&
\begin{minipage}[t]{\linewidth}
\begin{itemize}[topsep=0pt,itemsep=0pt,leftmargin=0pt]
    \item Preference completeness is not a requirement of rationality.
    \item EU maximization is intractable, only weakly informative of actual AI.
\end{itemize}
\end{minipage}
&
\begin{minipage}[t]{\linewidth}
\begin{itemize}[topsep=0pt,itemsep=0pt,leftmargin=0pt]
    \item Mechanistic analyses of AI behavior, grounded in existing AI paradigms.
    \item Economic or evolutionary analyses.
    \item Resource rational analyses.
\end{itemize}
\end{minipage}
\\
\midrule
Globally coherent EU maximizers as a (necessary) design target for advanced AI.
&
\begin{minipage}[t]{\linewidth}
\begin{itemize}[topsep=0pt,itemsep=0pt,leftmargin=0pt]
    \item Global coherence is unfaithful to locally coherent human values.
    \item Global coherence at odds with tool-like boundedness and locality.
\end{itemize}
\end{minipage}
&
\begin{minipage}[t]{\linewidth}
\begin{itemize}[topsep=0pt,itemsep=0pt,leftmargin=0pt]
    \item Local coherence through AI systems with local / bounded scopes.
    \item Local coherence via locally (non-globally) complete preferences.
\end{itemize}
\end{minipage}
\\
\midrule
EUT as a complete theory of rationality and reasoning.
&
\begin{minipage}[t]{\linewidth}
\begin{itemize}[topsep=0pt,itemsep=0pt,leftmargin=0pt]
    \item Lacks an account of (normative) reasoning about preferences.
    \item Only constrains instrumental preferences, not ``intrinsic'' preferences.
\end{itemize}
\end{minipage}
&
\begin{minipage}[t]{\linewidth}
\begin{itemize}[topsep=0pt,itemsep=0pt,leftmargin=0pt]
    \item Theories of normative reasoning (argumentation \& deontic logics, etc.).
    \item Learning from human normative judgments, guided by theories.
\end{itemize}
\end{minipage}
\\
\bottomrule
\end{tabular}
\end{adjustbox}
\vspace{3pt}
\caption{Assumptions, limitations, and alternatives to expected utility theory (EUT) as a normative standard for rationality and reasoning.}
\label{tab:EUT}
\end{table}
\fi

In the previous section, we described how research in AI alignment often assumes approximate utility maximization as a \emph{descriptive} model of human behavior, then highlighted the shortcomings of this approach. However, this leaves open whether utility maximization is a desirable \emph{normative} standard for both human and machine behavior --- that is, whether agents ought to maximize the satisfaction of their preferences as a condition of ideal rationality, regardless of whether they actually do so.

\partitle{Coherence arguments for EUT.} There is a long history of debate regarding the validity of this normative standard. Arguments in favor of expected utility theory (EUT) include the utility representation theorems mentioned earlier \citep{samuelson1938note,savage1972foundations,bolker1967simultaneous,jeffrey1991logic,von_neumann1944theory}, which start from an axiomatization of what preferences count as rational, then demonstrate that any agent that acts in accordance with such preferences must act as if they are an expected utility maximizer.\footnote{In von Neumann and Morgenstern (VNM) theory, the four axioms are: \textit{completeness}, any two distributions over outcomes can be ranked by preference; \textit{transitivity}, if a (probabilistic) outcome A is preferred over outcome B, and outcome B over outcome C, then outcome A is preferred over outcome C; \textit{continuity}, preferences vary continuously with how probable an outcome is; and \textit{independence}, a preference between (probabilistic) outcomes A and B does not change when there is some fixed probability of getting some third outcome C whether or not one chooses A or B. Variants of these axioms are used in the Savage and Bolker-Jeffrey representation theorems, which extend VNM theory to allow for subjective probabilities.} In the AI alignment literature, these results are often treated as ``coherence theorems'' about the nature of rational agency, either by taking the rationality axioms for granted, or by providing arguments in defense of the axioms \citep{omohundro2007nature,yudkowsky2019coherent,demski2018complete}. For example, Dutch book arguments can be used to show that an agent's betting odds must obey certain axioms of probability theory in order to avoid exploitation by others \citep{vineberg2011dutch}, and money pump arguments can be used to show that an agent's preferences should be acyclic in order to avoid guaranteed losses \citep{gustafsson2022money}. 

\partitle{AI alignment as EU maximizer alignment.} In light of these arguments, AI alignment researchers have traditionally assumed that advanced AI systems will act as if they are expected utility (EU) maximizers \citep{omohundro2008basic,yudkowsky2016ai}. As a result, many have framed the challenge of aligning AI as the problem of aligning an EU maximizer, with various proposals focused on how to circumvent the dangers of utility maximization \citep{taylor2016quantilizers,armstrong2017low,turner2020avoiding}, or on accurately learning the correct utility function to maximize \citep{dewey2011learning,armstrong2019research}. After all, if advanced AI systems will inevitably comply with EUT, then our only hope for aligning such systems is to stay within its confines. Furthermore, if EU maximization is rationally required --- and if intelligence implies rationality --- then any sufficiently intelligent agent that acts on the basis of human values must eventually coherentize those values into a utility function.

\subsection{Beyond expected utility theory as an analytical lens}

\partitle{Coherence is not rationally required.} However, coherence arguments for expected utility theory are not as strong as the AI alignment literature has often presumed. The most extensive version of these arguments is given by \citet{gustafsson2022money}, who provides a money pump argument for preference completeness, and then uses completeness to derive arguments for transitivity, continuity, and independence. Yet, as \citet{thornley2023coherence} points out, the argument for completeness depends on particular assumptions about how agents are permitted to choose when offered a series of potentially exploitative trades, which can be avoided as long as agents do not accept offers that are less preferred than options they previously turned down.\footnote{Note that whereas \citet{gustafsson2022money} is focused on justifying the VNM axioms as requirements of rationality (in part by introducing and arguing for other principles of rationality, such as Decision-Tree Separability), \citet{thornley2023coherence} is focused on whether the VNM axioms will apply to advanced AI systems, and takes no position on whether they are rationally required. Here we go one step further, and suggest that arguments by \citet{thornley2023coherence} and \citet{petersen2023invulnerable} place strong pressure on Gustafsson's acceptance of rationality principles like Decision-Tree Separability, and hence the argument that the VNM axioms are rationally required.} \citet{petersen2023invulnerable} formalizes this counter-argument further, proposing a dynamic choice rule that ensures agents with incomplete preferences are invulnerable to money pumps.\footnote{Analogous arguments have made in defense of imprecise probabilities \citep{bradley2014should}, since they imply incomplete preferences. See also \citet{laibson2007safety} on how non-EU preferences are protected by competitive markets, and \citet{von2008evolution} on how non-EU preferences can be evolutionarily stable.} Indeed, it is accepted by many decision theorists that preference completeness is not a requirement of rationality; instead, all that is required is for an agent's preferences to be \emph{coherently extendible} \citep{steele2020decision}. In turn, this implies that rational agents need not be representable as EU maximizers.

\partitle{Coherent EU maximization is intractable.} But let us imagine that coherence arguments do go through after all. Even if this were the case, it is far from obvious that advanced intelligences would comply with the axioms of utility theory (or be incentivized to do so) in the face of computational and practical limitations. As \citet{bales2023will} argues, behaving as an expected utility maximizer can come with considerable costs, while only providing limited benefits. In fact, as we noted in Section \ref{sec:rat_choice}, most utility functions are \emph{computationally intractable} to coherently maximize: \citet{camara2022computationally} shows that while certain simple classes of utility functions allow for rational choice behavior to be computed in polynomial time, for a large class of other utility functions, agents cannot tractably compute choice behavior that complies with the rationality axioms, and must instead resort to approximately maximizing their utility function. Alternatively, agents may insist on complying with the rationality axioms, but give up on even approximate optimality with respect to their original utility functions. In other words, it is not always resource rational to maximize expected utility.

\partitle{Coherence alone is not informative.} Suppose we could set aside these tractability worries as well.\footnote{Perhaps because it is proven that P = NP, or because advanced AI systems will have such vast resources at their disposal that all relevant intractable problems will be solvable in practice.}  
Even so, it is unclear what information EUT provides us. As discussed by \citet{shah2018coherence}, \citet{ngo2019coherent}, and \citet{bales2023will}, many kinds of behavior can trivially be described in terms of utility maximization, including an ``agent'' that does nothing at all. This means that EUT alone does not say much about the kinds of goals that advanced AI systems are likely to pursue, or what they are likely to do in order to pursue them. While it is possible to draw some conclusions about utility maximizing agents  \citep{soares2015corrigibility,turner2021optimal,everitt2021agent,carroll2023characterizing}, further assumptions are typically needed (e.g. constraints on the space of utility functions) before one can obtain stronger analytical results. Moreover, many deployed AI systems cannot be fully analyzed by EUT, as they are highly approximate (e.g. deep reinforcement learning agents).

\partitle{Alternative analytical lenses to EUT.} What alternatives might one turn to instead to ground understanding, prediction, and alignment of advanced AI systems? Since many others have already addressed some version of these questions, we offer here a brief taxonomy of approaches.

\partitle{Mechanistic analyses.} The most common of such approaches are \emph{mechanistic analyses}, which reason about the likely properties of AI systems by assuming specific classes of training processes or algorithmic procedures. For example, reasoning about the training dynamics of deep (reinforcement) learning systems can suggest pathways to power-seeking or deceptive behavior \citep{ngo2022alignment,langosco2022goal,krakovna2023power}, or give us confidence that deceptive alignment is unlikely \citep{wheaton2023deceptive}. Similarly, knowledge of the workings of general-purpose algorithms, such as model-based search techniques or approximate Bayesian inference methods, can deliver us predictions or even provable guarantees regarding the risk or safety of an AI system \citep{yudkowsky2015kansi,bengio2023scientists,dalrymple2024guaranteed}.

\partitle{Economic and evolutionary analyses.} One downside of mechanistic analyses is that they are tied to particular hypotheses about how AI systems are likely to be built. Given uncertainty about which AI paradigms will ultimately reign dominant, we might want to abstract away from the details of any particular class of AI architectures. While this was the original appeal of EUT analyses, other approaches may hold more promise: \emph{economic analyses} and \emph{evolutionary analyses} can respectively ground predictions about the behavior and capabilities of AI systems in what is likely to be economically competitive, or what is likely to be evolutionary successful. For example, economic incentives could imply that AI services are more likely to proliferate than AI agents \citep{drexler2019reframing}, while evolutionary arguments can help us reason about whether increasingly capable AI systems are likely to displace human control over the economy \citep{hendrycks2023natural}.

\partitle{Resource-rational analyses.} Finally, it may be possible to analyze AI systems through the lens of \emph{computational tractability} and \emph{resource rationality}, applying ideas from the study of human cognition to understanding the potential capabilities and limits of artificial cognition \citep{van2008tractable,lieder2020resource}. For instance, AI safety via debate can theoretically solve PSPACE problems if optimal play is assumed\footnote{Note that achieving optimal play, formalized as finding a Nash equilibrium, is itself computationally intractable for most games.} \citep{irving2018ai}, while \citet{zhixuan2022what} cites intractability as a reason to avoid centralized AI planners as an alignment solution, and \citet{van2024reclaiming} provide an intractability argument against the possibility of human-like AI via imitation learning. By and large, however, resource rational analyses of AI systems appear to be neglected. It is thus a potentially fruitful avenue for better analyzing future AI systems --- one which retains many of the appealing features of expected utility theory, but adopts a more feasible normative standard.

\subsection{Beyond globally coherent agents as design targets}

If agents are neither rationally required nor practically required to act as if they are expected utility maximizers, this opens up the design space of (advanced) AI systems that we might hope to build and align. In particular, we have the option of building AI systems that do not comply with one or more of the axioms of expected utility theory --- systems that are not \emph{globally coherent} in the way that expected utility maximizers are required to be.

\partitle{Non-globally coherent AI may be more faithfully and safely aligned.} Why might this be desirable? There are two broad reasons. One reason is \emph{faithfulness}. As we discussed in Section \ref{sec:rat_choice}, many human preferences may be incomplete due to incommensurable values, and we might want AI systems to faithfully represent that preferential structure when making decisions \citep{eckersley2018impossibility}. Otherwise, such systems might reliably take actions that promote certain outcomes over others, even though we have yet to form a preference over which of those outcomes is better.\footnote{For example, AI systems that influence or manipulate humans into choosing particular career paths or societal structures because they are programmed to regard them as the best options, instead of respecting our initially incomplete preferences over careers or societal structures.} Another reason is \emph{safety} --- for a wide range of (time unbounded) utility functions, expected utility maximizers have been shown to seek power over their environment \citep{turner2021optimal}, and avoid being shut down by their creators \citep{soares2015corrigibility},\footnote{Provided that such utility maximizers are aware of the existence of a shutdown button.} suggesting that sufficiently capable utility maximizers will create considerable risks if their utility functions are not compatible with human safety \citep{carlsmith2022power}.

\partitle{AI tools as locally coherent agents.} A general class of AI systems that seem to largely satisfy faithfulness and safety are what we might intuitively think of as \emph{tools}. We use tools to perform tasks that are \emph{context-specific} --- the goals we use them for vary by context --- as well as \emph{local} --- we do not expect or want them to reliably affect the world beyond the contexts of their use. Insofar as these tools can be thought of as agents, they are at best \emph{locally coherent} ones. In this sense, they mimic the role-specific nature of human preferences. Just as people have differing goals and obligations depending on whether they are in the role of a parent or a worker \citep{anderson1995value}, tools take on the aims and constraints of their users, whether those involve classifying images or generating code. Within each context, we are typically willing to commensurate our values such that our preferences can be represented as a local utility function, even if we are unwilling to do so in general.

\partitle{Tool-like locality through local scope.} How can we build AI systems that function as tools? The answer, of course, is that we already have: Most AI systems that exist today are best thought of as tools. This is not due to any special care on our part as designers, but only because functioning as a tool is the default nature of rule-bound, computationally limited algorithms with no representation of their own existence in the world. Such algorithms execute a bounded amount of computation in response to some input, terminating when they find an answer or if time runs out. They exhibit no preference for altering the conditions of their termination, or for gaining control over more of their environment, because they cannot even represent the environment they exist in. In other words, such systems are \emph{local in scope}. This is the case even for systems that we might be tempted to call agents due to their long horizon reasoning abilities (e.g. classical planners, theorem provers) or relative autonomy (e.g. self-driving cars, robot vacuums). To the extent that such systems can be represented as utility maximizers, they can often be viewed as having local, time-bounded utility functions, which provide no incentive for continued operation beyond a certain time or resource bound \citep{dalrymple2022you}. Very plausibly, we could even build highly advanced, economically transformative AI systems by composing these bounded tools \citep{drexler2022open,dalrymple2024safeguarded}.

\partitle{Maintaining locality despite global scope.} Suppose, however, that some actors want to build advanced AI systems that are \emph{not} bounded in these ways. For example, many AI companies are keen to develop general purpose AI assistants, which follow human instructions in a wide range of domains and contexts, remain operational across contexts, and possess enough understanding of the wider world that they can represent both themselves and their users as entities in that world model. LLMs are increasingly being used in this way, and while their reasoning capabilities remain unreliable and limited \citep{valmeekam2023on,dziri2023faith,momennejad2024evaluating}, one might imagine augmenting or embedding them within systems with more coherent representations and reasoning abilities \citep{parisi2022talm,sumers2024cognitive}. Can we ensure that such systems continue to function as tools, despite their increasingly \emph{global} scope?

\partitle{Contextual reward functions are insufficient for locality.} We suggest that the answer may depend on whether such systems remain local in terms of the \emph{completeness} of their preferences, despite having global scope. What does it mean for preferences to be only locally complete? Consider one tempting but unsuccessful way to formalize this idea: We design our system to have a \emph{context-sensitive reward function} $R(s, c)$, where $s$ is the current state, and $c$ is the current context (e.g. an instruction or prompt given to a LLM-based assistant). The hope is that users will be able to set $c$ to whatever they like, and the system will change the task it optimizes for. Within the context $c$, the system exhibits locally coherent behavior, since its preferences are given by the reward function $R(\cdot, c)$. However, since our system has global scope, it also cares about rewards \emph{across} contexts: its utility function for a trajectory $\xi = ((s_1, c_1), ..., (s_T, c_T))$ is $U(\xi) = \sum_{t=1}^T R(s_t, c_t)$. This means that the system will have a \emph{context manipulation incentive}, i.e. an incentive to enter and remain within contexts that deliver more reward. For example, it might persuade or manipulate the user to give it instructions that are easier to satisfy.\footnote{This can be viewed as a generalization of the shutdown problem \citep{soares2015corrigibility}: Shutdown implies switching from a context that delivers some reward to a context which never delivers reward.} The reason for this is that the system's preferences are still globally complete --- they are represented by a global utility function, despite being context-sensitive.  

\partitle{Tool-like locality through local completeness.} How could locally complete preferences avoid these context-manipulating incentives? Following recent work by \citet{thornley2024shutdown} on circumventing the shutdown problem via incomplete preferences, we formulate local preference completeness as follows: Within each class of trajectories with a fixed schedule of $k$ contexts $(c_1, ..., c_k)$ that take effect at times $(1, t_1, ..., t_{k-1})$, there is a complete preference ordering over trajectories. Across these classes, trajectories are \emph{incomparable}, leading to preferential gaps \footnote{This construction builds upon the incomplete preference condition described in \citet{thornley2024shutdown} for building agents that are neither shutdown-seeking nor avoiding.}. Agents with such preferences would still optimize their behavior while within each context. At the same time, they would exhibit no reliable disposition towards being in some contexts more than others, or manipulating the schedule of contexts. At least in the sense we identified earlier, they would function as tools.

In making this proposal, we do not mean to imply that it is impossible to align or ensure the safety of globally coherent agents --- it may be possible to avoid pathological incentives by maintaining uncertainty over the utility function to maximize \citep{hadfield2016cooperative,hadfield2017off}, or by carefully balancing utilities across contexts \citep{armstrong2017indifference,holtman2019corrigibility}. We also do not claim that incompleteness is \emph{necessary} for tool-like AI --- if we coordinate to ensure that powerful AI systems always remain bounded and local in scope, then we may never need to explicitly engineer incompleteness. Indeed, it remains unclear how to perform such engineering at scale.\footnote{\citet{thornley2024towards} describes a reinforcement learning scheme, but it may not apply to context switching.} Nevertheless, if we want to build AI systems that safely respect our preferences and values, it makes sense to keep our options open, and look beyond the default theoretical assumption of globally coherent agents.

\subsection{Beyond preferences as the normative basis of action}

\partitle{EUT does not explain when our preferences are normatively acceptable.} Up to this point, we have primarily critiqued the normativity of expected utility theory on formal grounds, drawing upon arguments from decision theory and computational complexity theory. But an arguably deeper problem with EUT is that it fails to ground the normativity of our preferences. EUT is a theory of \emph{instrumental rationality} not \emph{value rationality}:\footnote{A distinction introduced by \citet{weber1978economy}.} It tells us how to choose our actions in order to satisfy our preferences, and imposes constraints on what those preferences can be, but it does not say anything further about where those preferences can or should come from. Yet, as we have elaborated in Section \ref{sec:rat_choice}, human preferences are not fundamental, but \emph{derivative} --- they derive from our values and reasons. EUT is thus woefully incomplete. It might tell us how to derive instrumental preferences from intrinsic ones,\footnote{In the sense that the expected utility of some state or action can be derived from the expected utility of the  states it allows us to achieve.} but it provides no guidance on many questions of great normative importance, such as why and how to value human and animal lives, whether and when it is permissible to give up equality for efficiency in a democracy, or how to judge the desirability and relevance of EUT itself.

\partitle{Normative judgments are increasingly automated.} Reasoning about these normative questions has traditionally been the purview of humans alone. Indeed, there are many reasons to preserve that state of affairs, lest we cede our moral and political autonomy entirely to machines \citep{van2019critiquing}. But even without replacing human autonomy over normative affairs, we are already building AI systems that automate normative judgments, assist us with normative reasoning, or operate under normative uncertainty. For example, machine learning methods are routinely used to moderate content that may be regarded as toxic and offensive \citep{gorwa2020algorithmic}, or to steer LLMs towards producing outputs that are less harmful \citep{bai2022training}. More ambitiously, AI writing assistants are being used to draft legal arguments by mimicking certain aspects of legal reasoning \citep{iu2023chatgpt,lohr2023ai}. If these trends continue, then increasing amounts of work will have to be done to ensure that AI systems produce normatively appropriate behavior. Humans will either have to do work upfront --- a difficult task, given the combinatorially large space of situations that increasingly autonomous systems might encounter --- or we will have to imbue AI systems with some semblance of normative reasoning.

\partitle{The need for theories of normative reasoning.} What options do we have for doing this? What would it look like to reason about the preferences and values one \emph{ought} to have? Given the complexity of these questions, one might hope to sidestep the need for a formal account like EUT entirely, and instead train AI systems to \emph{imitate} human normative reasoning. This is exemplified by the standard training objective of LLMs, which incentivizes replication of human-generated text. By training such systems on normative human judgments, one might hope that LLMs will learn the reasoning patterns that produce such judgments \citep{jiang2021can}. Recent methods such as Constitutional AI \citep{bai2022constitutional} take this idea one step further, bootstrapping an LLM's ability to approximate human normative judgments by generating self-critiques \citep{saunders2022self} and revisions, then finetuning the LLM on its own revisions. However, even strong LLMs currently struggle to reproduce human judgments on sufficiently nuanced normative questions \citep{jin2022make,kwon2023neuro}, and there are reasons to doubt whether LLMs can learn to reliably reason through either imitation \citep{van2024reclaiming,dziri2023faith} or self-critique \citep{stechly2023gpt,valmeekam2023can}. This unreliability suggests that we might want formal theories of normative reasoning after all. Without such theories, we would have no general way of evaluating whether an AI system reasons ``correctly'', beyond comparison to often fallible human judgments.\footnote{While formal theories of reasoning will ultimately have to be evaluated against human judgments themselves, they deliver systematicity and precision that many AI systems do not. Just as with mathematics, logic, and probability theory, formal reasoning systems can succinctly express what we would reflectively endorse, provided that we accept certain principles of reasoning as sound.} Perhaps imitation or self-critique will be enough for the majority of everyday situations, but if we want AI systems to address normative questions that are increasingly far afield from past human experience, the ability to validate or produce long chains of normative reasoning may be crucial for both system evaluation and scalable oversight.

\partitle{Computational theories of normative reasoning.} Thankfully, alignment researchers do not have to develop theories of normative reasoning from scratch. Across philosophy, AI, and legal computing, there have been numerous attempts to formalize the logic of argumentation, preferences, and duties, providing systems for reasoning about what we ought to endorse, prefer, or act upon. Abstract argumentation frameworks can be used to compute sets of acceptable arguments given a system of attack relations \citep{dung1995acceptability}. Preference logics can be used to express and deduce preferences for some propositions over others \citep{von1972logic,liu2011reasoning}. Deontic logics can be used reason about what norms must be complied with, and which norms are entailed by others \citep{von1951deontic}. Many extensions and combinations exist, including argumentation frameworks that allow for reasoning over preferences \citep{amgoud1998acceptability,modgil2009reasoning}, or reformulations of deontic logic using preference logic \citep{hansson1990preference,liu2011reasoning}. Uncertainty over normative arguments and conclusions can also be handled through weighted argumentation frameworks \citep{amgoud2017acceptability} and probabilistic logics \citep{ng1992probabilistic,de2003probabilistic}, allowing us to avoid over-extrapolation of our normative judgments and dogmatism about ``normative truths''. For the purposes of AI alignment, the work that remains to be done is not so much one of formalization, but \emph{integration}: How can these reasoning systems interface with or augment the standard formalisms of probability theory and decision theory? And how can they be combined with algorithms for machine learning and decision-making?

\partitle{Integrating normative reasoning with machine learning}. One relatively straightforward path to integration might be to use normative reasoning frameworks as synthetic data generators: Instead of directly training machine learning systems on human normative judgments, algorithms for normative reasoning could be used to produce sets of internally consistent arguments that can be derived from an initial set of human-provided judgments. Similar to deductive closure training for classical logic \citep{akyurek2024deductive}, machine learning systems (e.g. LLMs) could then be trained on the sets of derived judgments and arguments,\footnote{Note there might be \emph{multiple} sets of valid or defensible arguments, since an initial set of normative premises might be in conflict without decisively ruling each other out \citep{dung1995acceptability} Maintaining this multiplicity may be crucial to avoid normative dogmatism.} which would hopefully strengthen their ability to produce sound argumentative conclusions, while improving performance at distinguishing incompatible judgments and identifying self-consistent sets of normative claims. Normative reasoning frameworks could also be used to scaffold and validate the outputs of machine learned systems \citep{castagna2024can}, improving interpretability and correctness while still allowing the overall AI system to work with open-ended (e.g. language) inputs. Finally, one might hope to minimize the role of uninterpretable machine-learned systems altogether, using them primarily for the translation of inputs and outputs while performing most of the reasoning (normative or otherwise) via symbolic model-based algorithms \citep{wong2023word,kwon2023neuro}. On this route, the main challenge will be to integrate normative reasoning with frameworks for model-based inference and planning, such as probabilistic programming \citep{van2018introduction,cusumano2019gen}.

Considerable work needs to be done before we can design AI that reasons flexibly and generally about preferences and values.
Still, there exist many opportunities for research that are under-explored. By taking advantage of them, we might hope to build systems that handle the true normative complexity of the situations we are deploying them into.

\section{Beyond single-principal AI alignment as preference matching}
\label{sec:single_align}

\ifshowtables
\begin{table}[h]
\centering
\begin{adjustbox}{center, max width=\paperwidth}
\footnotesize
\begin{tabular}{p{3.5cm}p{4.5cm}p{4.5cm}}
\toprule
\textbf{Assumption} & \textbf{Limitations} & \textbf{Alternatives} \\ 
\midrule
Alignment by learning and optimizing a scalar, acontextual reward function. &
\begin{minipage}[t]{\linewidth}
\begin{itemize}[topsep=0pt,itemsep=0pt,leftmargin=0pt]
    \item Assumes that values are commensurable within and across contexts.
    \item Implies that ideal behavior in one context applies to other contexts.
\end{itemize}
\end{minipage}
&
\begin{minipage}[t]{\linewidth}
\begin{itemize}[topsep=0pt,itemsep=0pt,leftmargin=0pt]
    \item Limit scalar reward optimization to bounded, task-specific AI systems.
    \item Context-sensitive rewards for AI that cannot optimize across contexts.
\end{itemize}
\end{minipage}
\\
\midrule
Alignment with static and asocial representations of an individual's preferences. &
\begin{minipage}[t]{\linewidth}
\begin{itemize}[topsep=0pt,itemsep=0pt,leftmargin=0pt]
    \item Preferences change via learning, reflection, value transformation.
    \item Assumes preferences reflect individual welfare, not societal norms.
\end{itemize}
\end{minipage}
&
\begin{minipage}[t]{\linewidth}
\begin{itemize}[topsep=0pt,itemsep=0pt,leftmargin=0pt]
    \item Where possible, alignment with informed (post-reflective) preferences.
    \item Where appropriate, alignment with socially-dependent preferences.
\end{itemize}
\end{minipage}
\\
\midrule
An individual's preferences as the target of single-principal AI alignment.
&
\begin{minipage}[t]{\linewidth}
\begin{itemize}[topsep=0pt,itemsep=0pt,leftmargin=0pt]
    \item Underspecified behavior due to preference change, incompleteness.
    \item Individual's preferences may be normatively unacceptable.
\end{itemize}
\end{minipage}
&
\begin{minipage}[t]{\linewidth}
\begin{itemize}[topsep=0pt,itemsep=0pt,leftmargin=0pt]
     \item Alignment with task \& role-specific normative criteria.   \item Alignment with the normative ideal of a good assistant.
\end{itemize}
\end{minipage}
\\
\bottomrule
\end{tabular}
\end{adjustbox}
\vspace{3pt}
\caption{Assumptions, limitations, and alternatives to single-principal AI alignment by matching and optimizing an individual's preferences.}
\label{tab:single_align}
\end{table}
\fi

If rational choice theory is an inadequate description of human behavior and values, and expected utility theory is an unsatisfactory account of rational decision-making, what does this imply for the practice of AI alignment? Though there is growing awareness of the limits of these preferentist assumptions \citep{casper2023open,lambert2023entangled}, most applied methods for AI alignment continue to treat alignment as the problem of \emph{preference matching}: Given an AI system, the goal is to ensure that its behavior conforms with the preferences of a human user or developer. 

\partitle{Reward learning as alignment via preference matching.} At present, the most prominent of such methods is reinforcement learning from human feedback (RLHF). Similar to other reward learning methods such as inverse reinforcement learning \citep{ng2000algorithms}, RLHF learns an estimate of a user's presumed reward function --- a \emph{reward model} -- from a dataset of their stated preferences. The AI system is then trained to optimize the learned reward model, with the aim of producing behavior that better conforms to the user's preferences. Since the development of RLHF for classical control problems \citep{knox2011augmenting,griffith2013policy,akrour2014programming}, the method has been extended to train increasingly complex AI systems in increasingly open-ended domains, including deep neural networks for robotic control \citep{christiano2017deep} and large language models \citep{ouyang2022training,bai2022training}. This latter development has led to an explosion of interest in RLHF, given the unprecedented capabilities and general purpose nature of LLMs.

\partitle{Foundational limitations of reward learning.} For all its success, RLHF faces numerous technical challenges \citep{casper2023open}, ranging from issues with preference elicitation \citep{knox2024learning} and scalable oversight \citep{leike2018scalable} to over-optimization \citep{gao2023scaling,moskovitz2024confronting} and stable training \citep{hejna2024contrastive}. Our focus, however, is more foundational, and applies to not just RLHF but \emph{any} alignment method derived from reward learning:\footnote{This includes Direct Preference Optimization \citep{rafailov2024direct}, Contrastive Preference Learning \citep{hejna2024contrastive}, and Distributional Preference Learning \citep{siththaranjan2024distributional}.} By committing to a reward representation of human preferences or values, reward learning suffers from all the representational limits we discussed in Section \ref{sec:rat_choice}. Furthermore, by treating reward as something to be optimized, reward-based methods adopt EUT as a normative standard, with all the issues that Section \ref{sec:eut_rat} describes.

\partitle{The limited scope of reward learning and preference matching.} In this section, we discuss what it would require for AI alignment research to take these challenges seriously. Importantly, we do not claim that reward-based methods are never appropriate. Rather, we argue that reward-based alignment --- and preference matching more generally --- is only appropriate for AI systems with sufficiently \emph{local} uses and scopes. In other words, it is adequate for only the \emph{narrow} or \emph{minimalist} versions of the value alignment problem, where the values and norms at stake can be summarized as a reward function specific to the system's scope. For sufficiently \emph{ambitious} or \emph{maximalist} attempts at AI alignment,\footnote{The distinction between ``narrow'' and ``ambitious'' value learning is due to \citet{christiano2015ambitious}, while the analogous distinction between ``minimalist'' and ``maximalist'' value alignment is due to \citet{gabriel2020artificial}.} more is necessary: AI systems will have to learn how each person's preferences are dynamically constructed, and be aligned to the underlying values that generate those preferences. Furthermore, when preferences are incomplete, or conflict across time, they have to be aligned with normative ideals about how to assist in such situations. While versions of these points have been made before \citep{hadfield-menell2018incomplete,gabriel2020artificial,yao2023instructions}, we aim to make precise the connection between values, norms, and preferences, and to illustrate concrete possibilities.

\subsection{Beyond alignment with scalar and acontexual rewards}

Two aspects of reward functions are important for determining their role in the practice of AI alignment. The first is whether they are \emph{scalar}. As explained in Section \ref{sec:rat_choice}, this corresponds to the question of whether values are treated as fully commensurable, and whether the preferences they represent are complete. The second, often underappreciated aspect, is whether they are \emph{contextual}: Is the reward function understood to be a representation of context-specific preference judgments, or of an individual's overall preferences?

\partitle{Scalar rewards are only appropriate in narrow decision contexts.} Scalar rewards are generally inadequate, since (as elaborated in Section \ref{sec:rat_choice}) they assume away the possibility of incomplete human preferences. But as long as these rewards are also understood to be contextual, then reward-based alignment can be appropriate. In relatively narrow decision contexts without sharp practical or moral dilemmas, it is not unreasonable to assume that people are willing to commensurate their values \citep{anderson1995value}. In these contexts (e.g. buying groceries, travel planning, solving math homework) it is often clear to us how to weight different values against others (e.g. quality vs. cost, time vs. comfort, correctness vs. verbosity), leading to a complete preference ordering that it is representable by scalar reward. Learning a reward function is thus not inherently problematic. If this learned reward function is then optimized by a \emph{bounded} AI system --- the kind of local, tool-like system we discussed in Section \ref{sec:eut_rat} --- then the downsides are also limited. A poorly learned reward function may still result in negative outcomes \citep{zhuang2020consequences}, but the system will not reliably bring about unintended non-local effects.

\partitle{Models of context-specific preferences will not generalize across contexts.} By and large, this is the setting within which methods like RLHF are applied. Reward models are learned from human preferences, but these preferences typically represent context-specific \emph{goodness-of-a-kind} judgments like ``How well does this robot achieve its goal?'' \citep{christiano2017deep} or ``How well do these responses follow the provided instructions?'' \citep{ouyang2022training} While such judgments may implicitly aggregate a number of underlying values like ``harmlessness'' or ``helpfulness'' \citep{bai2022training}, they are not judgments of goodness \emph{simpliciter}, or of goodness for the user as a whole. This means that the resulting reward models are only useful for narrow alignment. They can serve as reasonable guides to in-context behavior, but are unlikely to generalize beyond that context \citep{lambert2023alignment}. In particular, such reward models do not represent human preferences \emph{across} contexts, over an extended period of time.

\partitle{Context-sensitive preference models as an intermediate solution.} What would it take to align an AI system that operates across contexts? One option is the use of context-sensitive reward functions \citep{pitis2024improving}, as described in Section~\ref{sec:eut_rat}. Though this approach runs the risk of context-manipulating incentives, it may well be adequate for sufficiently bounded systems. Similar to our ECD proposal in Section \ref{sec:rat_choice}, context-sensitivity could be achieved by \emph{per context commensuration} of multiple values, perhaps by learning \emph{separate} reward or preference models for each value \citep{wu2024fine,go2024compositional,xu2024perfect}, then aggregating their rewards with different weights depending on the downstream context. Context switches could then be triggered by users by selecting a desired ``mode'' \citep{edwards2023ai} or specifying a system prompt \citep{pitis2024improving}.

Still, all that the above amounts to is solving several instances of the narrow alignment problem, then stitching together the answers. If society is on a path towards more general AI systems --- say, the globally-scoped AI assistants we discussed in Section \ref{sec:eut_rat} --- then we will need more general solutions.

\subsection{Beyond alignment with static and asocial preferences}

How should one build an AI system that is aligned not to a user in a particular context, but to assist a person over an extended period of time? Addressing this challenge requires a significantly more ambitious solution to the value alignment problem --- one that not only avoids the pathologies of expected utility maximization across global scopes (cf. Section \ref{sec:eut_rat}), but also accounts for the dynamically and socially constructed nature of our preferences.\footnote{Of course, it is always an option to avoid taking up this challenge; there are many transformative uses of AI that do not involve globally scoped personal assistance. Nonetheless, if AI researchers do aim for something like this goal, they should be clear about what it requires.}

Most alignment methods do not adequately account for these aspects of preference construction. Instead, they assume that elicited preferences are \emph{static} --- they do not change over time --- and \emph{asocial} --- they are independent of other agent's preferences and societal norms. These are reasonable assumptions if AI systems are only interacting with users over relatively short timescales, and if such interactions can be decoupled from their wider social context. Unfortunately, neither of these assumptions is true in general.

\partitle{Preferences change via adaptation, drift, learning, reflection, or volition.} Contra the first assumption, preferences are \emph{dynamic}: They change, shift, and grow over time \citep{franklin2022recognising}. This is partly the result of context, as we have discussed, and partly a feature of human psychology: per Kahneman, our stated preferences about an experience can vary with the time of elicitation \citep{kahneman2005living}; per Sen and Nussbaum, our preferences adapt to the conditions of what is available to us \citep{sen1999commodities,nussbaum2001symposium}. More generally preference change is the result of being agents who learn about the world and ourselves as we grow \citep{loewenstein2003predicting}, and who reflect upon and reconsider what we value and desire. As we change our beliefs about what is true, what we find instrumentally valuable changes accordingly. As we discover what we experience as pleasant or unpleasant, what we consider to be intrinsically valuable may also change. We can also \emph{voluntarily} change our values \citep{ammann2023value}, perhaps by practicing an art form so that we may appreciate it better, or by adopting a new way of life \citep{chang2009voluntarist,paul2014transformative}.

\partitle{Alignment with informed preferences as a partial solution.} Can standard techniques for AI alignment be transplanted to the dynamic context? One modification is to assume that preference change is due only to people learning about their desires over time. In this model, there is still a true underlying preference structure, albeit one initially unknown to the human, and the AI system can just treat those preferences as the target of alignment \citep{chan2019assistive}. Similar modifications can be applied to the case of changing empirical beliefs: Instead of satisfying a person's revealed preferences, the AI system aims to satisfy what their preferences would be if they were more informed \citep{reddy2018you}. This idea might even be extended to encompass \emph{reflection} upon preferences and values \citep{cath2016reflective}: By modeling people as bounded reasoners \citep{zhi2020online,alanqary2021modeling}, and integrating such models with frameworks for normative reasoning, AI systems could infer what people \emph{would come to want}, if they thought harder about what they truly value. 

\partitle{The challenge of genuine value change.} However, alignment with informed preferences avoids the deeper normative questions raised by \emph{genuine} value change: How should an AI system assist someone whose informed preferences change over time due to drift, volition, or transformation? Or what if a person's preferences adapt in response to (potentially oppressive or addictive) environments \citep{sen1999commodities,nussbaum2001symposium}? Unlike preference change due to learning or reasoning, there is no sense in which the resulting preferences are more informed or ``rational'' than they were before. Perhaps AI systems could optimize for a person's \emph{current} preferences, but this risks shifting or manipulating their preferences in undesirable ways \citep{ashton2022problem,carroll2022estimating,carroll2024ai}. Avoiding such shifts would require delineating the kinds of value change that are legitimate or illegitimate \citep{ammann2023value}, but as \citet{carroll2024ai} discuss, it is not obvious how to do so. Alternatively, one might hope to aggregate preferences across the \emph{time-slices} that make up a person \citep{hedden2015reasons}, but this introduces difficult questions about how to weight past, present, and future time-slices \citep{paul2014transformative,pettigrew2019choosing}, while ignoring the practical unity that individuates a person as a \emph{person} \citep{korsgaard1989personal,schechtman2014staying}, not just a collection of consciousness moments. 

\partitle{Preferences are socially constructed.} We shall return to these normative questions shortly. Before doing so, let us consider the assumption that preferences are \emph{asocial}. In rational choice theory, preferences are typically understood to be an individual's comparative judgments about the outcomes that would be best for them and them alone. These self-regarding preferences are often treated as the target of AI alignment \citep{hadfield2016cooperative,russell2019human}. But of course, many of our preferences are not asocial in this way. Instead, they are \emph{interdependent} \citep{sobel2005interdependent}: formed not in isolation, but influenced by the preferences, values, and norms of our social and moral circles. Sometimes this influence is merely instrumental --- one might prefer to follow a social norm just because it is convenient to do so. But sometimes the influence is \emph{constitutive} --- as in a parent's concern for their child's well-being, or a feminist's desire to uphold a norm of equality. If we are to align an AI system with an individual, we will need some way of accounting for these influences.

\partitle{Recursive preference modeling as a partial solution.} As an intermediate solution to the challenge of socially constructed preferences, one might hope to align AI systems with \emph{recursive or interdependent preferences} --- preferences which depend on the preferences of others \citep{sobel2005interdependent}. Such preferences can be modeled with recursive utility functions, which assign weight to the posited utility functions of other agents \citep{kleiman2017learning,kim2018computational}, or more general models of preference interdependence \citep{yang2003modeling}. Preferences or ``rewards'' can also depend on social and moral norms \citep{bicchieri2005grammar,oldenburg2024learning}, reflecting how people predict and respond to the \emph{normative infrastructure} of their society \citep{hadfield-menell2018incomplete}.

Yet, by keeping preferences or utility functions as the target of alignment, recursive preference modeling still faces the many of the limitations we have surveyed. In particular, it still runs the risk of treating preferences as normatively basic, rather than the values and norms that generate those preferences. It also limits our ability to reason about such values and principles, and whether they are \emph{appropriately} influencing an individual's preferences. After all, many social norms and influences are oppressive or otherwise undesirable \citep{lukacs1972history,althusser2006ideology}, shaping preferences in ways we intuitively regard as contrary to an individual's best interests. In this sense, the problem of interdependent preferences is similar to the problem of dynamic preferences. In both cases, a range of preference orderings are at play, and without additional normative considerations, it is not clear which set of preferences an AI system should be aligned with \citep{carroll2024ai}.

\subsection{Beyond preferences as the target of alignment}

In light of the challenges introduced by contextual, dynamic, and interdependent preferences, it is difficult to see how they can serve as a coherent alignment target. This also follows from our discussion in Sections \ref{sec:rat_choice} and \ref{sec:eut_rat}: If preferences are neither psychologically nor normatively basic, then it is not clear what justifies their being the target of value learning and alignment.

\partitle{Alignment with role-specific normative criteria.} This basic point, of course, is not new: As many have long appreciated, identifying someone's welfare or best interests with their preferences runs into a thicket of philosophical issues \citep{sen1999commodities,nussbaum2001symposium}. Recognizing these issues, \citet{gabriel2020artificial} argues for an explicitly moral conception of alignment: ``the agent does what it morally ought to do, as defined by the individual
or society''.\footnote{\citet{gabriel2020artificial} uses ``values'' to describe this alignment target, though in a slightly narrower sense than ours. Whereas we have primarily used ``values'' to refer to evaluative concepts and judgments in general, Gabriel's use implicitly picks out the values that are normatively relevant to AI system behavior.} Others have proposed similar approaches, though they replace ``morally ought'' with what an agent or humanity as a whole would reflectively endorse, as in ideal observer theories \citep{firth1952ethical,brandt1955definition} or coherent extrapolated volition \citep{yudkowsky2004coherent}. However, it is far from clear how to operationalize these abstract principles. To make progress, we suggest a conception of single-principal alignment that is significantly more constrained: When an AI system only serves an individual in performing a particular task or role, it should be aligned with \emph{the normative ideals or criteria that are appropriate for that role}. For narrow systems, this requires task-specific determination of appropriate normative criteria. For general-purpose AI assistants, this implies alignment with the \emph{normative ideal of an assistant}, rather than alignment to an individual's preferences, or to human normativity writ large.

\partitle{Existing methods effectively align AI with role-specific norms.} Before discussing the case of general-purpose assistants, it is worth noting that many existing alignment methods effectively \emph{function} to align AI systems with task and role-specific norms, even though they are \emph{described} as methods for alignment with human preferences.\footnote{This preferentist focus is explicit in e.g. \citet{ouyang2022training}, who introduce an application of RLHF to LLMs that, in their words, ``aligns the behavior of GPT-3 to the stated preferences of a specific group of people.''} As discussed earlier, the pairwise judgments provided by human annotators in RLHF are typically not their preferences as end users, but instead context-specific \emph{goodness-of-a-kind} judgments. These judgments are provided in response to questions about whether an AI system's output complies with specific normative criteria --- for example, helpfulness, harmlessness, and truthfulness \citep{ouyang2022training, bai2022training}. As such, insofar as these judgments can be called preferences, they are \emph{derivative} of normative standards like harmlessness, not the alignment target themselves. Preferences merely serve as data so that machines can \emph{learn} some approximation of these standards. The typical language used to describe reward-learning methods like RLHF is thus misconceived: As used, they are not methods for alignment with any one human's preferences, or for recovering the ``true reward function'' in some person's head, \footnote{Supposing the concept of a ``human reward function'' is even coherent. See \citet{butlin2021ai} for a discussion.} but for aligning AI systems with contextually-appropriate normative criteria.

\partitle{Normative criteria for general purpose AI assistants.} What then are the normative criteria for general-purpose AI assistants --- those globally scoped AI systems for which questions of preference change and incompleteness seem the most pressing? While we cannot give a definitive answer --- indeed, as we shall discuss, we think this is something that society will have to collectively decide --- we suggest that progress can be made by reflecting on the \emph{normative ideal} of a good assistant.

How does this ideal address the issues with preference alignment that we have raised? Here are a number of suggestions: First, a good assistant does not presume certainty about a person's preferences and values \citep{hadfield2016cooperative}. This means maintaining an awareness of their own ignorance, while avoiding unwarranted extrapolation of preferences from one context to another, including Knightian uncertainty about how preferences extrapolate \citep{dalrymple2024guaranteed}. Second, a good assistant is aware that some choices are hard, and some options may seem incomparable \citep{chang1997incommensurability}. When helping someone with such a choice, the assistant does not pretend to know which option is better, or try to optimize that person's life; instead, the assistant respects their autonomy, and \emph{empowers} them to make the most informed choice possible \citep{du2020ave}, while ultimately remaining agnostic as to which choice is ``best''. Third, a good assistant understands and respects the values of the person they are assisting. This means recognizing that a person's preferences often derive from their values, which can take priority over their immediate requests and preferences \citep{london2024beneficent}. The assistant also enables those values to grow and change through normatively acceptable forms of exploration, reflection, volition, or even drift, while avoiding manipulating them or restricting them \citep{ammann2023value}. Finally, a good assistant, being situated in wider society, respects the preferences and values of \emph{others} \citep{kirk2024benefits}. When assisting someone who wishes harm out of anger, the assistant might dissuade them from acting against their better nature. When asked to directly harm others, the assistant might refuse.

\partitle{Pathways to aligning general purpose assistants.} In a past era of AI development, these principles might have seemed too vague to formalize or implement. Yet, as our discussion of RLHF suggests, it now seems like we have at least one path towards aligning globally-scoped AI assistants: Train them to comply with human judgments and standards for ideal assistive behavior. Methods such as harmless and helpful RLHF \citep{bai2022training}, (collective) constitutional AI \citep{bai2022constitutional,huang2024collective}, and moral graph elicitation \citep{klingefjord2024human} are already taking steps in this direction, each of them making more explicit that the targets of alignment are not preferences, but normative principles for assistance. Such systems still have to learn the preference of each user they assist, but this is separate from learning \emph{how} to provide assistance in light of those preferences.

Within this broad approach, we can embed many of the proposals we have made in earlier sections. Rich but structured models of human decision-making can serve as the AI assistant's ``theory-of-mind'', producing well-calibrated estimates of user goals and preferences while avoiding the deficiencies of unstructured approaches \citep{zhi2024pragmatic,kim2023fantom}. Mechanisms for preference incompleteness could be engineered or trained into the AI assistant if this turns out to remove incentives for shutdown avoidance and context manipulation \citep{thornley2024shutdown}. Theories of normative reasoning could be integrated into AI systems, allowing them to reason about human-provided judgments and principles, while aiding us in deliberating about what counts as good assistance. Each of these proposals may turn out to be strictly unnecessary for the task. Even so, they can provide us helpful guidance as we refine and implement our normative ideals of assistance.

\section{Beyond multi-principal AI alignment as preference aggregation}
\label{sec:multi_align}

\ifshowtables
\begin{table}[h]
\centering
\begin{adjustbox}{center, max width=\paperwidth}
\footnotesize
\begin{tabular}{p{3.5cm}p{4.5cm}p{4.5cm}}
\toprule
\textbf{Assumption} & \textbf{Limitations} & \textbf{Alternatives} \\ 
\midrule
Naively utilitarian aggregation (i.e. equal weighting) of elicited preference data. &
\begin{minipage}[t]{\linewidth}
\begin{itemize}[topsep=0pt,itemsep=0pt,leftmargin=0pt]
    \item Conflates task-specific preferences with overall/welfare preferences.
    \item Exclusionary majority preferences can cause harmful/unjust outcomes.
\end{itemize}
\end{minipage}
&
\begin{minipage}[t]{\linewidth}
\begin{itemize}[topsep=0pt,itemsep=0pt,leftmargin=0pt]
    \item Task or role-specific aggregation of normative judgments.
    \item Prioritarian, egalitarian, or contractualist elicitation and aggregation.
\end{itemize}
\end{minipage}
\\
\midrule
Aggregate human preferen-ces as the target of multi-principal AI alignment. &
\begin{minipage}[t]{\linewidth}
\begin{itemize}[topsep=0pt,itemsep=0pt,leftmargin=0pt]
    \item Computationally intractable due to the difficulty of central planning.
    \item Politically infeasible given the divergent incentives of AI developers.
    \item At odds with the plurality of AI uses and human interests.
\end{itemize}
\end{minipage}
&
\begin{minipage}[t]{\linewidth}
\begin{itemize}[topsep=0pt,itemsep=0pt,leftmargin=0pt]
    \item Alignment with a plurality of norms for a plurality of AI systems.
    \item Norms as a strategically viable solution given our divergent interests.
    \item Mutual agreement as the normative basis for norm-oriented alignment.
\end{itemize}
\end{minipage}
\\
\bottomrule
\end{tabular}
\end{adjustbox}
\vspace{3pt}
\caption{Assumptions, limitations, and alternatives to multi-principal AI alignment by matching and optimizing aggregate human preferences.}
\label{tab:multi_align}
\vspace{-12pt}
\end{table}
\fi

Having argued against a preference-based conception of single-principal alignment, we now turn to the problem of multi-principal alignment: Given the multitude of humans that we share this planet with, and the plurality of values that we hold, what, if anything, should AI systems be aligned to? At least at first glance, it does not seem as though our assistive account of AI alignment can readily be extended to this context. What it means to assist a single person is relatively clear. What it means to assist multiple people --- especially people with conflicting values --- is far less obvious.

\partitle{A theoretical argument for preference aggregation.} A traditional answer to this question is that AI systems should be aligned to the \emph{aggregate} preferences of humans. Why so? Part of this may be the normative appeal of a preference utilitarian ethic \citep{hare1981moral}. In the AI alignment literature, however, the argument for preference aggregation is usually more technical \citep{critch2017servant,demski2018complete}, appealing to Harsanyi's social aggregation theorem as justification \citep{harsanyi1955cardinal}. Suppose we require that the AI system complies with the (VNM) axioms of expected utility theory. Suppose further that all humans also do so, such that the preferences of each individual $i$ can be represented by a utility function $U_i(x)$ over outcomes $x$.\footnote{Harsanyi's theorem also requires that all humans have \emph{common beliefs} \citep{critch2017servant}.} Finally, assume \emph{unanimity} as a minimal requirement of rational social choice --- if \emph{all} humans prefer some (probabilistic) outcome $x$ over outcome $y$, then the AI system should prefer $x$ over $y$ as well. Then Harsanyi's theorem says that the AI system's utility function $U(x)$ \emph{must} be a weighted aggregate of individual utility functions:
$$U(x) = w_1 U_1(x) + w_2 U_2(x) + \dots + w_n U_n(x)$$
where the weights $w_i$ are fixed values independent of the outcome $x$. By a veil-of-ignorance argument, Harsanyi also proposed that these weights should be \emph{equal}, reasoning that a risk-neutral decision-maker should assign equal probability as to which person they could become \citep{harsanyi1975can}.

\partitle{Preference aggregation in the practice of alignment.} However convincing one finds this theoretical argument, preference aggregation is often found in the practice of AI alignment as well. A notable example is, once again, RLHF: Despite having been originally designed for single-human contexts, in practice, RLHF is almost always applied to preference datasets collected from \emph{multiple} human labelers \citep{christiano2017deep,ouyang2022training, bai2022training}. This practice has recently been shown equivalent to the Borda count voting rule \citep{siththaranjan2024distributional}; in effect, each labeler's choices are weighted according to their \emph{ordinal} ranking among the set of possible alternatives.

\partitle{Practical, political, and foundational limits to preference aggregation.} In this section, we critically examine preference aggregation in AI alignment at the practical, political, and foundational levels. At the practical level, we contend that preference aggregation is often misinterpreted and misapplied, such that even if one accepts Harsanyi-style utility aggregation as a normative ideal, it may often be better to use various non-utilitarian aggregation rules in practice. At the political level, we critique the idealized nature of aggregationist approaches, arguing that approaches grounded in bargaining and social contract theory are more politically tractable given our diverse and contested values.
At the foundational level, we build upon our arguments against EUT and preference matching from the earlier sections, elaborating them into a critique of the normativity of utilitarian aggregation.

\subsection{Beyond naïve utilitarian aggregation of elicited preferences}

\partitle{Different types of preferences are subject to aggregation.} Discussion of preference aggregation and its uses is often afflicted by confusion about the nature of preferences. Are these all-things-considered preferences, or goodness-of-a-kind judgments? Are these preferences over outcomes \citep{harsanyi1953cardinal}, or preferences over ethical views \citep{baum2020social}? Are these self-regarding preferences, social preferences, or some combination of the two? For clarity, we shall use the term \emph{welfare preferences} \citep{rubinstein2012eliciting} to refer to those preferences that Harsanyi's theorem most intuitively applies to: These are self-regarding preferences over outcomes that affect one's individual welfare, which exclude consideration of others' welfare. We distinguish this concept from \emph{all-things-considered preferences}, which are preferences about overall goodness (including social or moral considerations), and from \emph{elicited preferences}, which refers to any kind of preference elicited while applying some alignment technique.

\partitle{Aggregation of elicited preferences need not track aggregate welfare or goodness.} The first thing to note is that elicited preferences, welfare preferences, and all-things-considered preferences may all come apart. This crucially affects why and how we aggregate preferences, and whether some utilitarian aggregation procedure should be used. Consider a hypothetical example in the context of RLHF: Users are asked whether they would personally enjoy an LLM that can generate copyrighted short stories, and most of them say yes. If what we care about is aggregate (immediate) welfare, then uniform aggregation of the elicited preferences seems to achieve that goal. But if we what we care about aggregating are all-things-considered value judgments --- including legal and moral considerations --- then uniform aggregation no longer seems so appropriate. 

Similar issues arise when trying to aggregate \emph{toxicity} or \emph{harmfulness} judgments across multiple humans \citep{bai2022training,davani2022dealing}. In these cases, the elicited preferences are goodness-of-a-kind judgments, and their connection to aggregate welfare (or all-things-considered goodness) is many steps removed. As such, uniform or majoritarian aggregation can easily fail to achieve social goals. If most human annotators are insensitive to certain forms of identity discrimination (e.g. sexually demeaning images, trans-exclusionary rhetoric, or anti-semitic tropes), then AI systems trained on such data will almost certainly cause harm \citep{richardson2019dirty,okidegbe2021discredited}. Uniform preference aggregation may thus constitute a form of epistemic injustice \citep{fricker2007epistemic,symons2022epistemic,hull2023dirty}, which in turn leads to downstream injustice and harm.

\partitle{Non-utilitarian aggregation may be beneficial on normative or epistemic grounds.} What aggregation procedures might we use instead? And what justifies their use? In the case of potential copyright violations, we might want to grant veto power to copyright holders, allowing them to \emph{reasonably reject} the welfare-oriented majority preference for copying their work. This veto right could be justified as an instantiation of Scanlon's contractualism \citep{scanlon2000what}, on the principle that mutual respect among persons necessitates taking claims of intellectual ownership seriously. Alternatively, it could simply be understood as a policy that everyone would reflectively prefer, once they properly understood the costs and benefits of a copyright veto.

As for harmfulness judgments, it may often be preferable to apply \emph{prioritarian} \citep{lumer2005prioritarian,holtug2017prioritarianism} or \emph{egalitarian} \citep{rawls1971theory} approaches to aggregation. For example, one might select annotators who are \emph{most directly impacted} by potential harms \citep{gordon2022jury}, thereby prioritizing certain segments of the population. In cases of significant disagreement, one might even place all weight on the individual with the strongest dispreference \citep{leben2017rawlsian,bakker2022fine-tuning,weidinger2023using}. Again, there are many possible justifications for such procedures. Prioritarian selection could be justified on normative grounds, or because of its epistemic benefits --- after all, those most impacted by harms also tend to be \emph{more informed} about their effects \citep{dror2023there}. 

\partitle{Distinguishing aggregation procedures from standards of rightness.} Whatever procedure one favors, it is important not to confuse the aggregation rules used in AI systems with our ultimate social objectives. In practice, these aggregation rules are merely parts of the overall decision procedure implemented by (training) an AI system, and as many philosophers have pointed out, such procedures should be distinguished from standards of rightness \citep{railton1993alienation,frazier1994act,stark1997decision}. Rather than directly instantiating a particular standard (or its mathematical formalization) into a preference aggregation procedure, we should consider which aggregation procedures best satisfy the standard(s) we care about, taking into account practical and informational constraints. In doing so, we should recognize that elicited preferences are typically not the objects of our concern, but simply \emph{information} as to what we truly care about.

\subsection{Beyond aggregate preferences as the target of alignment}

Suppose we recognize that any particular set of elicited preferences is merely a guide or estimate to what we care about. Even so, one could still imagine taking humanity's aggregate preferences as the \emph{target} of AI alignment. For example, suppose that humanity eventually builds a single powerful AI system --- a singleton --- that actively infers the preferences of all humans, uses those preferences to estimate humanity's social welfare function, then optimizes its best estimates of that function. In doing so, we might create the ideal utilitarian central planner, achieving what welfare economists and utopian socialists could only dream of \citep{ng1997case,bastani2019fully}.

\partitle{Theoretical difficulties for preference aggregation.} Unfortunately, taking aggregate preferences as an alignment target immediately runs into theoretical difficulties. While these issues have been studied at length by social choice theorists,\footnote{See \citet{baum2020social,korinek2022aligned,mishra2023ai} and \citet{conitzer2024position} for discussions of the challenge of applying social choice to AI alignment.} one that is especially challenging for standard utilitarian aggregation is incomparability. As we noted earlier, justifications for preference aggregation typically assume that each individual's preferences can be represented as a utility function, and furthermore that utility can be compared across persons \citep{harsanyi1953cardinal,harsanyi1975can}. But as we have elaborated Section \ref{sec:rat_choice}, these assumptions are very much in doubt. Even within a single individual, preferences may be incomplete due to incomparable choices, or not clearly comparable across time \citep{carroll2024ai}. Having to compare the goodness of choices across individuals only makes the difficulty more severe \citep{korinek2022aligned}. This is not to say that the preferability of some outcome can \emph{never} be compared across people,\footnote{For example, if the choice of person A not wearing a mask would lead to less inconvenience for person A but severe illness for person B, we should intuitively give a stronger weight to person B's preference against severe illness over person A's preference against inconvenience.} but that any such comparison stands in need of further normative justification \citep{sen1970collective,clayton1999egalitarian} --- justification that, as we argued in Section \ref{sec:eut_rat}, utility theory alone cannot provide.

\partitle{The computational intractability of aggregate preference optimization.} Let us suppose, however, that these theoretical challenges can be addressed.\footnote{Perhaps by using frameworks that allow for partial comparability of welfare across individuals \citep{sen1970interpersonal}, or by aligning AI with partial social preferences \citep{korinek2022aligned}.} Even so, aggregate preference optimization still faces serious practical challenges. For one, such optimization is \emph{computationally intractable}: As Austrian economists have long argued, central planning runs into the economic calculation problem \citep{von1990economic}, a problem made worse by the sheer complexity of inferring human preferences under limited information, coordinating global production to maximize aggregate preferences, and planning for the future under uncertainty \citep{hayek1945use,murphy2006cantor,cwik2024revisiting}.\footnote{These difficulties can be formalized with the aid of theoretical computer science, which shows that optimal planning under uncertainty is sometimes undecidable, and even when decidable, remains anywhere from PSPACE to EXPTIME-complete \citep{papadimitriou1987complexity,chatterjee2016decidable}.} In contrast, decentralized decision-making (in the form of e.g. competitive markets) can sometimes be exponentially more efficient in computational cost than central planning \citep{rust1996dealing}, while achieving optimal informational efficiency \citep{mount1974informational,jordan1982competitive}. As such, even if not a practical impossibility, optimizing humanity's aggregate preferences with a single AI system is likely to be considerably less efficient than more pluralistic alternatives \citep{siddarth2022how}.

\partitle{The politically infeasibility of impartially benevolent AI.} Perhaps even more importantly, the project of building AI that optimizes humanity's aggregate preferences is \emph{politically infeasible}: Even if impartially benevolent AI planners were possible to develop, building such systems would be incompatible with the incentives of every AI developer with a realistic chance of doing so. This is the case even for AI developers with expressedly pro-social missions, which are still subject to market incentives as a result of the need to raise capital \citep{toner2024aifirms}, and are still governed by the laws and regulations of the countries they are based in.
Allowing the creation of such AI systems would also risk the centralization of immense power: However virtuous the goal of impartial preference optimization might seem, the history of central planning should tell us that optimal social outcomes are far from likely to be achieved \citep{scott1998seeing,verdery2005socialism}. Instead, we are more likely to see a tyranny of creator values, with potentially disastrous consequences for everyone with a contrary way of life.

\partitle{Pluralistic alignment as a politically feasible alternative.} In light of these challenges, how should we reconceive the goals of multi-principal AI alignment? One constraint in doing so is incentive compatibility: Whatever our vision of AI alignment is, it should account for divergent interests and contested values, credibly enabling collective safety and stability by ensuring incentives for cooperation and minimizing the chances of conflict \citep{critch2020ai,dafoe2020open}. A related constraint is political feasibility: Alternative targets for alignment should be achievable given the political economy of actually existing AI --- an economy that consists of a wide variety of AI services developed and deployed by a large number of self-interested actors \citep{drexler2019reframing}. Although these are negative constraints, they pair well with a more positive, pluralistic vision of what alignment could enable: A world where increasingly advanced AI systems serve a diversity of individual, communal, and universal ends, without catastrophically endangering anyone’s interests \citep{zhixuan2022what,gabriel2020artificial,siddarth2023whitepaper}.

\partitle{Enabling pluralism through political constraints.} What would it require to enact this pluralistic vision? As a starting point, consider our principles for AI assistance from Section \ref{sec:single_align}. While an AI assistant primarily serves a single person, and might be personalized to do so in many ways \citep{sorensen2024roadmap}, our presumptive norms for ideal assistance do not permit disregard for others. Rather, they endorse a \emph{circumscribed} promotion of the person's interests and values, such that the assistant avoids harming and endangering other individuals.\footnote{This might be viewed as an instantiation of the Harm Principle \citep{mill1859liberty} for AI assistants.} These norms function as political constraints, allowing assistants to provide value for individual users without imposing unreasonable externalities upon others.\footnote{See also \citet{kirk2024benefits} on the bounds of personalization in LLM assistants, and \citet{gabriel2024matter} for an explicitly political conception of AI alignment.} In doing so, they reduce the chance of conflict and non-cooperation.

\partitle{Alignment with politically negotiated normative standards.} Our suggestion then, is that this approach can be generalized to broadly \emph{contractualist} account of AI alignment \citep{zhixuan2022what}:\footnote{We use ``contractualist'' here in a broad sense, which includes both contractarianism \citep{cudd2021contractarianism} and Rawlsian \citep{rawls1971theory} or Scanlonian contractualism \citep{scanlon2000what}.} Rather than learning humanity's preferences in order to maximally satisfy them, AI systems should be aligned with normative standards and criteria that we collectively forge and negotiate --- standards exemplified by social, legal, and moral norms. These norms may be constructed as we design each system, or can be decided in advance for entire classes of AI systems. Returning to our earlier discussion of role-specific alignment, what is important is that these norms are tailored to the scope and uses of each system: Just as AI assistants should avoid harmful language, self-driving cars should follow the rules of the road. By negotiating norms and constraints for each of AI's social functions, we can enable a plurality of uses for AI while limiting the costs and harms to all stakeholders involved.

\partitle{The practical benefits of contractualist alignment.} What benefits does a contractualist approach to alignment offer? In our view, its primary benefits are practical ones: Unlike aggregate preference optimization, contractualist alignment does not require unrealistic amounts of benevolence from any one actor. Instead, it aims for a regime where largely self-interested actors stand to mutually benefit from the development and deployment of AI. Well-designed norms and institutions enable this, stabilizing cooperation by making it costly for relevant parties to defect or withdraw from cooperation \citep{kalai1975other,gintis2010social}. Aligning AI systems to comply with cooperative norms (and perhaps even to enforce them) thus reduces the chance of AI-caused or mediated conflict, or the risk of (catastrophically) endangering anyone's interests. Norms also limit the computational and informational cost of ensuring aligned behavior: Rather than inferring a large number of preferences, norm-aligned agents just have to (learn to) comply with a limited set of constraints \citep{oldenburg2024learning}. Finally, by centering norms and principles as the targets of AI alignment, political deliberation becomes more feasible and widely accessible \citep{huang2024collective}: Stakeholders need not negotiate over every last detail over how an AI system is built or trained, but can instead agree upon high-level requirements and standards for how the system should behave.\footnote{This does not preclude lower-level forms feedback such as participatory data labeling \citep{gordon2022jury} or end-user audits \citep{lam2022end}, which can complement the aim of mutually-acceptable AI design.}

\partitle{The normative grounds of contractualist alignment.} Besides its practical benefits, contractualist alignment can also be grounded in normative foundations that are more compatible with a pluralistic world. While it might be possible to justify broadly contractualist principle-setting on rule consequentialist grounds \citep{parfit2011matters}, we contend that the normative appeal of contractualist alignment is precisely that it avoids a universal account of what consequences are better or worse.\footnote{Similar arguments are made by \citet{gabriel2020artificial} and \citet{gabriel2024matter}.} Given the difficulties with comparability that we have examined, it is unlikely that people will ever agree upon a single scale of value for ranking all consequences. Instead, contractualist alignment aims to align AI systems with goals, standards, and principles that are mutually agreed upon by people despite our disparate preferences and values, deriving its normative force from the fair and impartial agreement of relevantly-situated rational actors.

\partitle{Conditions for fair and impartial agreement.} What makes an agreement impartial or fair? As in contractarian moral and political theories \citep{gauthier1986morals,binmore1994game}, it may be enough that all stakeholders benefit relative to an originally fair bargaining position, subject to additional symmetry constraints. Or as \citet{rawls1993political} and \citet{scanlon2000what} respectively argue, a thicker conception of public reason and the mutual recognition of each other as reasonable persons may be necessary to decide which agreements are fair. While examining these questions would take us beyond the scope of this paper, we believe our critique of expected utility theory lends itself to thicker conceptions of fair and reasonable agreement. On such conceptions, AI systems should not just be aligned with goals and standards that achieve mutual benefit.\footnote{After all, mutual benefit is not always achievable. In such cases, it is still possible to reach agreements that are viewed as fair, as in an agreement to compensate someone for harm.} Instead, AI goals and standards should be \emph{justified} to each stakeholder, on grounds that none can reasonably reject. Insofar as these AI systems are used to exercise power over others, they should also act in accordance with standards that are not just fair, but \emph{legitimate} \citep{lazar2024legitimacy,stone2024legitimate}.

\partitle{Alignment in the absence of agreement.} A natural worry for contractualist alignment is the possibility that agreement between different stakeholders may not be obtained (let alone agreement that is impartial and fair). Yet, this worry is not as acute as it may initially seem. First, rather than aligning AI systems with norms that have \emph{actually} been agreed upon, we could align them with norms that would \emph{hypothetically} be agreed upon, in the spirit of virtual bargaining \citep{misyak2014unwritten,chater2023could}. This would generally be necessary to handle incompletely specified agreements and contracts \citep{hadfield-menell2018incomplete}, while sharply lowering the cost and frequency of actual negotiations. Second, there are many cases where the operation of an AI system imposes minimal externalities upon others, and hence the cost of disagreement between AI stakeholders is merely that the gains of cooperation cannot be realized. In such cases, it is no great loss if each party operates their own AI system aligned with their individual goals, rather than having a shared AI system aligned with collective goals and norms. It is only when AI systems do impose substantial negative externalities that disagreement about their operation is more dangerous. These situations could well lead to mutually destructive conflict, as in prisoner's dilemma scenarios, or exploitative outcomes, where some AI operators benefit significantly at the expense of others. Even so, humans still have the political agency to shape which agreements are feasible and fair, and there is reason to hope that parties will negotiate to avoid at least the worst AI outcomes (e.g. in the form of minimal safety standards). Finally, achieving agreement over norms and principles is likely to be far easier than agreeing on a metric for globally ranking all consequences or comparing all people's preferences. As such, unless one is willing to allow a small set of actors to decide how all of humanity's preferences should be weighted and compared, utilitarian preference aggregation faces an even sharper risk of disagreement and conflict than a contractualist approach.

\partitle{Technical avenues toward contractualist alignment.} If we accept this contractualist understanding of multi-principal alignment, then much work remains to be done. On the technical front, there need to be advances in the theory and implementation of cooperative or contractualist decision-making. While recent alignment techniques show how language-based AI assistants can be aligned with collectively elicited norms and values, and how divergences in norms, opinions, and values can be reconciled through agreement \citep{huang2024collective}, iterative critique \citep{bakker2022fine-tuning}, or moral reflection \citep{klingefjord2024human}, these methods are specialized to a particular type of AI system, and have yet to be situated in a more general theoretical framework. To develop such a framework, we suspect that it will be necessary to unite ideas from game theory \citep{dafoe2020open}, bargaining theory \citep{chater2023could}, and social choice \citep{conitzer2024position} with formal approaches to argumentation \citep{amgoud1998acceptability} and negotiation \citep{rahwan2003argumentation}, along with insights from the science of human normativity \citep{binmore1994game,hadfield-menell2018incomplete,levine2023resource}. In particular, by developing computational theories of how humans rapidly learn extant norms and conventions \citep{tan2019bayesian,hadfield2019legible,hawkins2019emergence}, recognize institutional structure \citep{jara2024institutional,baker2024roles}, and engage in contractualist reasoning about social and moral norms \citep{levine2023resource,levine2024rules}, we can inform the design of AI systems with social and normative competence: AI that is not just aligned with stakeholder values in a once-off process, but which flexibly adapts to our norms and institutions as they evolve \citep{oldenburg2024learning}, reasons about their applicability in novel situations \citep{kwon2023not}, and perhaps even aids us in negotiating new contracts and norms \citep{christoffersen2023get,jarrett2023language,tessler2024ai}.

\partitle{Social and political avenues toward contractualist alignment.} Of course, if we take fair and impartially negotiated standards as the target of AI alignment, then technical advances will not be enough; we also need to foster the development of social, economic, and political orders that provide the conditions for free and fair agreement. This might involve the creation of new economic and political mechanisms that elicit and consolidate the interests of AI stakeholders \citep{siddarth2023whitepaper}, the establishment of democratic processes and bodies that can exercise legitimate authority over AI systems \citep{ovadya2023reimagining}, or the expansion of participatory approaches to AI development and design \citep{birhane2022power,suresh2024participation}. Without these social and political investments, we will lack the capacity to surface our reasons and values to AI systems that act on our behalf, and the accountability to ensure that each of our interests is fairly represented. After all, if we are going to align AI systems with normative standards we would collectively endorse, then we had better make sure that a ``we'' exists to endorse them.

\section{Conclusion}

Preference is a central concept in both the theory and practice of AI alignment. Yet as we have seen, its multiple scopes and meanings are often poorly understood. In this paper, we have sought not only to better contextualize the nature of preferences, but also to challenge its centrality in approaches to AI alignment.  In doing so, we hope to have established the goals of AI alignment on firmer normative ground. Crucially, we do not do so by rejecting all preference-based frameworks in alignment, but by reinterpreting what preferences do for us: Since they are \emph{constructed} from our values, norms, and reasons, they are \emph{informative} of those underlying structures. As such, preferences can serve as proxies for our values, but not targets of alignment in and of themselves.

What would AI alignment look like if it took these challenges seriously? It would move away from naive rational choice models of human decision making, towards richer models that include how we evaluate, commensurate, and act upon our values in boundedly rational ways. It would no longer take for granted expected utility theory, and instead explore systems for reasoning about the normativity of our preferences and values. It would learn to distinguish goodness-of-a-kind preferences from all-things-considered preferences, and identify which of those are operative in any particular decision. It would let go of preference matching as a crisp formalization of alignment, and instead lean into the normative complexity of scoping and defining AI's social roles. And it would move beyond alignment with aggregate preferences, towards a more pluralistic and contractualist understanding of what it means to live together with AI. If successful, then perhaps the world we can look forward to is not just one we will prefer, but one that we will truly have reason to value.

\section*{Acknowledgments}

This paper benefited from comments and feedback provided by participants at the closing retreat of the 2023 Principles of Intelligent Behavior in Biological and Social Systems (PIBBSS) Summer Fellowship, where an early version of this work was presented. We would also like to thank participants of the 2024 Sociotechnical AI Safety Workshop in Rio de Janeiro for their engagement and suggestions, and Seth Lazar for organizing the workshop. Conversations with many individuals informed the development and presentation of the ideas in this paper, including members of the 2019--2020 MIT AI Alignment Reading Group, Tushita Jha, Jonathan Stray, Iason Gabriel, Nora Ammann, Cecilia Wood, Mateusz Bagiński, Joe Kwon, Sydney Levine, Max Kleiman-Weiner, Max Langenkamp, Saffron Huang, Divya Siddarth, Gillian Hadfield, Vikash Mansinghka, and Joshua Tenenbaum.  Finally, we thank our anonymous reviewers for their highly detailed feedback and suggestions, which improved the clarity of our paper on many technical and conceptual points. Tan Zhi-Xuan is funded by the Open Philanthropy AI Fellowship. Micah Carroll is funded by the NSF Fellowship. Matija Franklin was funded by a UCL demonstratorship.

\bibliographystyle{apalike}
\setlength{\bibhang}{0pt}
\bibliography{bibliography}

\begin{thebibliography}{}

\bibitem[Ab~Azar et~al., 2020]{ab2020inverse}
Ab~Azar, N., Shahmansoorian, A., and Davoudi, M. (2020).
\newblock From inverse optimal control to inverse reinforcement learning: A
  historical review.
\newblock {\em Annual Reviews in Control}, 50:119--138.

\bibitem[Abbeel and Ng, 2004]{abbeel2004apprenticeship}
Abbeel, P. and Ng, A.~Y. (2004).
\newblock Apprenticeship learning via inverse reinforcement learning.
\newblock In {\em Proceedings of the Twenty-First International Conference on
  Machine Learning}, page~1.

\bibitem[Abel et~al., 2021]{abel2021expressivity}
Abel, D., Dabney, W., Harutyunyan, A., Ho, M.~K., Littman, M., Precup, D., and
  Singh, S. (2021).
\newblock On the expressivity of {M}arkov reward.
\newblock {\em Advances in Neural Information Processing Systems}, 34.

\bibitem[Akrour et~al., 2014]{akrour2014programming}
Akrour, R., Schoenauer, M., Sebag, M., and Souplet, J.-C. (2014).
\newblock Programming by feedback.
\newblock In {\em International Conference on Machine Learning}, pages
  1503--1511. JMLR. org.

\bibitem[Aky{\"u}rek et~al., 2024]{akyurek2024deductive}
Aky{\"u}rek, A.~F., Aky{\"u}rek, E., Choshen, L., Wijaya, D., and Andreas, J.
  (2024).
\newblock Deductive closure training of language models for coherence,
  accuracy, and updatability.
\newblock In {\em Findings of the Association for Computational Linguistics ACL
  2024}. Association for Computational Linguistics.

\bibitem[Alanqary et~al., 2021]{alanqary2021modeling}
Alanqary, A., Lin, G.~Z., Le, J., Zhi-Xuan, T., Mansinghka, V.~K., and
  Tenenbaum, J.~B. (2021).
\newblock Modeling the {M}istakes of {B}oundedly {R}ational {A}gents {W}ithin a
  {B}ayesian {T}heory of {M}ind.
\newblock In {\em Proceedings of the Annual Meeting of the Cognitive Science
  Society}, volume~43.

\bibitem[Althusser et~al., 2006]{althusser2006ideology}
Althusser, L. et~al. (2006).
\newblock {I}deology and {I}deological {S}tate {A}pparatuses.
\newblock {\em The Anthropology of the State: A Reader}, 9(1):86--98.

\bibitem[Amgoud et~al., 2017]{amgoud2017acceptability}
Amgoud, L., Ben-Naim, J., Doder, D., and Vesic, S. (2017).
\newblock Acceptability semantics for weighted argumentation frameworks.
\newblock In {\em Twenty-Sixth International Joint Conference on Artificial
  Intelligence (IJCAI 2017)}. International Joint Conferences on Artifical
  Intelligence (IJCAI).

\bibitem[Amgoud and Cayrol, 1998]{amgoud1998acceptability}
Amgoud, L. and Cayrol, C. (1998).
\newblock On the acceptability of arguments in preference-based argumentation.
\newblock In {\em Proceedings of the Fourteenth Conference on Uncertainty in
  Artificial Intelligence}.

\bibitem[Ammann, 2023]{ammann2023value}
Ammann, N. (2023).
\newblock The {Value} {Change} {Problem} (sequence).
\newblock {\em AI Alignment Forum}.
\newblock \url{https://www.alignmentforum.org/s/3QXNgNKXoLrdXJwWE}.

\bibitem[Amplayo et~al., 2019]{amplayo2019evaluating}
Amplayo, R.~K., Hwang, S.-w., and Song, M. (2019).
\newblock Evaluating research novelty detection: Counterfactual approaches.
\newblock In {\em Proceedings of the Thirteenth Workshop on Graph-Based Methods
  for Natural Language Processing (TextGraphs-13)}, pages 124--133.

\bibitem[Anderson, 1995]{anderson1995value}
Anderson, E. (1995).
\newblock {\em Value in Ethics and Economics}.
\newblock Harvard University Press.

\bibitem[Armstrong, 2019]{armstrong2019research}
Armstrong, S. (2019).
\newblock Synthesising a human’s preferences into a utility function.
\newblock {\em AI Alignment Forum}.
\newblock \url{https://www.alignmentforum.org/posts/CSEdLLEkap2pubjof}.

\bibitem[Armstrong and Levinstein, 2017]{armstrong2017low}
Armstrong, S. and Levinstein, B. (2017).
\newblock Low {Impact} {Artificial} {Intelligences}.
\newblock {\em arXiv:1705.10720 [cs]}.
\newblock arXiv: 1705.10720.

\bibitem[Armstrong and Mindermann, 2018]{armstrong2018occam}
Armstrong, S. and Mindermann, S. (2018).
\newblock Occam's razor is insufficient to infer the preferences of irrational
  agents.
\newblock {\em Advances in Neural Information Processing Systems}, 31.

\bibitem[Armstrong and O'Rourke, 2017]{armstrong2017indifference}
Armstrong, S. and O'Rourke, X. (2017).
\newblock Indifference methods for managing agent rewards.
\newblock {\em arXiv preprint arXiv:1712.06365}.

\bibitem[Ashton and Franklin, 2022]{ashton2022problem}
Ashton, H. and Franklin, M. (2022).
\newblock The {Problem} of {Behaviour} and {Preference} {Manipulation} in {AI}
  {Systems}.
\newblock In {\em The {AAAI}-22 {Workshop} on {Artificial} {Intelligence}
  {Safety} ({SafeAI} 2022)}.

\bibitem[Azari~Soufiani et~al., 2013]{azari_soufiani2013generalized}
Azari~Soufiani, H., Diao, H., Lai, Z., and Parkes, D.~C. (2013).
\newblock Generalized random utility models with multiple types.
\newblock {\em Advances in Neural Information Processing Systems}, 26.

\bibitem[Baber, 2011]{baber2011preference}
Baber, H.~E. (2011).
\newblock Preference-{Satisfaction}.
\newblock In Chatterjee, D.~K., editor, {\em Encyclopedia of {Global}
  {Justice}}, pages 890--896. Springer Netherlands, Dordrecht.

\bibitem[Bai et~al., 2022a]{bai2022training}
Bai, Y., Jones, A., Ndousse, K., Askell, A., Chen, A., DasSarma, N., Drain, D.,
  Fort, S., Ganguli, D., Henighan, T., et~al. (2022a).
\newblock Training a helpful and harmless assistant with reinforcement learning
  from human feedback.
\newblock {\em arXiv preprint arXiv:2204.05862}.

\bibitem[Bai et~al., 2022b]{bai2022constitutional}
Bai, Y., Kadavath, S., Kundu, S., Askell, A., Kernion, J., Jones, A., Chen, A.,
  Goldie, A., Mirhoseini, A., McKinnon, C., Chen, C., Olsson, C., Olah, C.,
  Hernandez, D., Drain, D., Ganguli, D., Li, D., Tran-Johnson, E., Perez, E.,
  Kerr, J., Mueller, J., Ladish, J., Landau, J., Ndousse, K., Lukosuite, K.,
  Lovitt, L., Sellitto, M., Elhage, N., Schiefer, N., Mercado, N., DasSarma,
  N., Lasenby, R., Larson, R., Ringer, S., Johnston, S., Kravec, S., Showk,
  S.~E., Fort, S., Lanham, T., Telleen-Lawton, T., Conerly, T., Henighan, T.,
  Hume, T., Bowman, S.~R., Hatfield-Dodds, Z., Mann, B., Amodei, D., Joseph,
  N., McCandlish, S., Brown, T., and Kaplan, J. (2022b).
\newblock Constitutional {AI}: {Harmlessness} from {AI} {Feedback}.
\newblock {\em arXiv preprint arXiv:2212.08073}.

\bibitem[Baker et~al., 2024]{baker2024roles}
Baker, A., Dunham, Y., and Jara-Ettinger, J. (2024).
\newblock Roles guide rapid inferences about agent knowledge and behavior.
\newblock In {\em Proceedings of the Annual Meeting of the Cognitive Science
  Society}, volume~46.

\bibitem[Baker et~al., 2009]{baker2009action}
Baker, C.~L., Saxe, R., and Tenenbaum, J.~B. (2009).
\newblock Action understanding as inverse planning.
\newblock {\em Cognition}, 113(3):329--349.

\bibitem[Bakker et~al., 2022]{bakker2022fine-tuning}
Bakker, M., Chadwick, M., Sheahan, H., Tessler, M., Campbell-Gillingham, L.,
  Balaguer, J., McAleese, N., Glaese, A., Aslanides, J., Botvinick, M., et~al.
  (2022).
\newblock Fine-tuning language models to find agreement among humans with
  diverse preferences.
\newblock {\em Advances in Neural Information Processing Systems},
  35:38176--38189.

\bibitem[Bales, 2023]{bales2023will}
Bales, A. (2023).
\newblock Will {AI} avoid exploitation? {A}rtificial {G}eneral {I}ntelligence
  and {E}xpected {U}tility {T}heory.
\newblock {\em Philosophical Studies}, pages 1--20.

\bibitem[Barreto et~al., 2017]{barreto2017successor}
Barreto, A., Dabney, W., Munos, R., Hunt, J.~J., Schaul, T., van Hasselt,
  H.~P., and Silver, D. (2017).
\newblock Successor features for transfer in reinforcement learning.
\newblock {\em Advances in Neural Information Processing Systems}, 30.

\bibitem[Bastani, 2019]{bastani2019fully}
Bastani, A. (2019).
\newblock {\em Fully Automated Luxury Communism}.
\newblock Verso Books.

\bibitem[Baum, 2020]{baum2020social}
Baum, S.~D. (2020).
\newblock Social choice ethics in {A}rtificial {I}ntelligence.
\newblock {\em AI \& Society}, 35(1):165--176.

\bibitem[Bengio, 2023]{bengio2023scientists}
Bengio, Y. (2023).
\newblock {AI} scientists: Safe and useful {AI}?
\newblock
  \url{https://yoshuabengio.org/2023/05/07/ai-scientists-safe-and-useful-ai/}.

\bibitem[Bentham, 1789]{bentham1789introduction}
Bentham, J. (1789).
\newblock {\em An Introduction to the Principles of Morals and Legislation}.
\newblock T. Payne and Son.

\bibitem[Berger, 2013]{berger2013statistical}
Berger, J. (2013).
\newblock {\em Statistical Decision Theory: Foundations, Concepts, and
  Methods}.
\newblock Springer Science \& Business Media.

\bibitem[Berke et~al., 2023]{berke2023thinking}
Berke, M., Tenenbaum, A., Sterling, B., and Jara-Ettinger, J. (2023).
\newblock Thinking about thinking as rational computation.
\newblock In {\em Proceedings of the Annual Conference of the Cognitive Science
  Society}.

\bibitem[Bicchieri, 2005]{bicchieri2005grammar}
Bicchieri, C. (2005).
\newblock {\em The {G}rammar of {S}ociety: The Nature and Dynamics of Social
  Norms}.
\newblock Cambridge University Press.

\bibitem[Binmore, 1994]{binmore1994game}
Binmore, K.~G. (1994).
\newblock {\em {G}ame {T}heory and the {S}ocial {C}ontract}.
\newblock MIT Press.

\bibitem[Birhane et~al., 2022]{birhane2022power}
Birhane, A., Isaac, W., Prabhakaran, V., Diaz, M., Elish, M.~C., Gabriel, I.,
  and Mohamed, S. (2022).
\newblock Power to the people? opportunities and challenges for participatory
  {AI}.
\newblock In {\em Proceedings of the 2nd ACM Conference on Equity and Access in
  Algorithms, Mechanisms, and Optimization}, pages 1--8.

\bibitem[Blili-Hamelin and Hancox-Li, 2023]{blili2023making}
Blili-Hamelin, B. and Hancox-Li, L. (2023).
\newblock Making intelligence: Ethical values in {IQ} and {ML} benchmarks.
\newblock In {\em Proceedings of the 2023 ACM Conference on Fairness,
  Accountability, and Transparency}, pages 271--284.

\bibitem[Blili-Hamelin et~al., 2024]{blili2024unsocial}
Blili-Hamelin, B., Hancox-Li, L., and Smart, A. (2024).
\newblock {Unsocial} {Intelligence}: {An} {Investigation} of the {Assumptions}
  of {AGI} {Discourse}.
\newblock In {\em Proceedings of the AAAI/ACM Conference on AI, Ethics, and
  Society}, volume~7, pages 141--155.

\bibitem[Bobu et~al., 2024]{bobu2024aligning}
Bobu, A., Peng, A., Agrawal, P., Shah, J., and Dragan, A.~D. (2024).
\newblock Aligning robot and human representations.
\newblock In {\em Proceedings of the 2024 ACM/IEEE International Conference on
  Human-Robot Interaction}. Association for Computing Machinery.

\bibitem[Bobu et~al., 2022]{bobu2022inducing}
Bobu, A., Wiggert, M., Tomlin, C., and Dragan, A.~D. (2022).
\newblock Inducing structure in reward learning by learning features.
\newblock {\em The International Journal of Robotics Research}, 41(5):497--518.

\bibitem[Bolker, 1967]{bolker1967simultaneous}
Bolker, E.~D. (1967).
\newblock A {Simultaneous} {Axiomatization} of {Utility} and {Subjective}
  {Probability}.
\newblock {\em Philosophy of Science}, 34(4):333--340.

\bibitem[Booth et~al., 2023]{booth2023perils}
Booth, S., Knox, W.~B., Shah, J., Niekum, S., Stone, P., and Allievi, A.
  (2023).
\newblock The perils of trial-and-error reward design: {M}isdesign through
  overfitting and invalid task specifications.
\newblock In {\em Proceedings of the AAAI Conference on Artificial
  Intelligence}, volume~37, pages 5920--5929.

\bibitem[Bossaerts et~al., 2019]{bossaerts2019uncertainty}
Bossaerts, P., Yadav, N., and Murawski, C. (2019).
\newblock Uncertainty and computational complexity.
\newblock {\em Philosophical Transactions of the Royal Society B},
  374(1766):20180138.

\bibitem[Boudon, 2003]{boudon2003beyond}
Boudon, R. (2003).
\newblock Beyond rational choice theory.
\newblock {\em Annual Review of Sociology}, 29(1):1--21.

\bibitem[Boutilier et~al., 2004]{boutilier2004cp-nets}
Boutilier, C., Brafman, R.~I., Domshlak, C., Hoos, H.~H., and Poole, D. (2004).
\newblock {CP}-nets: {A} {Tool} for {Representing} and {Reasoning} with
  {Conditional} {Ceteris} {Paribus} {Preference} {Statements}.
\newblock {\em Journal of Artificial Intelligence Research}, 21:135--191.

\bibitem[Bowling et~al., 2023]{bowling2023settling}
Bowling, M., Martin, J.~D., Abel, D., and Dabney, W. (2023).
\newblock Settling the {R}eward {H}ypothesis.
\newblock In {\em International Conference on Machine Learning}, pages
  3003--3020. PMLR.

\bibitem[Bradley and Steele, 2014]{bradley2014should}
Bradley, S. and Steele, K. (2014).
\newblock Should subjective probabilities be sharp?
\newblock {\em Episteme}, 11(3):277--289.

\bibitem[Brandt, 1955]{brandt1955definition}
Brandt, R.~B. (1955).
\newblock The definition of an ``ideal observer'' theory in ethics.
\newblock {\em Philosophy and Phenomenological Research}, 15(3):407--413.

\bibitem[Bratman, 1987]{bratman1987intention}
Bratman, M. (1987).
\newblock {\em Intention, Plans, and Practical Reason}.
\newblock Cambridge, MA: Harvard University Press, Cambridge.

\bibitem[Bratman et~al., 1988]{bratman1988plans}
Bratman, M.~E., Israel, D.~J., and Pollack, M.~E. (1988).
\newblock Plans and resource-bounded practical reasoning.
\newblock {\em Computational Intelligence}, 4(3):349--355.

\bibitem[Butlin, 2021]{butlin2021ai}
Butlin, P. (2021).
\newblock {AI} alignment and human reward.
\newblock In {\em Proceedings of the 2021 AAAI/ACM Conference on AI, Ethics,
  and Society}, pages 437--445.

\bibitem[Callaway et~al., 2022]{callaway2022rational}
Callaway, F., van Opheusden, B., Gul, S., Das, P., Krueger, P.~M., Griffiths,
  T.~L., and Lieder, F. (2022).
\newblock Rational use of cognitive resources in human planning.
\newblock {\em Nature Human Behaviour}, 6(8):1112--1125.

\bibitem[Camara, 2022]{camara2022computationally}
Camara, M.~K. (2022).
\newblock Computationally tractable choice.
\newblock In {\em Proceedings of the 23rd ACM Conference on Economics and
  Computation}, pages 28--28.

\bibitem[Cao et~al., 2021]{cao2021identifiability}
Cao, H., Cohen, S., and Szpruch, L. (2021).
\newblock Identifiability in inverse reinforcement learning.
\newblock {\em Advances in Neural Information Processing Systems},
  34:12362--12373.

\bibitem[Carlsmith, 2022]{carlsmith2022power}
Carlsmith, J. (2022).
\newblock Is power-seeking {AI} an existential risk?
\newblock {\em arXiv preprint arXiv:2206.13353}.

\bibitem[Carroll et~al., 2023]{carroll2023characterizing}
Carroll, M., Chan, A., Ashton, H., and Krueger, D. (2023).
\newblock Characterizing manipulation from {AI} systems.
\newblock In {\em Proceedings of the 3rd ACM Conference on Equity and Access in
  Algorithms, Mechanisms, and Optimization}, pages 1--13.

\bibitem[Carroll et~al., 2024]{carroll2024ai}
Carroll, M., Foote, D., Siththaranjan, A., Russell, S., and Dragan, A. (2024).
\newblock {AI} alignment with changing and influenceable reward functions.
\newblock In {\em Proceedings of the 41st International Conference on Machine
  Learning}, pages 5706--5756.

\bibitem[Carroll et~al., 2022]{carroll2022estimating}
Carroll, M.~D., Dragan, A., Russell, S., and Hadfield-Menell, D. (2022).
\newblock Estimating and penalizing induced preference shifts in recommender
  systems.
\newblock In {\em International Conference on Machine Learning}, pages
  2686--2708. PMLR.

\bibitem[Casper et~al., 2023]{casper2023open}
Casper, S., Davies, X., Shi, C., Gilbert, T.~K., Scheurer, J., Rando, J.,
  Freedman, R., Korbak, T., Lindner, D., Freire, P., Wang, T., Marks, S.,
  Segerie, C.-R., Carroll, M., Peng, A., Christoffersen, P., Damani, M.,
  Slocum, S., Anwar, U., Siththaranjan, A., Nadeau, M., Michaud, E.~J., Pfau,
  J., Krasheninnikov, D., Chen, X., Langosco, L., Hase, P., Bıyık, E.,
  Dragan, A., Krueger, D., Sadigh, D., and Hadfield-Menell, D. (2023).
\newblock Open {Problems} and {Fundamental} {Limitations} of {Reinforcement}
  {Learning} from {Human} {Feedback}.
\newblock {\em Transactions on Machine Learning Research}.

\bibitem[Castagna et~al., 2024]{castagna2024can}
Castagna, F., Sassoon, I., and Parsons, S. (2024).
\newblock Can formal argumentative reasoning enhance {LLMs'} performances?
\newblock {\em arXiv preprint arXiv:2405.13036}.

\bibitem[Cath, 2016]{cath2016reflective}
Cath, Y. (2016).
\newblock Reflective equilibrium.
\newblock {\em The Oxford Handbook of Philosophical Methodology}, 1.

\bibitem[Cettolin and Riedl, 2019]{cettolin2019revealed}
Cettolin, E. and Riedl, A. (2019).
\newblock Revealed preferences under uncertainty: Incomplete preferences and
  preferences for randomization.
\newblock {\em Journal of Economic Theory}, 181:547--585.

\bibitem[Chan et~al., 2021]{chan2021human}
Chan, L., Critch, A., and Dragan, A. (2021).
\newblock Human irrationality: {B}oth bad and good for reward inference.
\newblock {\em arXiv preprint arXiv:2111.06956}.

\bibitem[Chan et~al., 2019]{chan2019assistive}
Chan, L., Hadfield-Menell, D., Srinivasa, S., and Dragan, A. (2019).
\newblock The {Assistive} {Multi}-{Armed} {Bandit}.
\newblock In {\em 2019 14th ACM/IEEE International Conference on Human-Robot
  Interaction (HRI)}, pages 354--363. IEEE.

\bibitem[Chang, 1997]{chang1997incommensurability}
Chang, R., editor (1997).
\newblock {\em Incommensurability, Incomparability, and Practical Reason}.
\newblock Harvard, Cambridge, MA, USA.

\bibitem[Chang, 2004]{chang2004can}
Chang, R. (2004).
\newblock Can desires provide reasons for action.
\newblock In Wallace, R.~J., Pettit, P., Scheffler, S., and Smith, M., editors,
  {\em Reason and Value: Themes From the Moral Philosophy of Joseph Raz}, pages
  56--90. Oxford University Press.

\bibitem[Chang, 2009]{chang2009voluntarist}
Chang, R. (2009).
\newblock {Voluntarist} {Reasons} and the {Sources} of {Normativity}.
\newblock In Sobel, D. and Wall, S., editors, {\em Reasons for Action}, pages
  243--71. Cambridge University Press.

\bibitem[Chang, 2021]{chang2021how}
Chang, R. (2021).
\newblock How to prevent {AI} from taking over the world.
\newblock
  \url{https://www.newstatesman.com/ideas/2021/02/how-prevent-ai-taking-over-world}.

\bibitem[Chater, 2023]{chater2023could}
Chater, N. (2023).
\newblock How could we make a social robot? {A} virtual bargaining approach.
\newblock {\em Philosophical Transactions of the Royal Society A},
  381(2251):20220040.

\bibitem[Chater and Oaksford, 1999]{chater1999ten}
Chater, N. and Oaksford, M. (1999).
\newblock Ten years of the rational analysis of cognition.
\newblock {\em Trends in Cognitive Sciences}, 3(2):57--65.

\bibitem[Chatterjee et~al., 2016]{chatterjee2016decidable}
Chatterjee, K., Chmelik, M., and Tracol, M. (2016).
\newblock What is decidable about partially observable {M}arkov decision
  processes with $\omega$-regular objectives.
\newblock {\em Journal of Computer and System Sciences}, 82(5):878--911.

\bibitem[Christiano, 2015a]{christiano2015ambitious}
Christiano, P. (2015a).
\newblock Ambitious vs. narrow value learning.
\newblock
  \url{https://ai-alignment.com/ambitious-vs-narrow-value-learning-99bd0c59847e}.

\bibitem[Christiano, 2015b]{christiano2015easy}
Christiano, P. (2015b).
\newblock The easy goal inference problem is still hard.
\newblock {\em AI Alignment Forum}.

\bibitem[Christiano et~al., 2017]{christiano2017deep}
Christiano, P.~F., Leike, J., Brown, T., Martic, M., Legg, S., and Amodei, D.
  (2017).
\newblock Deep reinforcement learning from human preferences.
\newblock {\em Advances in Neural Information Processing Systems}, 30.

\bibitem[Christoffersen et~al., 2023]{christoffersen2023get}
Christoffersen, P.~J., Haupt, A.~A., and Hadfield-Menell, D. (2023).
\newblock Get it in writing: {F}ormal contracts mitigate social dilemmas in
  multi-agent {RL}.
\newblock In {\em Proceedings of the 2023 International Conference on
  Autonomous Agents and Multiagent Systems}, pages 448--456.

\bibitem[Clayton and Williams, 1999]{clayton1999egalitarian}
Clayton, M. and Williams, A. (1999).
\newblock Egalitarian justice and interpersonal comparison.
\newblock {\em European Journal of Political Research}, 35(4):445--464.

\bibitem[Conitzer et~al., 2024]{conitzer2024position}
Conitzer, V., Freedman, R., Heitzig, J., Holliday, W.~H., Jacobs, B.~M.,
  Lambert, N., Moss{\'e}, M., Pacuit, E., Russell, S., Schoelkopf, H., Tewolde,
  E., and Zwicker, W.~S. (2024).
\newblock Position: Social choice should guide {AI} alignment in dealing with
  diverse human feedback.
\newblock In {\em Forty-first International Conference on Machine Learning}.

\bibitem[Cornelio et~al., 2013]{cornelio2013updates}
Cornelio, C., Goldsmith, J., Mattei, N., Rossi, F., and Venable, K.~B. (2013).
\newblock Updates and {Uncertainty} in {CP}-{Nets}.
\newblock In Hutchison, D., Kanade, T., Kittler, J., Kleinberg, J.~M., Mattern,
  F., Mitchell, J.~C., Naor, M., Nierstrasz, O., Pandu~Rangan, C., Steffen, B.,
  Sudan, M., Terzopoulos, D., Tygar, D., Vardi, M.~Y., Weikum, G., Cranefield,
  S., and Nayak, A., editors, {\em {AI} 2013: {Advances} in {Artificial}
  {Intelligence}}, volume 8272, pages 301--312. Springer International
  Publishing, Cham.
\newblock Series Title: Lecture Notes in Computer Science.

\bibitem[Critch and Krueger, 2020]{critch2020ai}
Critch, A. and Krueger, D. (2020).
\newblock {AI} research considerations for human existential safety ({ARCHES}).
\newblock {\em arXiv preprint arXiv:2006.04948}.

\bibitem[Critch and Russell, 2017]{critch2017servant}
Critch, A. and Russell, S. (2017).
\newblock Servant of many masters: Shifting priorities in {P}areto-optimal
  sequential decision-making.
\newblock {\em arXiv preprint arXiv:1711.00363}.

\bibitem[Cudd and Eftekhari, 2021]{cudd2021contractarianism}
Cudd, A. and Eftekhari, S. (2021).
\newblock {Contractarianism}.
\newblock In Zalta, E.~N., editor, {\em The {Stanford} Encyclopedia of
  Philosophy}. Metaphysics Research Lab, Stanford University, {W}inter 2021
  edition.

\bibitem[Cusumano-Towner et~al., 2019]{cusumano2019gen}
Cusumano-Towner, M.~F., Saad, F.~A., Lew, A.~K., and Mansinghka, V.~K. (2019).
\newblock Gen: {A} general-purpose probabilistic programming system with
  programmable inference.
\newblock In {\em Proceedings of the 40th ACM SIGPLAN Conference on Programming
  Language Design and Implementation}, pages 221--236.

\bibitem[Cwik and Engelhardt, 2024]{cwik2024revisiting}
Cwik, P. and Engelhardt, L. (2024).
\newblock Revisiting the computation problem.
\newblock {\em Quarterly Journal of Austrian Economics}, 26(3).

\bibitem[Dafoe et~al., 2020]{dafoe2020open}
Dafoe, A., Hughes, E., Bachrach, Y., Collins, T., McKee, K.~R., Leibo, J.~Z.,
  Larson, K., and Graepel, T. (2020).
\newblock Open problems in cooperative {AI}.
\newblock {\em arXiv preprint arXiv:2012.08630}.

\bibitem[Dalrymple, 2022]{dalrymple2022you}
Dalrymple, D.~D. (2022).
\newblock You can still fetch the coffee today if you're dead tomorrow.
\newblock {\em AI Alignment Forum}.
\newblock
  \url{https://www.alignmentforum.org/posts/dzDKDRJPQ3kGqfER9/you-can-still-fetch-the-coffee-today-if-you-re-dead-tomorrow}.

\bibitem[Dalrymple, 2024]{dalrymple2024safeguarded}
Dalrymple, D.~D. (2024).
\newblock Safeguarded {AI}: {C}onstructing guaranteed safety.
\newblock Technical report, ARIA.

\bibitem[Dalrymple et~al., 2024]{dalrymple2024guaranteed}
Dalrymple, D.~D., Skalse, J., Bengio, Y., Russell, S., Tegmark, M., Seshia, S.,
  Omohundro, S., Szegedy, C., Goldhaber, B., Ammann, N., Abate, A., Halpern,
  J., Barrett, C., Zhao, D., Zhi-Xuan, T., Wing, J., and Tenenbaum, J. (2024).
\newblock Towards guaranteed safe {AI}: {A} framework for ensuring robust and
  reliable {AI} systems.
\newblock {\em arXiv preprint arXiv:2405.06624}.

\bibitem[Davani et~al., 2022]{davani2022dealing}
Davani, A.~M., D{\'\i}az, M., and Prabhakaran, V. (2022).
\newblock Dealing with disagreements: Looking beyond the majority vote in
  subjective annotations.
\newblock {\em Transactions of the Association for Computational Linguistics},
  10:92--110.

\bibitem[Davidson et~al., 2024]{davidson2024goals}
Davidson, G., Todd, G., Togelius, J., Gureckis, T.~M., and Lake, B.~M. (2024).
\newblock Goals as reward-producing programs.
\newblock {\em arXiv preprint arXiv:2405.13242}.

\bibitem[De~Raedt and Kersting, 2003]{de2003probabilistic}
De~Raedt, L. and Kersting, K. (2003).
\newblock Probabilistic logic learning.
\newblock {\em ACM SIGKDD Explorations Newsletter}, 5(1):31--48.

\bibitem[Demski, 2018]{demski2018complete}
Demski, A. (2018).
\newblock Complete class: Consequentialist foundations.
\newblock {\em AI Alignment Forum}.
\newblock \url{https://www.alignmentforum.org/posts/sZuw6SGfmZHvcAAEP}.

\bibitem[Denoeux and Shenoy, 2020]{denoeux2020interval}
Denoeux, T. and Shenoy, P.~P. (2020).
\newblock An interval-valued utility theory for decision making with
  {Dempster-Shafer} belief functions.
\newblock {\em International Journal of Approximate Reasoning}, 124:194--216.

\bibitem[Dewey, 2011]{dewey2011learning}
Dewey, D. (2011).
\newblock Learning what to value.
\newblock In {\em {Artificial} {General} {Intelligence}: 4th {International}
  {Conference}, {AGI} 2011, {Mountain} {View}, {CA}, {USA}, {August} 3-6, 2011.
  {Proceedings} 4}, pages 309--314. Springer, Springer.

\bibitem[Di~Langosco et~al., 2022]{langosco2022goal}
Di~Langosco, L.~L., Koch, J., Sharkey, L.~D., Pfau, J., and Krueger, D. (2022).
\newblock Goal misgeneralization in deep reinforcement learning.
\newblock In {\em International Conference on Machine Learning}, pages
  12004--12019. PMLR.

\bibitem[Drexler, 2022]{drexler2022open}
Drexler, E. (2022).
\newblock The open agency model.
\newblock {\em AI Alignment Forum}.
\newblock \url{https://www.alignmentforum.org/posts/5hApNw5f7uG8RXxGS}.

\bibitem[Drexler, 2019]{drexler2019reframing}
Drexler, K.~E. (2019).
\newblock Reframing {S}uperintelligence: {Comprehensive} {AI} services as
  general intelligence.
\newblock Technical {Report} 2019-1, Future of Humanity Institute, Oxford.

\bibitem[Dror, 2023]{dror2023there}
Dror, L. (2023).
\newblock Is there an epistemic advantage to being oppressed?
\newblock {\em No{\^u}s}, 57(3):618--640.

\bibitem[Du et~al., 2020]{du2020ave}
Du, Y., Tiomkin, S., Kiciman, E., Polani, D., Abbeel, P., and Dragan, A.
  (2020).
\newblock {AVE}: Assistance via empowerment.
\newblock {\em Advances in Neural Information Processing Systems},
  33:4560--4571.

\bibitem[Dumoulin et~al., 2024]{dumoulin2023density}
Dumoulin, V., Johnson, D.~D., Castro, P.~S., Larochelle, H., and Dauphin, Y.
  (2024).
\newblock A density estimation perspective on learning from pairwise human
  preferences.
\newblock {\em Transactions on Machine Learning Research}.

\bibitem[Dung, 1995]{dung1995acceptability}
Dung, P.~M. (1995).
\newblock On the acceptability of arguments and its fundamental role in
  nonmonotonic reasoning, logic programming and n-person games.
\newblock {\em Artificial Intelligence}, 77(2):321--357.

\bibitem[Dziri et~al., 2023]{dziri2023faith}
Dziri, N., Lu, X., Sclar, M., Li, X.~L., Jian, L., Lin, B.~Y., West, P.,
  Bhagavatula, C., Bras, R.~L., Hwang, J.~D., et~al. (2023).
\newblock Faith and fate: Limits of transformers on compositionality.
\newblock {\em Advances in Neural Information Processing Systems}, 36.

\bibitem[Eckersley, 2018]{eckersley2018impossibility}
Eckersley, P. (2018).
\newblock Impossibility and uncertainty theorems in {AI} value alignment (or
  why your {AGI} should not have a utility function).
\newblock {\em arXiv preprint arXiv:1901.00064}.

\bibitem[Edwards, 2023]{edwards2023ai}
Edwards, B. (2023).
\newblock {AI}-powered {Bing} chat gains three distinct personalities.
\newblock {\em Ars Technica}.

\bibitem[Espeland and Stevens, 1998]{espeland1998commensuration}
Espeland, W.~N. and Stevens, M.~L. (1998).
\newblock Commensuration as a social process.
\newblock {\em Annual Review of Sociology}, 24(1):313--343.

\bibitem[Evans et~al., 2016]{evans2016learning}
Evans, O., Stuhlm{\"u}ller, A., and Goodman, N. (2016).
\newblock Learning the preferences of ignorant, inconsistent agents.
\newblock In {\em Proceedings of the AAAI Conference on Artificial
  Intelligence}, volume~30.

\bibitem[Everitt et~al., 2021]{everitt2021agent}
Everitt, T., Carey, R., Langlois, E.~D., Ortega, P.~A., and Legg, S. (2021).
\newblock Agent incentives: A causal perspective.
\newblock In {\em Proceedings of the AAAI Conference on Artificial
  Intelligence}, volume~35, pages 11487--11495.

\bibitem[Fickinger et~al., 2020]{fickinger2020multi-principal}
Fickinger, A., Zhuang, S., Hadfield-Menell, D., and Russell, S. (2020).
\newblock Multi-{Principal} {Assistance} {Games}.
\newblock {\em arXiv:2007.09540 [cs]}.
\newblock arXiv: 2007.09540.

\bibitem[Fikes and Nilsson, 1971]{fikes1971strips}
Fikes, R.~E. and Nilsson, N.~J. (1971).
\newblock {STRIPS}: A new approach to the application of theorem proving to
  problem solving.
\newblock {\em Artificial Intelligence}, 2(3-4):189--208.

\bibitem[Firth, 1952]{firth1952ethical}
Firth, R. (1952).
\newblock Ethical absolutism and the ideal observer.
\newblock {\em Philosophy and Phenomenological Research}, 12(3):317--345.

\bibitem[Franklin et~al., 2022]{franklin2022recognising}
Franklin, M., Ashton, H., Gorman, R., and Armstrong, S. (2022).
\newblock Recognising the importance of preference change: {A} call for a
  coordinated multidisciplinary research effort in the age of {AI}.
\newblock In {\em AAAI-22 Workshop on AI For Behavior Change}.

\bibitem[Frazier, 1994]{frazier1994act}
Frazier, R.~L. (1994).
\newblock Act utilitarianism and decision procedures.
\newblock {\em Utilitas}, 6(1):43--53.

\bibitem[Fricker, 2007]{fricker2007epistemic}
Fricker, M. (2007).
\newblock {\em Epistemic Injustice: Power and the Ethics of Knowing}.
\newblock Oxford University Press.

\bibitem[Gabriel, 2020]{gabriel2020artificial}
Gabriel, I. (2020).
\newblock {A}rtificial {I}ntelligence, {V}alues, and {A}lignment.
\newblock {\em Minds and Machines}, 30(3):411--437.

\bibitem[Gabriel and Keeling, 2024]{gabriel2024matter}
Gabriel, I. and Keeling, G. (2024).
\newblock A matter of principle? {AI} alignment as the fair treatment of
  claims.
\newblock \textit{Under Review}.

\bibitem[Gao et~al., 2023]{gao2023scaling}
Gao, L., Schulman, J., and Hilton, J. (2023).
\newblock Scaling laws for reward model overoptimization.
\newblock In {\em International Conference on Machine Learning}, pages
  10835--10866. PMLR.

\bibitem[Garrabrant, 2022]{garrabrant2022geometric}
Garrabrant, S. (2022).
\newblock Geometric rationality.
\newblock {\em AI Alignment Forum}.
\newblock \url{https://www.alignmentforum.org/s/4hmf7rdfuXDJkxhfg}.

\bibitem[Gauthier, 1986]{gauthier1986morals}
Gauthier, D. (1986).
\newblock {\em Morals by Agreement}.
\newblock Clarendon Press.

\bibitem[Gerevini and Long, 2005]{gerevini2005plan}
Gerevini, A. and Long, D. (2005).
\newblock Plan constraints and preferences in {PDDL3}.
\newblock Technical report, Department of Electronics for Automation,
  University of Brescia.

\bibitem[Gershman et~al., 2015]{gershman2015computational}
Gershman, S.~J., Horvitz, E.~J., and Tenenbaum, J.~B. (2015).
\newblock Computational rationality: A converging paradigm for intelligence in
  brains, minds, and machines.
\newblock {\em Science}, 349(6245):273--278.

\bibitem[Ghosal et~al., 2023]{ghosal2023effect}
Ghosal, G.~R., Zurek, M., Brown, D.~S., and Dragan, A.~D. (2023).
\newblock The effect of modeling human rationality level on learning rewards
  from multiple feedback types.
\newblock In {\em Proceedings of the AAAI Conference on Artificial
  Intelligence}, volume~37, pages 5983--5992.

\bibitem[Gigerenzer, 2008]{gigerenzer2008why}
Gigerenzer, G. (2008).
\newblock Why heuristics work.
\newblock {\em Perspectives on psychological science}, 3(1):20--29.
\newblock Publisher: SAGE Publications Sage CA: Los Angeles, CA.

\bibitem[Gintis, 2010]{gintis2010social}
Gintis, H. (2010).
\newblock Social norms as choreography.
\newblock {\em Politics, Philosophy \& Economics}, 9(3):251--264.

\bibitem[Go et~al., 2024]{go2024compositional}
Go, D., Korbak, T., Kruszewski, G., Rozen, J., and Dymetman, M. (2024).
\newblock Compositional preference models for aligning {LM}s.
\newblock In {\em The Twelfth International Conference on Learning
  Representations}.

\bibitem[Gordon et~al., 2022]{gordon2022jury}
Gordon, M.~L., Lam, M.~S., Park, J.~S., Patel, K., Hancock, J., Hashimoto, T.,
  and Bernstein, M.~S. (2022).
\newblock Jury learning: Integrating dissenting voices into machine learning
  models.
\newblock In {\em Proceedings of the 2022 CHI Conference on Human Factors in
  Computing Systems}, pages 1--19.

\bibitem[Gorwa et~al., 2020]{gorwa2020algorithmic}
Gorwa, R., Binns, R., and Katzenbach, C. (2020).
\newblock Algorithmic content moderation: Technical and political challenges in
  the automation of platform governance.
\newblock {\em Big Data \& Society}, 7(1):2053951719897945.

\bibitem[Griffith et~al., 2013]{griffith2013policy}
Griffith, S., Subramanian, K., Scholz, J., Isbell, C.~L., and Thomaz, A.~L.
  (2013).
\newblock Policy {Shaping}: {Integrating} {Human} {Feedback} with
  {Reinforcement} {Learning}.
\newblock In Burges, C.~J., Bottou, L., Welling, M., Ghahramani, Z., and
  Weinberger, K.~Q., editors, {\em Advances in {Neural} {Information}
  {Processing} {Systems}}, volume~26. Curran Associates, Inc.

\bibitem[Gustafsson, 2022]{gustafsson2022money}
Gustafsson, J.~E. (2022).
\newblock {\em Money-pump arguments}.
\newblock Cambridge University Press.

\bibitem[Hadfield-Menell et~al., 2019]{hadfield2019legible}
Hadfield-Menell, D., Andrus, M., and Hadfield, G. (2019).
\newblock Legible normativity for {AI} alignment: The value of silly rules.
\newblock In {\em Proceedings of the 2019 AAAI/ACM Conference on AI, Ethics,
  and Society}, pages 115--121.

\bibitem[Hadfield-Menell et~al., 2017a]{hadfield2017off}
Hadfield-Menell, D., Dragan, A., Abbeel, P., and Russell, S. (2017a).
\newblock The off-switch game.
\newblock In {\em Workshops at the Thirty-First AAAI Conference on Artificial
  Intelligence}.

\bibitem[Hadfield-Menell and Hadfield, 2018]{hadfield-menell2018incomplete}
Hadfield-Menell, D. and Hadfield, G.~K. (2018).
\newblock Incomplete contracting and {AI} alignment.
\newblock {\em USC CLASS Research Papers Series No. CLASS18-10}.

\bibitem[Hadfield-Menell et~al., 2017b]{hadfield2017inverse}
Hadfield-Menell, D., Milli, S., Abbeel, P., Russell, S.~J., and Dragan, A.
  (2017b).
\newblock Inverse reward design.
\newblock {\em Advances in Neural Information Processing Systems}, 30.

\bibitem[Hadfield-Menell et~al., 2016]{hadfield2016cooperative}
Hadfield-Menell, D., Russell, S.~J., Abbeel, P., and Dragan, A. (2016).
\newblock Cooperative inverse reinforcement learning.
\newblock {\em Advances in Neural Information Processing Systems}, 29.

\bibitem[Halpern and Pass, 2015]{halpern2015algorithmic}
Halpern, J.~Y. and Pass, R. (2015).
\newblock Algorithmic rationality: Game theory with costly computation.
\newblock {\em Journal of Economic Theory}, 156:246--268.

\bibitem[Hansson, 1990]{hansson1990preference}
Hansson, S.~O. (1990).
\newblock Preference-based deontic logic.
\newblock {\em Journal of Philosophical Logic}, 19:75--93.

\bibitem[Hare, 1981]{hare1981moral}
Hare, R.~M. (1981).
\newblock {\em Moral Thinking: Its Levels, Method, and Point}.
\newblock Oxford: Clarendon Press; New York: Oxford University Press.

\bibitem[Harsanyi, 1953]{harsanyi1953cardinal}
Harsanyi, J.~C. (1953).
\newblock Cardinal utility in welfare economics and in the theory of
  risk-taking.
\newblock {\em Journal of Political Economy}, 61(5):434--435.
\newblock Publisher: The University of Chicago Press.

\bibitem[Harsanyi, 1955]{harsanyi1955cardinal}
Harsanyi, J.~C. (1955).
\newblock Cardinal welfare, individualistic ethics, and interpersonal
  comparisons of utility.
\newblock {\em Journal of Political Economy}, 63(4):309--321.

\bibitem[Harsanyi, 1975]{harsanyi1975can}
Harsanyi, J.~C. (1975).
\newblock Can the maximin principle serve as a basis for morality? {A} critique
  of {John} {Rawls}'s theory.
\newblock {\em American Political Science Review}, 69(2):594--606.

\bibitem[Hawkins et~al., 2019]{hawkins2019emergence}
Hawkins, R.~X., Goodman, N.~D., and Goldstone, R.~L. (2019).
\newblock The emergence of social norms and conventions.
\newblock {\em Trends in Cognitive Sciences}, 23(2):158--169.

\bibitem[Hayden and Niv, 2021]{hayden2021case}
Hayden, B.~Y. and Niv, Y. (2021).
\newblock The case against economic values in the orbitofrontal cortex (or
  anywhere else in the brain).
\newblock {\em Behavioral Neuroscience}, 135(2):192.

\bibitem[Hayek, 1945]{hayek1945use}
Hayek, F. (1945).
\newblock The use of knowledge in society.
\newblock {\em American Economic Review}, 35(4).

\bibitem[Hedden, 2015]{hedden2015reasons}
Hedden, B. (2015).
\newblock {\em Reasons Without Persons: Rationality, Identity, and Time}.
\newblock OUP Oxford.

\bibitem[Hejna et~al., 2024]{hejna2024contrastive}
Hejna, J., Rafailov, R., Sikchi, H., Finn, C., Niekum, S., Knox, W.~B., and
  Sadigh, D. (2024).
\newblock Contrastive preference learning: Learning from human feedback without
  reinforcement learning.
\newblock In {\em The Twelfth International Conference on Learning
  Representations}.

\bibitem[Hendrycks, 2023]{hendrycks2023natural}
Hendrycks, D. (2023).
\newblock Natural selection favors {AI}s over humans.
\newblock {\em arXiv preprint arXiv:2303.16200}.

\bibitem[Hill et~al., 2017]{hill2017efficient}
Hill, D.~N., Nassif, H., Liu, Y., Iyer, A., and Vishwanathan, S. (2017).
\newblock An efficient bandit algorithm for realtime multivariate optimization.
\newblock In {\em Proceedings of the 23rd ACM SIGKDD International Conference
  on Knowledge Discovery and Data Mining}, pages 1813--1821.

\bibitem[Ho and Griffiths, 2022]{ho2022cognitive}
Ho, M.~K. and Griffiths, T.~L. (2022).
\newblock Cognitive science as a source of forward and inverse models of human
  decisions for robotics and control.
\newblock {\em Annual Review of Control, Robotics, and Autonomous Systems},
  5:33--53.

\bibitem[Holtman, 2019]{holtman2019corrigibility}
Holtman, K. (2019).
\newblock Corrigibility with utility preservation.
\newblock {\em arXiv preprint arXiv:1908.01695}.

\bibitem[Holtug, 2017]{holtug2017prioritarianism}
Holtug, N. (2017).
\newblock Prioritarianism.
\newblock In {\em Oxford Research Encyclopedia of Politics}. Oxford University
  Press.

\bibitem[Horowitz et~al., 1994]{horowitz1994advances}
Horowitz, J.~L., Bolduc, D., Divakar, S., Geweke, J., G{\"o}n{\"u}l, F.,
  Hajivassiliou, V., Koppelman, F.~S., Keane, M., Matzkin, R., Rossi, P.,
  et~al. (1994).
\newblock Advances in random utility models.
\newblock {\em Marketing Letters}, 5:311--322.

\bibitem[Huang et~al., 2024]{huang2024collective}
Huang, S., Siddarth, D., Lovitt, L., Liao, T.~I., Durmus, E., Tamkin, A., and
  Ganguli, D. (2024).
\newblock {C}ollective {C}onstitutional {AI}: Aligning a language model with
  public input.
\newblock In {\em Proceedings of the 2024 ACM Conference on Fairness,
  Accountability, and Transparency (FAccT 2024)}. ACM.

\bibitem[Hubinger et~al., 2019]{hubinger2019risks}
Hubinger, E., van Merwijk, C., Mikulik, V., Skalse, J., and Garrabrant, S.
  (2019).
\newblock Risks from learned optimization in advanced machine learning systems.
\newblock {\em arXiv preprint arXiv:1906.01820}.

\bibitem[Hull, 2023]{hull2023dirty}
Hull, G. (2023).
\newblock Dirty data labeled dirt cheap: {E}pistemic injustice in machine
  learning systems.
\newblock {\em Ethics and Information Technology}, 25(3):38.

\bibitem[Icarte et~al., 2022]{icarte2022reward}
Icarte, R.~T., Klassen, T.~Q., Valenzano, R., and McIlraith, S.~A. (2022).
\newblock Reward machines: Exploiting reward function structure in
  reinforcement learning.
\newblock {\em Journal of Artificial Intelligence Research}, 73:173--208.

\bibitem[Irving et~al., 2018]{irving2018ai}
Irving, G., Christiano, P., and Amodei, D. (2018).
\newblock {AI} safety via debate.
\newblock {\em arXiv preprint arXiv:1805.00899}.
\newblock arXiv:1805.00899 [cs, stat].

\bibitem[Iu and Wong, 2023]{iu2023chatgpt}
Iu, K.~Y. and Wong, V. M.-Y. (2023).
\newblock {ChatGPT by OpenAI}: The end of litigation lawyers?
\newblock {\em Available at SSRN 4339839}.

\bibitem[Jacob et~al., 2024]{jacob2024modeling}
Jacob, A.~P., Gupta, A., and Andreas, J. (2024).
\newblock Modeling boundedly rational agents with latent inference budgets.
\newblock In {\em The Twelfth International Conference on Learning
  Representations}.

\bibitem[Jara-Ettinger and Dunham, 2024]{jara2024institutional}
Jara-Ettinger, J. and Dunham, Y. (2024).
\newblock The institutional stance.
\newblock {\em PsyArXiv}.

\bibitem[Jara-Ettinger et~al., 2020]{jara2020naive}
Jara-Ettinger, J., Schulz, L.~E., and Tenenbaum, J.~B. (2020).
\newblock The naive utility calculus as a unified, quantitative framework for
  action understanding.
\newblock {\em Cognitive Psychology}, 123:101334.

\bibitem[Jarrett et~al., 2021]{jarrett2021inverse}
Jarrett, D., H{\"u}y{\"u}k, A., and Van Der~Schaar, M. (2021).
\newblock Inverse decision modeling: Learning interpretable representations of
  behavior.
\newblock In {\em International Conference on Machine Learning}, pages
  4755--4771. PMLR.

\bibitem[Jarrett et~al., 2023]{jarrett2023language}
Jarrett, D., Pislar, M., Bakker, M.~A., Tessler, M.~H., Koster, R., Balaguer,
  J., Elie, R., Summerfield, C., and Tacchetti, A. (2023).
\newblock Language agents as digital representatives in collective
  decision-making.
\newblock In {\em NeurIPS 2023 Foundation Models for Decision Making Workshop}.

\bibitem[Jaynes, 1968]{jaynes1968prior}
Jaynes, E.~T. (1968).
\newblock Prior probabilities.
\newblock {\em IEEE Transactions on Systems Science and Cybernetics},
  4(3):227--241.

\bibitem[Jeffrey, 1991]{jeffrey1991logic}
Jeffrey, R.~C. (1991).
\newblock {\em The {L}ogic of {D}ecision}.
\newblock Chicago University Press, 2nd edition.

\bibitem[Jeon et~al., 2020]{jeon2020reward-rational}
Jeon, H.~J., Milli, S., and Dragan, A. (2020).
\newblock Reward-rational (implicit) choice: {A} unifying formalism for reward
  learning.
\newblock {\em Advances in Neural Information Processing Systems},
  33:4415--4426.

\bibitem[Jiang et~al., 2021]{jiang2021can}
Jiang, L., Hwang, J.~D., Bhagavatula, C., Bras, R.~L., Liang, J., Dodge, J.,
  Sakaguchi, K., Forbes, M., Borchardt, J., Gabriel, S., et~al. (2021).
\newblock Can machines learn morality? {T}he {D}elphi experiment.
\newblock {\em arXiv preprint arXiv:2110.07574}.

\bibitem[Jin et~al., 2022]{jin2022make}
Jin, Z., Levine, S., Gonzalez~Adauto, F., Kamal, O., Sap, M., Sachan, M.,
  Mihalcea, R., Tenenbaum, J., and Sch{\"o}lkopf, B. (2022).
\newblock When to make exceptions: Exploring language models as accounts of
  human moral judgment.
\newblock {\em Advances in Neural Information Processing Systems},
  35:28458--28473.

\bibitem[Jordan, 1982]{jordan1982competitive}
Jordan, J.~S. (1982).
\newblock The competitive allocation process is informationally efficient
  uniquely.
\newblock {\em Journal of Economic Theory}, 28(1):1--18.

\bibitem[Kahneman and Riis, 2005]{kahneman2005living}
Kahneman, D. and Riis, J. (2005).
\newblock Living, and thinking about it: Two perspectives on life.
\newblock {\em The Science of Well-Being}, 1:285--304.

\bibitem[Kahneman and Tversky, 1979]{kahneman1979prospect}
Kahneman, D. and Tversky, A. (1979).
\newblock Prospect theory: {An} analysis of decision under risk.
\newblock {\em Econometrica}, 47.

\bibitem[Kalai and Smorodinsky, 1975]{kalai1975other}
Kalai, E. and Smorodinsky, M. (1975).
\newblock Other solutions to {N}ash's bargaining problem.
\newblock {\em Econometrica: Journal of the Econometric Society}, pages
  513--518.

\bibitem[Kasenberg et~al., 2018]{kasenberg2018norms}
Kasenberg, D., Arnold, T., and Scheutz, M. (2018).
\newblock Norms, rewards, and the intentional stance: Comparing machine
  learning approaches to ethical training.
\newblock In {\em Proceedings of the 2018 AAAI/ACM Conference on AI, Ethics,
  and Society}, pages 184--190.

\bibitem[Kasirzadeh and Gabriel, 2023]{kasirzadeh2023conversation}
Kasirzadeh, A. and Gabriel, I. (2023).
\newblock In {C}onversation with {A}rtificial {I}ntelligence: {A}ligning
  language models with human values.
\newblock {\em Philosophy \& Technology}, 36(2):27.

\bibitem[Keramati et~al., 2016]{keramati2016adaptive}
Keramati, M., Smittenaar, P., Dolan, R.~J., and Dayan, P. (2016).
\newblock Adaptive integration of habits into depth-limited planning defines a
  habitual-goal--directed spectrum.
\newblock {\em Proceedings of the National Academy of Sciences},
  113(45):12868--12873.

\bibitem[Kim et~al., 2023]{kim2023fantom}
Kim, H., Sclar, M., Zhou, X., Bras, R., Kim, G., Choi, Y., and Sap, M. (2023).
\newblock {FANToM}: {A} benchmark for stress-testing machine theory of mind in
  interactions.
\newblock In {\em Proceedings of the 2023 Conference on Empirical Methods in
  Natural Language Processing}, pages 14397--14413.

\bibitem[Kim et~al., 2021]{kim2021reward}
Kim, K., Garg, S., Shiragur, K., and Ermon, S. (2021).
\newblock Reward identification in inverse reinforcement learning.
\newblock In {\em International Conference on Machine Learning}, pages
  5496--5505. PMLR.

\bibitem[Kim et~al., 2018]{kim2018computational}
Kim, R., Kleiman-Weiner, M., Abeliuk, A., Awad, E., Dsouza, S., Tenenbaum,
  J.~B., and Rahwan, I. (2018).
\newblock A computational model of commonsense moral decision making.
\newblock In {\em Proceedings of the 2018 AAAI/ACM Conference on AI, Ethics,
  and Society}, pages 197--203.

\bibitem[Kirk et~al., 2024]{kirk2024benefits}
Kirk, H.~R., Vidgen, B., R{\"o}ttger, P., and Hale, S.~A. (2024).
\newblock The benefits, risks and bounds of personalizing the alignment of
  large language models to individuals.
\newblock {\em Nature Machine Intelligence}, pages 1--10.

\bibitem[Kleiman-Weiner et~al., 2017]{kleiman2017learning}
Kleiman-Weiner, M., Saxe, R., and Tenenbaum, J.~B. (2017).
\newblock Learning a commonsense moral theory.
\newblock {\em Cognition}, 167:107--123.

\bibitem[Klingefjord et~al., 2024]{klingefjord2024human}
Klingefjord, O., Lowe, R., and Edelman, J. (2024).
\newblock What are human values, and how do we align {AI} to them?
\newblock {\em arXiv preprint arXiv:2404.10636}.

\bibitem[Knox et~al., 2024a]{knox2024learning}
Knox, W.~B., Hatgis-Kessell, S., Adalgeirsson, S.~O., Booth, S., Dragan, A.,
  Stone, P., and Niekum, S. (2024a).
\newblock Learning optimal advantage from preferences and mistaking it for
  reward.
\newblock In {\em Proceedings of the AAAI Conference on Artificial
  Intelligence}, volume~38, pages 10066--10073.

\bibitem[Knox et~al., 2024b]{knox2024models}
Knox, W.~B., Hatgis-Kessell, S., Booth, S., Niekum, S., Stone, P., and Allievi,
  A.~G. (2024b).
\newblock Models of human preference for learning reward functions.
\newblock {\em Transactions on Machine Learning Research}.

\bibitem[Knox and Stone, 2011]{knox2011augmenting}
Knox, W.~B. and Stone, P. (2011).
\newblock Augmenting reinforcement learning with human feedback.
\newblock In {\em {ICML} 2011 {Workshop} on {New} {Developments} in {Imitation}
  {Learning} ({July} 2011)}, volume 855, page~3.

\bibitem[Korinek and Balwit, 2022]{korinek2022aligned}
Korinek, A. and Balwit, A. (2022).
\newblock Aligned with whom? {D}irect and social goals for {AI} systems.
\newblock Technical report, National Bureau of Economic Research.

\bibitem[Korsgaard, 1989]{korsgaard1989personal}
Korsgaard, C.~M. (1989).
\newblock Personal identity and the unity of agency: A {Kantian} response to
  {Parfit}.
\newblock {\em Philosophy \& Public Affairs}, pages 101--132.

\bibitem[Krakovna and Kramar, 2023]{krakovna2023power}
Krakovna, V. and Kramar, J. (2023).
\newblock Power-seeking can be probable and predictive for trained agents.
\newblock {\em arXiv preprint arXiv:2304.06528}.

\bibitem[Kwon et~al., 2023a]{kwon2023neuro}
Kwon, J., Levine, S., and Tenenbaum, J.~B. (2023a).
\newblock Neuro-symbolic models of human moral judgment: {LLMs} as automatic
  feature extractors.
\newblock {\em ICML 2023 Workshop on the Challenges of Deploying Generative
  AI}.

\bibitem[Kwon et~al., 2023b]{kwon2023not}
Kwon, J., Zhi-Xuan, T., Tenenbaum, J., and Levine, S. (2023b).
\newblock When it is not out of line to get out of line: The role of
  universalization and outcome-based reasoning in rule-breaking judgments.
\newblock In {\em Proceedings of the Annual Meeting of the Cognitive Science
  Society}, volume~45.

\bibitem[Laibson and Yariv, 2007]{laibson2007safety}
Laibson, D. and Yariv, L. (2007).
\newblock {Safety in Markets: An Impossibility Theorem for Dutch Books}.
\newblock Working Papers 2007-5, Princeton University. Economics Department.

\bibitem[Laidlaw and Dragan, 2022]{laidlaw2022boltzmann}
Laidlaw, C. and Dragan, A. (2022).
\newblock The {B}oltzmann policy distribution: Accounting for systematic
  suboptimality in human models.
\newblock In {\em International Conference on Learning Representations}.

\bibitem[Lam et~al., 2022]{lam2022end}
Lam, M.~S., Gordon, M.~L., Metaxa, D., Hancock, J.~T., Landay, J.~A., and
  Bernstein, M.~S. (2022).
\newblock End-user audits: A system empowering communities to lead large-scale
  investigations of harmful algorithmic behavior.
\newblock {\em Proceedings of the ACM on Human-Computer Interaction},
  6(CSCW2):1--34.

\bibitem[Lambert and Calandra, 2023]{lambert2023alignment}
Lambert, N. and Calandra, R. (2023).
\newblock {The Alignment Ceiling}: {Objective Mismatch in Reinforcement
  Learning from Human Feedback}.
\newblock {\em arXiv preprint arXiv:2311.00168}.

\bibitem[Lambert et~al., 2023]{lambert2023entangled}
Lambert, N., Gilbert, T.~K., and Zick, T. (2023).
\newblock {Entangled Preferences}: {The History and Risks of Reinforcement
  Learning and Human Feedback}.
\newblock {\em arXiv preprint arXiv:2310.13595}.

\bibitem[Lazar, 2024]{lazar2024legitimacy}
Lazar, S. (2024).
\newblock Legitimacy, authority, and democratic duties of explanation.
\newblock {\em Oxford Studies in Political Philosophy Volume 10}, page~28.

\bibitem[Lazar and Nelson, 2023]{lazar2023ai}
Lazar, S. and Nelson, A. (2023).
\newblock {AI} safety on whose terms?
\newblock {\em Science}, 381(6654):138--138.

\bibitem[Leben, 2017]{leben2017rawlsian}
Leben, D. (2017).
\newblock A {Rawlsian} algorithm for autonomous vehicles.
\newblock {\em Ethics and Information Technology}, 19(2):107--115.

\bibitem[Leike et~al., 2018]{leike2018scalable}
Leike, J., Krueger, D., Everitt, T., Martic, M., Maini, V., and Legg, S.
  (2018).
\newblock Scalable agent alignment via reward modeling: {A} research direction.
\newblock {\em arXiv preprint arXiv:1811.07871}.

\bibitem[Leshinskaya et~al., 2023]{leshinskaya2023value}
Leshinskaya, A., San~Franscisco, C., and Chakroff, A. (2023).
\newblock Value as semantics: Representations of human moral and hedonic value
  in large language models.
\newblock {\em NeurIPS 2023 Workshop: AI meets Moral Philosophy and Moral
  Psychology}.

\bibitem[Levine et~al., 2023]{levine2023resource}
Levine, S., Chater, N., Tenenbaum, J., and Cushman, F. (2023).
\newblock Resource-rational contractualism: A triple theory of moral cognition.
\newblock {\em PsyArXiv}.

\bibitem[Levine et~al., 2024]{levine2024rules}
Levine, S., Kleiman-Weiner, M., Chater, N., Cushman, F., and Tenenbaum, J.~B.
  (2024).
\newblock When rules are over-ruled: Virtual bargaining as a contractualist
  method of moral judgment.
\newblock {\em Cognition}, 250:105790.

\bibitem[Levine et~al., 2020]{levine2020logic}
Levine, S., Kleiman-Weiner, M., Schulz, L., Tenenbaum, J., and Cushman, F.
  (2020).
\newblock The logic of universalization guides moral judgment.
\newblock {\em Proceedings of the National Academy of Sciences},
  117(42):26158--26169.

\bibitem[Lewis et~al., 2014]{lewis2014computational}
Lewis, R.~L., Howes, A., and Singh, S. (2014).
\newblock Computational rationality: Linking mechanism and behavior through
  bounded utility maximization.
\newblock {\em Topics in Cognitive Science}, 6(2):279--311.

\bibitem[Li et~al., 2010]{li2010contextual}
Li, L., Chu, W., Langford, J., and Schapire, R.~E. (2010).
\newblock A contextual-bandit approach to personalized news article
  recommendation.
\newblock In {\em Proceedings of the 19th International Conference on World
  Wide Web}, pages 661--670.

\bibitem[Lichtenstein and Slovic, 2006]{lichtenstein2006construction}
Lichtenstein, S. and Slovic, P. (2006).
\newblock {\em The {Construction} of {Preference}}.
\newblock Cambridge University Press.

\bibitem[Lieder and Griffiths, 2020]{lieder2020resource}
Lieder, F. and Griffiths, T.~L. (2020).
\newblock Resource-rational analysis: Understanding human cognition as the
  optimal use of limited computational resources.
\newblock {\em Behavioral and Brain Sciences}, 43:e1.

\bibitem[Lieder et~al., 2018]{lieder2018overrepresentation}
Lieder, F., Griffiths, T.~L., and Hsu, M. (2018).
\newblock Overrepresentation of extreme events in decision making reflects
  rational use of cognitive resources.
\newblock {\em Psychological Review}, 125(1):1.

\bibitem[Lin et~al., 2022]{lin2022inferring}
Lin, J., Fried, D., Klein, D., and Dragan, A. (2022).
\newblock Inferring rewards from language in context.
\newblock In {\em Proceedings of the 60th Annual Meeting of the Association for
  Computational Linguistics}, pages 8546--8560.

\bibitem[Liu, 2011]{liu2011reasoning}
Liu, F. (2011).
\newblock {\em Reasoning about Preference Dynamics}, volume 354.
\newblock Springer Science \& Business Media.

\bibitem[Loewenstein and Angner, 2003]{loewenstein2003predicting}
Loewenstein, G. and Angner, E. (2003).
\newblock Predicting and indulging changing preferences.
\newblock In {\em Time and Decision: Economic and Psychological Perspectives on
  Intertemporal Choice}, pages 351--391. Russell Sage Foundation, New York, NY,
  US.

\bibitem[Logins, 2022]{logins2022normative}
Logins, A. (2022).
\newblock {\em Normative Reasons: Between Reasoning and Explanation}.
\newblock Cambridge University Press.

\bibitem[Lohr, 2023]{lohr2023ai}
Lohr, S. (2023).
\newblock {A.I.} is coming for lawyers, again.
\newblock {\em The New York Times}.

\bibitem[London and Heidari, 2024]{london2024beneficent}
London, A.~J. and Heidari, H. (2024).
\newblock Beneficent intelligence: a capability approach to modeling benefit,
  assistance, and associated moral failures through ai systems.
\newblock {\em Minds and Machines}, 34(4):41.

\bibitem[Luce, 1979]{luce1979individual}
Luce, R.~D. (1979).
\newblock {\em Individual Choice Behavior: A Theoretical Analysis}.
\newblock Greenwood Press, Westport, Conn.

\bibitem[Lukacs and Livingstone, 1972]{lukacs1972history}
Lukacs, G. and Livingstone, R. (1972).
\newblock {\em History and Class Consciousness: Studies in Marxist Dialectics}.
\newblock MIT Press.

\bibitem[Lumer et~al., 2005]{lumer2005prioritarian}
Lumer, C. et~al. (2005).
\newblock Prioritarian welfare functions: An elaboration and justification.

\bibitem[Mahowald, 2023]{mahowald2023discerning}
Mahowald, K. (2023).
\newblock A discerning several thousand judgments: {GPT-3} rates the article+
  adjective+ numeral+ noun construction.
\newblock In {\em Proceedings of the 17th Conference of the European Chapter of
  the Association for Computational Linguistics}, pages 265--273.

\bibitem[Mahowald et~al., 2024]{mahowald2024dissociating}
Mahowald, K., Ivanova, A.~A., Blank, I.~A., Kanwisher, N., Tenenbaum, J.~B.,
  and Fedorenko, E. (2024).
\newblock Dissociating language and thought in large language models.
\newblock {\em Trends in Cognitive Sciences}.

\bibitem[McInerney et~al., 2018]{mcinerney2018explore}
McInerney, J., Lacker, B., Hansen, S., Higley, K., Bouchard, H., Gruson, A.,
  and Mehrotra, R. (2018).
\newblock Explore, exploit, and explain: Personalizing explainable
  recommendations with bandits.
\newblock In {\em Proceedings of the 12th ACM Conference on Recommender
  Systems}, RecSys '18, page 31–39, New York, NY, USA. Association for
  Computing Machinery.

\bibitem[Mercier and Sperber, 2011]{mercier2011humans}
Mercier, H. and Sperber, D. (2011).
\newblock Why do humans reason? {A}rguments for an argumentative theory.
\newblock {\em Behavioral and Brain Sciences}, 34(2):57--74.

\bibitem[Mercier and Sperber, 2017]{mercier2017enigma}
Mercier, H. and Sperber, D. (2017).
\newblock {\em The {E}nigma of {R}eason}.
\newblock Harvard University Press.

\bibitem[Merrill et~al., 2024]{merrill2024can}
Merrill, W., Wu, Z., Naka, N., Kim, Y., and Linzen, T. (2024).
\newblock Can you learn semantics through next-word prediction? {T}he case of
  entailment.
\newblock In {\em Findings of the Association for Computational Linguistics ACL
  2024}. Association for Computational Linguistics.

\bibitem[Mill, 1859]{mill1859liberty}
Mill, J. (1859).
\newblock {\em On Liberty}.
\newblock J. W. Parker and Son.

\bibitem[Mishra, 2023]{mishra2023ai}
Mishra, A. (2023).
\newblock {AI} alignment and social choice: Fundamental limitations and policy
  implications.
\newblock {\em arXiv preprint arXiv:2310.16048}.

\bibitem[Mishra, 2014]{mishra2014decision}
Mishra, S. (2014).
\newblock Decision-making under risk: Integrating perspectives from biology,
  economics, and psychology.
\newblock {\em Personality and Social Psychology Review}, 18(3):280--307.

\bibitem[Misyak et~al., 2014]{misyak2014unwritten}
Misyak, J.~B., Melkonyan, T., Zeitoun, H., and Chater, N. (2014).
\newblock Unwritten rules: {V}irtual bargaining underpins social interaction,
  culture, and society.
\newblock {\em Trends in cognitive sciences}, 18(10):512--519.

\bibitem[Modgil, 2009]{modgil2009reasoning}
Modgil, S. (2009).
\newblock Reasoning about preferences in argumentation frameworks.
\newblock {\em Artificial Intelligence}, 173(9-10):901--934.

\bibitem[Molinaro and Collins, 2023]{molinaro2023goal}
Molinaro, G. and Collins, A.~G. (2023).
\newblock A goal-centric outlook on learning.
\newblock {\em Trends in Cognitive Sciences}.

\bibitem[Momennejad et~al., 2024]{momennejad2024evaluating}
Momennejad, I., Hasanbeig, H., Vieira~Frujeri, F., Sharma, H., Jojic, N.,
  Palangi, H., Ness, R., and Larson, J. (2024).
\newblock Evaluating cognitive maps and planning in large language models with
  {CogEval}.
\newblock {\em Advances in Neural Information Processing Systems}, 36.

\bibitem[Moskovitz et~al., 2024]{moskovitz2024confronting}
Moskovitz, T., Singh, A.~K., Strouse, D., Sandholm, T., Salakhutdinov, R.,
  Dragan, A.~D., and McAleer, S. (2024).
\newblock Confronting reward model overoptimization with constrained {RLHF}.
\newblock In {\em The Twelfth International Conference on Learning
  Representations}.

\bibitem[Mount and Reiter, 1974]{mount1974informational}
Mount, K. and Reiter, S. (1974).
\newblock The informational size of message spaces.
\newblock {\em Journal of Economic Theory}, 8(2):161--192.

\bibitem[Murphy, 2006]{murphy2006cantor}
Murphy, R. (2006).
\newblock Cantor’s diagonal argument: An extension to the socialist
  calculation debate.
\newblock {\em The Quarterly Journal of Austrian Economics}, 9(2):3--11.

\bibitem[Ng et~al., 1999]{ng1999policy}
Ng, A.~Y., Harada, D., and Russell, S.~J. (1999).
\newblock Policy invariance under reward transformations: Theory and
  application to reward shaping.
\newblock In {\em Proceedings of the Sixteenth International Conference on
  Machine Learning}, pages 278--287.

\bibitem[Ng and Russell, 2000]{ng2000algorithms}
Ng, A.~Y. and Russell, S.~J. (2000).
\newblock Algorithms for inverse reinforcement learning.
\newblock In {\em Proceedings of the Seventeenth International Conference on
  Machine Learning}, pages 663--670.

\bibitem[Ng and Subrahmanian, 1992]{ng1992probabilistic}
Ng, R. and Subrahmanian, V.~S. (1992).
\newblock Probabilistic logic programming.
\newblock {\em Information and computation}, 101(2):150--201.

\bibitem[Ng, 1997]{ng1997case}
Ng, Y.-K. (1997).
\newblock A case for happiness, cardinalism, and interpersonal comparability.
\newblock {\em The Economic Journal}, 107(445):1848--1858.

\bibitem[Ngo, 2019]{ngo2019coherent}
Ngo, R. (2019).
\newblock Coherent behaviour in the real world is an incoherent concept.
\newblock {\em AI Alignment Forum}.
\newblock \url{https://www.alignmentforum.org/posts/vphFJzK3mWA4PJKAg}.

\bibitem[Ngo et~al., 2022]{ngo2022alignment}
Ngo, R., Chan, L., and Mindermann, S. (2022).
\newblock The alignment problem from a deep learning perspective.
\newblock {\em arXiv preprint arXiv:2209.00626}.

\bibitem[Nielsen and Rigotti, 2023]{nielsen2023revealed}
Nielsen, K. and Rigotti, L. (2023).
\newblock Revealed incomplete preferences.
\newblock {\em Available at SSRN 4622145}.

\bibitem[Nussbaum, 2001]{nussbaum2001symposium}
Nussbaum, M.~C. (2001).
\newblock Symposium on {A}martya {S}en's philosophy: {A}daptive preferences and
  women's options.
\newblock {\em Economics \& Philosophy}, 17(1):67--88.

\bibitem[Okidegbe, 2021]{okidegbe2021discredited}
Okidegbe, N. (2021).
\newblock Discredited data.
\newblock {\em Cornell L. Rev.}, 107:2007.

\bibitem[Oldenburg and Zhi-Xuan, 2024]{oldenburg2024learning}
Oldenburg, N. and Zhi-Xuan, T. (2024).
\newblock Learning and sustaining shared normative systems via {Bayesian} rule
  induction in {Markov Games}.
\newblock In {\em Proceedings of the 23rd International Conference on
  Autonomous Agents and Multiagent Systems}.

\bibitem[Omohundro, 2007]{omohundro2007nature}
Omohundro, S.~M. (2007).
\newblock The nature of self-improving artificial intelligence.
\newblock {\em Singularity Summit}, 2008.
\newblock Publisher: Citeseer.

\bibitem[Omohundro, 2008]{omohundro2008basic}
Omohundro, S.~M. (2008).
\newblock The basic {AI} drives.
\newblock In {\em AGI}, volume 171, pages 483--492.

\bibitem[Ortega and Braun, 2013]{ortega2013thermodynamics}
Ortega, P.~A. and Braun, D.~A. (2013).
\newblock Thermodynamics as a theory of decision-making with
  information-processing costs.
\newblock {\em Proceedings of the Royal Society A: Mathematical, Physical and
  Engineering Sciences}, 469(2153):20120683.

\bibitem[Oulasvirta et~al., 2022]{oulasvirta2022computational}
Oulasvirta, A., Jokinen, J. P.~P., and Howes, A. (2022).
\newblock Computational {Rationality} as a {Theory} of {Interaction}.
\newblock In {\em Proceedings of the 2022 {CHI} {Conference} on {Human}
  {Factors} in {Computing} {Systems}}, {CHI} '22, pages 1--14, New York, NY,
  USA. Association for Computing Machinery.

\bibitem[Ouyang et~al., 2022]{ouyang2022training}
Ouyang, L., Wu, J., Jiang, X., Almeida, D., Wainwright, C.~L., Mishkin, P.,
  Zhang, C., Agarwal, S., Slama, K., Ray, A., Schulman, J., Hilton, J., Kelton,
  F., Miller, L., Simens, M., Askell, A., Welinder, P., Christiano, P., Leike,
  J., and Lowe, R. (2022).
\newblock Training language models to follow instructions with human feedback.
\newblock {\em Advances in Neural Information Processing Systems},
  35:27730--27744.

\bibitem[Ovadya, 2023]{ovadya2023reimagining}
Ovadya, A. (2023).
\newblock Reimagining democracy for {AI}.
\newblock {\em Journal of Democracy}, 34(4):162--170.

\bibitem[Papadimitriou and Tsitsiklis, 1987]{papadimitriou1987complexity}
Papadimitriou, C.~H. and Tsitsiklis, J.~N. (1987).
\newblock The {Complexity} of {Markov Decision Processes}.
\newblock {\em Mathematics of Operations Research}, 12(3):441--450.

\bibitem[Parfit, 2011]{parfit2011matters}
Parfit, D. (2011).
\newblock {\em On What Matters}, volume~1.
\newblock Oxford University Press, USA.

\bibitem[Parfit, 2018]{parfit2018rationality}
Parfit, D. (2018).
\newblock Rationality and reasons.
\newblock In {\em Exploring Practical Philosophy: From Action to Values}, pages
  17--39. Routledge.

\bibitem[Parisi et~al., 2022]{parisi2022talm}
Parisi, A., Zhao, Y., and Fiedel, N. (2022).
\newblock {TALM}: Tool augmented language models.
\newblock {\em arXiv preprint arXiv:2205.12255}.

\bibitem[Paul, 2014]{paul2014transformative}
Paul, L.~A. (2014).
\newblock {\em Transformative Experience}.
\newblock OUP Oxford.

\bibitem[Pennycook et~al., 2021]{pennycook2021shifting}
Pennycook, G., Epstein, Z., Mosleh, M., Arechar, A.~A., Eckles, D., and Rand,
  D.~G. (2021).
\newblock Shifting attention to accuracy can reduce misinformation online.
\newblock {\em Nature}, 592(7855):590--595.

\bibitem[Petersen, 2023]{petersen2023invulnerable}
Petersen, S. (2023).
\newblock Invulnerable incomplete preferences: A formal statement.
\newblock {\em AI Alignment Forum}.
\newblock \url{https://www.alignmentforum.org/posts/sHGxvJrBag7nhTQvb}.

\bibitem[Pettigrew, 2019]{pettigrew2019choosing}
Pettigrew, R. (2019).
\newblock {\em Choosing for Changing Selves}.
\newblock Oxford University Press.

\bibitem[Piantadosi and Hill, 2022]{piantadosi2022meaning}
Piantadosi, S.~T. and Hill, F. (2022).
\newblock Meaning without reference in large language models.
\newblock {\em arXiv preprint arXiv:2208.02957}.

\bibitem[Piantadosi and Jacobs, 2016]{piantadosi2016four}
Piantadosi, S.~T. and Jacobs, R.~A. (2016).
\newblock Four problems solved by the probabilistic language of thought.
\newblock {\em Current Directions in Psychological Science}, 25(1):54--59.

\bibitem[Pitis et~al., 2024]{pitis2024improving}
Pitis, S., Xiao, Z., Roux, N.~L., and Sordoni, A. (2024).
\newblock Improving context-aware preference modeling for language models.
\newblock {\em arXiv preprint arXiv:2407.14916}.

\bibitem[Prunkl and Whittlestone, 2020]{prunkl2020beyond}
Prunkl, C. and Whittlestone, J. (2020).
\newblock Beyond near-and long-term: Towards a clearer account of research
  priorities in {AI} ethics and society.
\newblock In {\em Proceedings of the AAAI/ACM Conference on AI, Ethics, and
  Society}, pages 138--143.

\bibitem[Quilty-Dunn et~al., 2023]{quilty2023best}
Quilty-Dunn, J., Porot, N., and Mandelbaum, E. (2023).
\newblock The best game in town: The reemergence of the language-of-thought
  hypothesis across the cognitive sciences.
\newblock {\em Behavioral and Brain Sciences}, 46:e261.

\bibitem[Rafailov et~al., 2024]{rafailov2024direct}
Rafailov, R., Sharma, A., Mitchell, E., Manning, C.~D., Ermon, S., and Finn, C.
  (2024).
\newblock Direct preference optimization: Your language model is secretly a
  reward model.
\newblock {\em Advances in Neural Information Processing Systems}, 36.

\bibitem[Rahwan et~al., 2003]{rahwan2003argumentation}
Rahwan, I., Ramchurn, S.~D., Jennings, N.~R., McBurney, P., Parsons, S., and
  Sonenberg, L. (2003).
\newblock Argumentation-based negotiation.
\newblock {\em The Knowledge Engineering Review}, 18(4):343--375.

\bibitem[Railton, 1993]{railton1993alienation}
Railton, P. (1993).
\newblock {\em Alienation, Consequentialism, and the Demands of Morality},
  pages 211--244.
\newblock Cornell University Press, Ithaca, NY.

\bibitem[Ramesh et~al., 2024]{ramesh2024compositional}
Ramesh, R., Lubana, E.~S., Khona, M., Dick, R.~P., and Tanaka, H. (2024).
\newblock Compositional capabilities of autoregressive transformers: A study on
  synthetic, interpretable tasks.
\newblock In {\em Forty-First International Conference on Machine Learning}.

\bibitem[Rawls, 1971]{rawls1971theory}
Rawls, J. (1971).
\newblock {\em A {Theory} of {Justice}: {Original} {Edition}}.
\newblock Harvard University Press.

\bibitem[Rawls, 1993]{rawls1993political}
Rawls, J. (1993).
\newblock {\em Political Liberalism}.
\newblock Columbia University Press.

\bibitem[Raz, 1999]{raz1999engaging}
Raz, J. (1999).
\newblock {\em Engaging Reason: On the Theory of Value and Action}.
\newblock Oxford University Press.

\bibitem[Reddy et~al., 2018]{reddy2018you}
Reddy, S., Dragan, A., and Levine, S. (2018).
\newblock Where do you think you're going?: Inferring beliefs about dynamics
  from behavior.
\newblock {\em Advances in Neural Information Processing Systems}, 31.

\bibitem[Richardson et~al., 2019]{richardson2019dirty}
Richardson, R., Schultz, J.~M., and Crawford, K. (2019).
\newblock Dirty data, bad predictions: How civil rights violations impact
  police data, predictive policing systems, and justice.
\newblock {\em NYUL Rev. Online}, 94:15.

\bibitem[Rubinstein and Salant, 2012]{rubinstein2012eliciting}
Rubinstein, A. and Salant, Y. (2012).
\newblock Eliciting welfare preferences from behavioural data sets.
\newblock {\em The Review of Economic Studies}, 79(1):375--387.

\bibitem[Russell, 2019]{russell2019human}
Russell, S.~J. (2019).
\newblock {\em Human Compatible: Artificial Intelligence and The Problem of
  Control}.
\newblock Allen Lane, an imprint of Penguin Books, London.

\bibitem[Russell and Subramanian, 1994]{russell1994provably}
Russell, S.~J. and Subramanian, D. (1994).
\newblock Provably bounded-optimal agents.
\newblock {\em Journal of Artificial Intelligence Research}, 2:575--609.

\bibitem[Rust, 1996]{rust1996dealing}
Rust, J.~P. (1996).
\newblock Dealing with the complexity of economic calculations.
\newblock {\em Available at SSRN 40780}.

\bibitem[Samuelson, 1938]{samuelson1938note}
Samuelson, P.~A. (1938).
\newblock A {Note} on the {Pure} {Theory} of {Consumer}'s {Behaviour}.
\newblock {\em Economica}, 5(17):61.

\bibitem[Saunders et~al., 2022]{saunders2022self}
Saunders, W., Yeh, C., Wu, J., Bills, S., Ouyang, L., Ward, J., and Leike, J.
  (2022).
\newblock Self-critiquing models for assisting human evaluators.
\newblock {\em arXiv preprint arXiv:2206.05802}.

\bibitem[Savage, 1972]{savage1972foundations}
Savage, L.~J. (1972).
\newblock {\em The Foundations of Statistics}.
\newblock Dover Publications, New York, 2d rev. ed edition.

\bibitem[Scanlon, 2000]{scanlon2000what}
Scanlon, T. (2000).
\newblock {\em What We Owe To Each Other}.
\newblock The Belknap Press of Harvard University Press, Cambridge,
  Massachusetts London, England.

\bibitem[Schechtman, 2014]{schechtman2014staying}
Schechtman, M. (2014).
\newblock {\em Staying Alive: Personal Identity, Practical Concerns, and the
  Unity of a Life}.
\newblock OUP Oxford.

\bibitem[Schroeder, 2004]{schroeder2004three}
Schroeder, T. (2004).
\newblock {\em Three Faces of Desire}.
\newblock Oxford University Press.

\bibitem[Scott, 1998]{scott1998seeing}
Scott, J.~C. (1998).
\newblock {\em Seeing Like A State: How Certain Schemes to Improve the Human
  Condition Have Failed}.
\newblock Yale University Press.

\bibitem[Sen, 1970a]{sen1970collective}
Sen, A. (1970a).
\newblock {\em Collective Choice and Social Welfare}.
\newblock Harvard University Press.

\bibitem[Sen, 1970b]{sen1970interpersonal}
Sen, A. (1970b).
\newblock Interpersonal aggregation and partial comparability.
\newblock {\em Econometrica: Journal of the Econometric Society}, pages
  393--409.

\bibitem[Sen et~al., 1999]{sen1999commodities}
Sen, A. et~al. (1999).
\newblock Commodities and capabilities.
\newblock {\em OUP Catalogue}.

\bibitem[Shah et~al., 2018]{shah2018bayesian}
Shah, A., Kamath, P., Shah, J.~A., and Li, S. (2018).
\newblock Bayesian inference of temporal task specifications from
  demonstrations.
\newblock {\em Advances in Neural Information Processing Systems}, 31.

\bibitem[Shah, 2018]{shah2018coherence}
Shah, R. (2018).
\newblock Coherence arguments do not entail goal-directed behavior.
\newblock {\em AI Alignment Forum}.
\newblock \url{https://www.alignmentforum.org/posts/NxF5G6CJiof6cemTw}.

\bibitem[Shah et~al., 2019]{shah2019feasibility}
Shah, R., Gundotra, N., Abbeel, P., and Dragan, A. (2019).
\newblock On the feasibility of learning, rather than assuming, human biases
  for reward inference.
\newblock In {\em International Conference on Machine Learning}, pages
  5670--5679. PMLR.

\bibitem[Siddarth et~al., 2022]{siddarth2022how}
Siddarth, D., Acemoglu, D., Allen, D., Crawford, K., Evans, J., Jordan, M., and
  Weyl, E.~G. (2022).
\newblock How {AI} {F}ails {U}s.
\newblock {\em Technology and Democracy Discussion Paper Series}.

\bibitem[Siddarth and Huang, 2023]{siddarth2023whitepaper}
Siddarth, D. and Huang, S. (2023).
\newblock Whitepaper.
\newblock {\em The Collective Intelligence Project}.
\newblock \url{https://cip.org/whitepaper}.

\bibitem[Silver et~al., 2021]{silver2021reward}
Silver, D., Singh, S., Precup, D., and Sutton, R.~S. (2021).
\newblock Reward is enough.
\newblock {\em Artificial Intelligence}, 299:103535.

\bibitem[Simon, 1957]{simon1957behavioral}
Simon, H.~A. (1957).
\newblock A behavioral model of rational choice.
\newblock {\em Models of Man, Social and Rational: Mathematical Essays on
  Rational Human Behavior in a Social Setting}, pages 241--260.

\bibitem[Simon, 1979]{simon1979rational}
Simon, H.~A. (1979).
\newblock Rational decision making in business organizations.
\newblock {\em The American Economic Review}, 69(4):493--513.
\newblock Publisher: JSTOR.

\bibitem[Sims, 2003]{sims2003implications}
Sims, C.~A. (2003).
\newblock Implications of rational inattention.
\newblock {\em Journal of Monetary Economics}, 50(3):665--690.

\bibitem[Singh et~al., 2009]{singh2009rewards}
Singh, S., Lewis, R.~L., and Barto, A.~G. (2009).
\newblock Where do rewards come from.
\newblock In {\em Proceedings of the Annual Conference of the Cognitive Science
  Society}, pages 2601--2606. Cognitive Science Society.

\bibitem[Sinhababu, 2017]{sinhababu2017humean}
Sinhababu, N. (2017).
\newblock {\em Humean Nature: How Desire Explains Action, Thought, and
  Feeling}.
\newblock Oxford University Press.

\bibitem[Siththaranjan et~al., 2024]{siththaranjan2024distributional}
Siththaranjan, A., Laidlaw, C., and Hadfield-Menell, D. (2024).
\newblock Distributional {Preference} {Learning}: {Understanding} and
  {Accounting} for {Hidden} {Context} in {RLHF}.
\newblock In {\em The Twelfth International Conference on Learning
  Representations}.

\bibitem[Skalse et~al., 2023]{skalse2023invariance}
Skalse, J. M.~V., Farrugia-Roberts, M., Russell, S., Abate, A., and Gleave, A.
  (2023).
\newblock Invariance in policy optimisation and partial identifiability in
  reward learning.
\newblock In {\em International Conference on Machine Learning}, pages
  32033--32058. PMLR.

\bibitem[Soares et~al., 2015]{soares2015corrigibility}
Soares, N., Fallenstein, B., Armstrong, S., and Yudkowsky, E. (2015).
\newblock Corrigibility.
\newblock In {\em Workshops at the Twenty-Ninth AAAI Conference on Artificial
  Intelligence}.

\bibitem[Sobel, 2005]{sobel2005interdependent}
Sobel, J. (2005).
\newblock Interdependent preferences and reciprocity.
\newblock {\em Journal of Economic Literature}, 43(2):392--436.

\bibitem[Sorensen et~al., 2024]{sorensen2024roadmap}
Sorensen, T., Moore, J., Fisher, J., Gordon, M., Mireshghallah, N., Rytting,
  C.~M., Ye, A., Jiang, L., Lu, X., Dziri, N., et~al. (2024).
\newblock A roadmap to pluralistic alignment.
\newblock In {\em Proceedings of the 41st International Conference on Machine
  Learning}, pages 46280--46302.

\bibitem[Stark, 1997]{stark1997decision}
Stark, C.~A. (1997).
\newblock Decision procedures, standards of rightness and impartiality.
\newblock {\em Nous}, 31(4):478--495.

\bibitem[Stechly et~al., 2023]{stechly2023gpt}
Stechly, K., Marquez, M., and Kambhampati, S. (2023).
\newblock {GPT-4} doesn't know it's wrong: An analysis of iterative prompting
  for reasoning problems.
\newblock {\em NeurIPS 2023 Workshop on Foundation Models for Decision Making.}

\bibitem[Steele and Stefánsson, 2020]{steele2020decision}
Steele, K. and Stefánsson, H.~O. (2020).
\newblock Decision {Theory}.
\newblock In Zalta, E.~N., editor, {\em The {Stanford} {Encyclopedia} of
  {Philosophy}}. Metaphysics Research Lab, Stanford University, winter 2020
  edition.

\bibitem[Steinhardt, 2017]{steinhardt2017latent}
Steinhardt, J. (2017).
\newblock Latent variables and model misspecification.
\newblock {\em AI Alignment Forum}.

\bibitem[Stone and Mittelstadt, 2024]{stone2024legitimate}
Stone, J. and Mittelstadt, B. (2024).
\newblock Legitimate power, illegitimate automation: The problem of ignoring
  legitimacy in automated decision systems.
\newblock {\em Proceedings of the 2024 ACM Conference on Fairness,
  Accountability, and Transparency (FAccT 2024)}.

\bibitem[Strotz, 1953]{strotz1953cardinal}
Strotz, R.~H. (1953).
\newblock Cardinal utility.
\newblock {\em The American Economic Review}, 43(2):384--397.

\bibitem[Sumers et~al., 2024]{sumers2024cognitive}
Sumers, T., Yao, S., Narasimhan, K., and Griffiths, T. (2024).
\newblock Cognitive architectures for language agents.
\newblock {\em Transactions on Machine Learning Research}.

\bibitem[Suresh et~al., 2024]{suresh2024participation}
Suresh, H., Tseng, E., Young, M., Gray, M., Pierson, E., and Levy, K. (2024).
\newblock Participation in the age of foundation models.
\newblock In {\em The 2024 ACM Conference on Fairness, Accountability, and
  Transparency}, pages 1609--1621.

\bibitem[Sutton and Barto, 2018]{sutton2018reinforcement}
Sutton, R.~S. and Barto, A.~G. (2018).
\newblock {\em Reinforcement Learning: An Introduction}.
\newblock MIT press.

\bibitem[Symons and Alvarado, 2022]{symons2022epistemic}
Symons, J. and Alvarado, R. (2022).
\newblock Epistemic injustice and data science technologies.
\newblock {\em Synthese}, 200(2):87.

\bibitem[Tan and Ong, 2019]{tan2019bayesian}
Tan, Z.-X. and Ong, D.~C. (2019).
\newblock Bayesian inference of social norms as shared constraints on behavior.
\newblock In {\em Proceedings of the Annual Meeting of the Cognitive Science
  Society}, volume~41, pages 2919--2925.

\bibitem[Taylor, 2016]{taylor2016quantilizers}
Taylor, J. (2016).
\newblock Quantilizers: A safer alternative to maximizers for limited
  optimization.
\newblock In {\em AAAI Workshop: AI, Ethics, and Society}.

\bibitem[Tessler et~al., 2024]{tessler2024ai}
Tessler, M.~H., Bakker, M.~A., Jarrett, D., Sheahan, H., Chadwick, M.~J.,
  Koster, R., Evans, G., Campbell-Gillingham, L., Collins, T., Parkes, D.~C.,
  et~al. (2024).
\newblock {AI} can help humans find common ground in democratic deliberation.
\newblock {\em Science}, 386(6719):eadq2852.

\bibitem[Thorburn et~al., 2022]{thorburn2022what}
Thorburn, L., Stray, J., and Bengani, P. (2022).
\newblock What does it mean to give someone what they want? the nature of
  preferences in recommender systems.
\newblock \url{https://medium.com/p/82b5a1559157}.

\bibitem[Thornley, 2023]{thornley2023coherence}
Thornley, E. (2023).
\newblock There are no coherence theorems.
\newblock {\em AI Alignment Forum}.

\bibitem[Thornley, 2024]{thornley2024shutdown}
Thornley, E. (2024).
\newblock The shutdown problem: an {AI} engineering puzzle for decision
  theorists.
\newblock {\em Philosophical Studies}, pages 1--28.

\bibitem[Thornley et~al., 2024]{thornley2024towards}
Thornley, E., Roman, A., Ziakas, C., Ho, L., and Thomson, L. (2024).
\newblock Towards shutdownable agents via stochastic choice.
\newblock {\em arXiv preprint arXiv:2407.00805}.

\bibitem[Toner and McCauley, 2024]{toner2024aifirms}
Toner, H. and McCauley, T. (2024).
\newblock {A}{I} firms mustn’t govern themselves, say ex-members of
  {O}pen{A}{I}’s board.
\newblock {\em The Economist}.

\bibitem[Turner et~al., 2020]{turner2020avoiding}
Turner, A., Ratzlaff, N., and Tadepalli, P. (2020).
\newblock Avoiding side effects in complex environments.
\newblock {\em Advances in Neural Information Processing Systems},
  33:21406--21415.

\bibitem[Turner et~al., 2021]{turner2021optimal}
Turner, A.~M., Smith, L., Shah, R., Critch, A., and Tadepalli, P. (2021).
\newblock Optimal policies tend to seek power.
\newblock In {\em Proceedings of the 35th International Conference on Neural
  Information Processing Systems}, pages 23063--23074.

\bibitem[Tversky and Kahneman, 1992]{tversky1992advances}
Tversky, A. and Kahneman, D. (1992).
\newblock Advances in prospect theory: {Cumulative} representation of
  uncertainty.
\newblock {\em Journal of Risk and Uncertainty}, 5(4):297--323.
\newblock Publisher: Springer.

\bibitem[Ulen, 1999]{ulen1999rational}
Ulen, T.~S. (1999).
\newblock Rational choice theory in law and economics.
\newblock {\em Encyclopedia of Law and Economics}, 1:790--818.

\bibitem[Ullman et~al., 2009]{ullman2009help}
Ullman, T., Baker, C., Macindoe, O., Evans, O., Goodman, N., and Tenenbaum, J.
  (2009).
\newblock Help or hinder: Bayesian models of social goal inference.
\newblock {\em Advances in Neural Information Processing Systems}, 22.

\bibitem[Valmeekam et~al., 2023a]{valmeekam2023can}
Valmeekam, K., Marquez, M., and Kambhampati, S. (2023a).
\newblock Can large language models really improve by self-critiquing their own
  plans?
\newblock {\em NeurIPS 2023 Workshop on Foundation Models for Decision Making.}

\bibitem[Valmeekam et~al., 2023b]{valmeekam2023on}
Valmeekam, K., Marquez, M., Sreedharan, S., and Kambhampati, S. (2023b).
\newblock On the planning abilities of large language models - a critical
  investigation.
\newblock In {\em Thirty-seventh Conference on Neural Information Processing
  Systems}.

\bibitem[Vamplew et~al., 2022]{vamplew2022scalar}
Vamplew, P., Smith, B.~J., Källström, J., Ramos, G., Rădulescu, R., Roijers,
  D.~M., Hayes, C.~F., Heintz, F., Mannion, P., Libin, P. J.~K., Dazeley, R.,
  and Foale, C. (2022).
\newblock Scalar reward is not enough: a response to {Silver}, {Singh},
  {Precup} and {Sutton} (2021).
\newblock {\em Autonomous Agents and Multi-Agent Systems}, 36(2):41.

\bibitem[van~de Meent et~al., 2018]{van2018introduction}
van~de Meent, J.-W., Paige, B., Yang, H., and Wood, F. (2018).
\newblock An introduction to probabilistic programming.
\newblock {\em arXiv preprint arXiv:1809.10756}.

\bibitem[van Rooij, 2008]{van2008tractable}
van Rooij, I. (2008).
\newblock The {T}ractable {C}ognition {T}hesis.
\newblock {\em Cognitive Science}, 32(6):939--984.

\bibitem[Van~Rooij et~al., 2024]{van2024reclaiming}
Van~Rooij, I., Guest, O., Adolfi, F., de~Haan, R., Kolokolova, A., and Rich, P.
  (2024).
\newblock Reclaiming ai as a theoretical tool for cognitive science.
\newblock {\em Computational Brain \& Behavior}, pages 1--21.

\bibitem[van Wynsberghe and Robbins, 2019]{van2019critiquing}
van Wynsberghe, A. and Robbins, S. (2019).
\newblock Critiquing the reasons for making artificial moral agents.
\newblock {\em Science and Engineering Ethics}, 25:719--735.

\bibitem[Verdery, 2005]{verdery2005socialism}
Verdery, K. (2005).
\newblock What was socialism, and why did it fall?
\newblock In {\em The Revolutions of 1989}, pages 73--94. Routledge.

\bibitem[Vineberg, 2011]{vineberg2011dutch}
Vineberg, S. (2011).
\newblock Dutch book arguments.
\newblock In Zalta, E., editor, {\em The {Stanford} Encyclopedia of
  Philosophy}. Metaphysics Research Lab, Stanford University.

\bibitem[von Mises, 1990]{von1990economic}
von Mises, L. (1990).
\newblock {\em Economic Calculation in the Socialist Commonwealth}.
\newblock Ludwig Von Mises Institute, Auburn University.

\bibitem[von Neumann and Morgenstern, 1944]{von_neumann1944theory}
von Neumann, J. and Morgenstern, O. (1944).
\newblock {\em Theory of Games and Economic Behavior}.
\newblock Princeton University Press.

\bibitem[von Widekind, 2008]{von2008evolution}
von Widekind, S. (2008).
\newblock {\em Evolution of Non-Expected Utility Preferences}, volume 606.
\newblock Springer Science \& Business Media.

\bibitem[von Wright, 1951]{von1951deontic}
von Wright, G.~H. (1951).
\newblock Deontic logic.
\newblock {\em Mind}, 60(237):1--15.

\bibitem[von Wright, 1972]{von1972logic}
von Wright, G.~H. (1972).
\newblock The logic of preference reconsidered.
\newblock {\em Theory and Decision}, 3:140--169.

\bibitem[Warren et~al., 2011]{warren2011values}
Warren, C., McGraw, A.~P., and Van~Boven, L. (2011).
\newblock Values and preferences: defining preference construction.
\newblock {\em WIREs Cognitive Science}, 2(2):193--205.

\bibitem[Weber, 1978]{weber1978economy}
Weber, M. (1978).
\newblock {\em Economy and Society: An Outline of Interpretive Sociology}.
\newblock University of California Press.

\bibitem[Weidinger et~al., 2023]{weidinger2023using}
Weidinger, L., McKee, K.~R., Everett, R., Huang, S., Zhu, T.~O., Chadwick,
  M.~J., Summerfield, C., and Gabriel, I. (2023).
\newblock Using the {Veil} of {Ignorance} to align {AI} systems with principles
  of justice.
\newblock {\em Proceedings of the National Academy of Sciences},
  120(18):e2213709120.

\bibitem[Wentworth, 2019]{wentworth2019subagents}
Wentworth, J. (2019).
\newblock Why subagents?
\newblock {\em AI Alignment Forum}.
\newblock \url{https://www.alignmentforum.org/posts/3xF66BNSC5caZuKyC}.

\bibitem[Wentworth, 2023]{wentworth2023subagents}
Wentworth, J. (2023).
\newblock Why not subagents?
\newblock {\em AI Alignment Forum}.
\newblock \url{https://www.alignmentforum.org/posts/bzmLC3J8PsknwRZbr}.

\bibitem[Wheaton, 2023]{wheaton2023deceptive}
Wheaton, D. (2023).
\newblock Deceptive alignment is $<$1\% likely by default.
\newblock {\em Less Wrong}.
\newblock \url{https://www.lesswrong.com/posts/RTkatYxJWvXR4Qbyd}.

\bibitem[Wong et~al., 2023]{wong2023word}
Wong, L., Grand, G., Lew, A.~K., Goodman, N.~D., Mansinghka, V.~K., Andreas,
  J., and Tenenbaum, J.~B. (2023).
\newblock From word models to world models: Translating from natural language
  to the probabilistic language of thought.
\newblock {\em arXiv preprint arXiv:2306.12672}.

\bibitem[Wu et~al., 2024]{wu2024fine}
Wu, Z., Hu, Y., Shi, W., Dziri, N., Suhr, A., Ammanabrolu, P., Smith, N.~A.,
  Ostendorf, M., and Hajishirzi, H. (2024).
\newblock Fine-grained human feedback gives better rewards for language model
  training.
\newblock {\em Advances in Neural Information Processing Systems}, 36.

\bibitem[Xu et~al., 2024]{xu2024perfect}
Xu, T., Helenowski, E., Sankararaman, K.~A., Jin, D., Peng, K., Han, E., Nie,
  S., Zhu, C., Zhang, H., Zhou, W., et~al. (2024).
\newblock {The Perfect Blend}: {Redefining RLHF with Mixture of Judges}.
\newblock {\em arXiv preprint arXiv:2409.20370}.

\bibitem[Yang and Allenby, 2003]{yang2003modeling}
Yang, S. and Allenby, G.~M. (2003).
\newblock Modeling interdependent consumer preferences.
\newblock {\em Journal of Marketing Research}, 40(3):282--294.

\bibitem[Yao et~al., 2023]{yao2023instructions}
Yao, J., Yi, X., Wang, X., Wang, J., and Xie, X. (2023).
\newblock From instructions to intrinsic human values--a survey of alignment
  goals for big models.
\newblock {\em arXiv preprint arXiv:2308.12014}.

\bibitem[Yudkowsky, 2004]{yudkowsky2004coherent}
Yudkowsky, E. (2004).
\newblock Coherent extrapolated volition.
\newblock {\em Singularity Institute for Artificial Intelligence}.

\bibitem[Yudkowsky, 2015]{yudkowsky2015kansi}
Yudkowsky, E. (2015).
\newblock {Known-Algorithm} {Non-Self-Improving} {Agent}.
\newblock {\em Arbital}.
\newblock \url{https://arbital.com/p/KANSI/}.

\bibitem[Yudkowsky, 2016]{yudkowsky2016ai}
Yudkowsky, E. (2016).
\newblock The {AI} alignment problem: {Why} it is hard, and where to start.
\newblock
  \url{https://intelligence.org/2016/12/28/ai-alignment-why-its-hard-and-where-to-start/}.

\bibitem[Yudkowsky, 2019]{yudkowsky2019coherent}
Yudkowsky, E. (2019).
\newblock Coherent decisions imply consistent utilities.
\newblock {\em Less Wrong}.
\newblock \url{https://www.lesswrong.com/posts/RQpNHSiWaXTvDxt6R}.

\bibitem[Zhi-Xuan, 2022]{zhixuan2022what}
Zhi-Xuan, T. (2022).
\newblock {W}hat {S}hould {AI} {O}we {T}o {U}s? {A}ccountable and {A}ligned
  {AI} {S}ystems via {C}ontractualist {AI} {A}lignment.
\newblock {\em AI Alignment Forum}.
\newblock \url{https://www.alignmentforum.org/posts/Cty2rSMut483QgBQ2}.

\bibitem[Zhi-Xuan et~al., 2024a]{zhixuan2024infinite}
Zhi-Xuan, T., Kang, G., Mansinghka, V., and Tenenbaum, J.~B. (2024a).
\newblock Infinite ends from finite samples: Open-ended goal inference as
  top-down {B}ayesian filtering of bottom-up proposals.
\newblock {\em Proceedings of the Annual Meeting of the Cognitive Science
  Society}, 46(46).

\bibitem[Zhi-Xuan et~al., 2020]{zhi2020online}
Zhi-Xuan, T., Mann, J., Silver, T., Tenenbaum, J., and Mansinghka, V. (2020).
\newblock Online {B}ayesian goal inference for boundedly rational planning
  agents.
\newblock {\em Advances in Neural Information Processing Systems},
  33:19238--19250.

\bibitem[Zhi-Xuan et~al., 2024b]{zhi2024pragmatic}
Zhi-Xuan, T., Ying, L., Mansinghka, V., and Tenenbaum, J.~B. (2024b).
\newblock Pragmatic instruction following and goal assistance via cooperative
  language-guided inverse planning.
\newblock In {\em Proceedings of the 23rd International Conference on
  Autonomous Agents and Multiagent Systems}, pages 2094--2103.

\bibitem[Zhou and Li, 2022]{zhou2022hierarchical}
Zhou, W. and Li, W. (2022).
\newblock A hierarchical {B}ayesian approach to inverse reinforcement learning
  with symbolic reward machines.
\newblock In {\em International Conference on Machine Learning}, pages
  27159--27178. PMLR.

\bibitem[Zhu et~al., 2023]{zhu2023principled}
Zhu, B., Jordan, M., and Jiao, J. (2023).
\newblock Principled reinforcement learning with human feedback from pairwise
  or k-wise comparisons.
\newblock In {\em International Conference on Machine Learning}, pages
  43037--43067. PMLR.

\bibitem[Zhuang and Hadfield-Menell, 2020]{zhuang2020consequences}
Zhuang, S. and Hadfield-Menell, D. (2020).
\newblock Consequences of {Misaligned} {AI}.
\newblock In {\em Proceedings of the 34th {International} {Conference} on
  {Neural} {Information} {Processing} {Systems}}.

\bibitem[Ziebart et~al., 2010]{ziebart2010modeling}
Ziebart, B.~D., Bagnell, J.~A., and Dey, A.~K. (2010).
\newblock Modeling interaction via the principle of maximum causal entropy.
\newblock In {\em Proceedings of the 27th {International} {Conference} on
  {Machine} {Learning}}, pages 1255--1262.

\bibitem[Ziebart et~al., 2008]{ziebart2008maximum}
Ziebart, B.~D., Maas, A.~L., Bagnell, J.~A., Dey, A.~K., et~al. (2008).
\newblock Maximum entropy inverse reinforcement learning.
\newblock In {\em Proceedings of the 23rd National Conference on Artificial
  Intelligence - Volume 3}, AAAI'08, page 1433–1438.

\end{thebibliography}

\end{document}